\documentclass[journal]{IEEEtran}
\ifCLASSINFOpdf
\else
\fi

\usepackage{enumitem}
\usepackage{afterpage}
\usepackage{algpseudocode}
\usepackage[ruled,linesnumbered]{algorithm2e}
\usepackage{xcolor}
\usepackage{epstopdf}
\usepackage{amsmath,bm}
\usepackage{multirow}
\usepackage{url}
\usepackage{amssymb}
\usepackage{bbding}
\usepackage{bbm}
\usepackage{marginnote}
\usepackage[noabbrev,capitalize]{cleveref}
\usepackage{makecell}
\usepackage{array}
\usepackage{rotating}
\usepackage{balance}
\usepackage[normalem]{ulem}
\useunder{\uline}{\ul}{}

\def\BibTeX{{\rm B\kern-.05em{\sc i\kern-.025em b}\kern-.08emT\kern-.1667em\lower.7ex\hbox{E}\kern-.125emX}}

\newcommand{\indzheng}{\color{black}}
\newcommand{\indshihao}{\color{black}}
\newcommand{\tcsvt}{\color{black}}
\newcommand{\tcsvtminor}{\color{black}}

\renewcommand{\Comment}{}
\newcommand{\SH}{}
\newcommand{\chengb}{}
\newcommand{\cheng}{}

\newcommand{\mm}{}

\begin{document}

\title{TrajSV: A Trajectory-based Model for Sports Video Representations and Applications}

\author{Zheng Wang\IEEEauthorrefmark{1} \IEEEcompsocitemizethanks{\IEEEcompsocthanksitem\IEEEauthorrefmark{1}Equal contribution.},
        Shihao Xu\IEEEauthorrefmark{1},
        Wei Shi\IEEEauthorrefmark{2} \IEEEcompsocitemizethanks{\IEEEcompsocthanksitem\IEEEauthorrefmark{2}Wei Shi is the corresponding author.}
\IEEEcompsocitemizethanks{
\IEEEcompsocthanksitem
Zheng Wang, Shihao Xu and Wei Shi are with Huawei Technologies, Co., Ltd. E-mail: \{wangzheng155,shihao.xu,w.shi\}@huawei.com\\
}}

\markboth{IEEE TRANSACTIONS ON CIRCUITS AND SYSTEMS FOR VIDEO TECHNOLOGY,~Vol.~1, No.~1, December~2024}%
{Shell \MakeLowercase{\textit{et al.}}: TrajSV: A Trajectory based Model for Sports Video Representations and Applications
}
\maketitle

%\IEEEtitleabstractindextext{%
\begin{abstract}
Sports analytics has received significant attention from both academia and industry in recent years. Despite the growing interest and efforts in this field, several issues remain unresolved, including (1) data unavailability, (2) lack of an effective trajectory-based framework, and (3) requirement for sufficient supervision labels.
In this paper, we present TrajSV, a trajectory-based framework that addresses various issues in existing studies. TrajSV comprises three components: data preprocessing, Clip Representation Network (CRNet), and Video Representation Network (VRNet). The data preprocessing module extracts player and ball trajectories from sports broadcast videos. CRNet utilizes a trajectory-enhanced Transformer module to learn clip representations based on these trajectories. Additionally, VRNet learns video representations by aggregating clip representations and visual features with an encoder-decoder architecture. Finally, a triple contrastive loss is introduced to optimize both video and clip representations in an unsupervised manner.
{\mm The experiments are conducted on three broadcast video datasets to verify the effectiveness of TrajSV for three types of sports (i.e., soccer, basketball, and volleyball) with three downstream applications (i.e., sports video retrieval, action spotting, and video captioning)}.
The results demonstrate that TrajSV achieves state-of-the-art performance in sports video retrieval, showcasing a nearly 70\% improvement. It outperforms baselines in action spotting, achieving state-of-the-art results in 9 out of 17 action categories, and demonstrates a nearly 20\% improvement in video captioning. Additionally, we introduce a deployed system along with the three applications based on TrajSV.

% (with {\Comment{11.1-28.1\%}} improvement in HR@1 over the best baseline method),
% % . It can also be easily incorporated into existing frameworks for fine-tuning, and it 
% {\cheng and} improves the baseline of action spotting by {\Comment{0.6\%}} and video captioning by up to {\Comment{3.52\%}}.

\end{abstract}

\begin{IEEEkeywords}
sports video representation, triple contrastive learning, trajectory
\end{IEEEkeywords}
%}

%\IEEEdisplaynontitleabstractindextext

\IEEEpeerreviewmaketitle

\section{INTRODUCTION}
\label{sec:intro}

\IEEEPARstart{W}{ith} the advancements in high-resolution cameras and sensors, significant amounts of data can now be gathered from sports events and subjected to analysis. This data analysis holds the potential to uncover valuable insights for a broad range of sports application scenarios, including coaching~\cite{DBLP:conf/kdd/WangLCJ19,DBLP:conf/iui/ShaLYCRM16,DBLP:conf/sdm/ZhangWLY22,DBLP:conf/kdd/DecroosHD18,DBLP:journals/tkdd/DiKSL18} and broadcasting~\cite{DBLP:conf/mmasia/HaruyamaTOH20,DBLP:conf/bigdataconf/ProbstRSSR18,DBLP:conf/cikm/KabaryS13,DBLP:conf/cvpr/CioppaDGGDGM20}.
Although {\cheng some} sports analytic techniques have {\cheng been developed for} the aforementioned applications, {\cheng they still suffer from some of the issues among (1) data unavailability, (2) lack of an effective trajectory-based framework, and (3) requirement of sufficient supervision labels, as summarized in Table~\ref{tab:summary}. Next, we discuss these issues and present our proposals for solving them.}

\begin{table}[]
\setlength{\tabcolsep}{13pt}
\centering
%\small
\caption{The issues in existing studies ($\surd$ indicates the issue exists and $\times$ otherwise). Issue-1: Data unavailability, Issue-2: Lack of an effective trajectory-based framework, and Issue-3: Requirement of sufficient supervision labels.
}
\vspace*{-3mm}
\begin{tabular}{lccc}
\hline
 & Issue-1 & Issue-2 & Issue-3 \\ \hline
{\mm play2vec}~\cite{DBLP:conf/kdd/WangLCJ19} &$\surd$  &$\times$  &$\times$ \\
{\mm Chalkboard}~\cite{DBLP:conf/iui/ShaLYCRM16} &$\surd$  &$\surd$  &$\times$ \\ 
{\mm LearnRank}~\cite{DBLP:journals/tkdd/DiKSL18} &$\surd$  &$\times$  &$\times$ \\
{\mm SimScene}~\cite{DBLP:conf/mmasia/HaruyamaTOH20} &$\times$  &$\surd$  &$\surd$ \\ 
{\mm  LRML}~\cite{DBLP:conf/mm/HaruyamaTOH19} &$\times$  &$\surd$  &$\surd$ \\ 
{\mm CALF}~\cite{DBLP:conf/cvpr/CioppaDGGDGM20} &$\times$  &$\surd$  &$\surd$ \\
{\mm LRRCN}~\cite{kong2019joint} &$\times$  &$\surd$  &$\surd$ \\
{\mm Baidu-VC}~\cite{Mkhallati2023SoccerNetCaption} &$\times$  &$\surd$  &$\surd$ \\
{\mm VCMG}~\cite{qi2019sports} &$\times$  &$\surd$  &$\surd$ \\
\hline
\end{tabular}
\label{tab:summary}
\vspace*{-5mm}
\end{table}

\textbf{(1) Data {\cheng unavailability}.} Many sports-related studies~\cite{DBLP:conf/kdd/WangLCJ19,DBLP:conf/iui/ShaLYCRM16,DBLP:journals/tkdd/DiKSL18} require the collection of player or ball movement data (i.e., trajectories) using professional equipment like GPS trackers or optical tracking cameras, which can be expensive. For example, studies on sports play retrieval~\cite{DBLP:conf/kdd/WangLCJ19, DBLP:conf/iui/ShaLYCRM16, DBLP:journals/tkdd/DiKSL18, DBLP:journals/tkde/WangLC23} utilize 4K resolution cameras in an optical tracking system, costing thousands of dollars for tracking a single game~\cite{datasource}, not including the expense of a professional team to operate it. Additionally, the collected data is limited to specific sports games (e.g., France's Ligue de Football Professionnel (LFP) in these studies), making it challenging to generalize the research results to games without available movement data. This highlights the need for more cost-effective and accessible methods of collecting sports data.
%
% One promising approach is 
{\cheng We propose} to use publicly available sports broadcast videos as an alternative data source. With the widespread availability of sports broadcasts on the internet, this approach has the potential to provide a large volume of data at a relatively low cost. 
% In the following sections, we will explore the potential of using sports broadcast videos for analyzing sports data.%, and present a data preprocessing approach for extracting the movement information (i.e., trajectories) from such broadcasts.

\textbf{(2) Lack of an effective trajectory-based framework.} Learning sports data representation is a widely used technique with applications in sports video retrieval~\cite{DBLP:conf/mmasia/HaruyamaTOH20,DBLP:conf/mm/HaruyamaTOH19}, action spotting~\cite{DBLP:conf/cvpr/CioppaDGGDGM20, kong2019joint}, and video captioning~\cite{Mkhallati2023SoccerNetCaption, qi2019sports}. However, few studies have explored a trajectory-based framework that can support different downstream tasks. In many cases, existing techniques focus on utilizing data for specific tasks, like sports video retrieval, often overlooking the underlying structure and information embedded in trajectories, which could enhance flexibility for various tasks. To address this gap, we investigate a trajectory-based framework that extracts spatio-temporal trajectories from raw sports broadcast videos, deriving both video-level and clip-level representations. These representations support a variety of downstream tasks and can be fine-tuned for improved results. The rationale of extracting trajectories is that it embeds both spatial and temporal features of a sports game, providing a shared foundation for different tasks. For example, the trajectory of a ball or player can be used for recognizing specific actions (action spotting) and serve as information for retrieving similar scenes based on extracted trajectory patterns (sports retrieval). Recognizing these common patterns behind trajectories presents an opportunity to develop techniques for various sports applications.
{\cheng We} note that some recent video models (e.g., VideoMAE~\cite{DBLP:journals/corr/abs-2203-12602} or X-CLIP~\cite{DBLP:conf/eccv/NiPCZMFXL22}) have shown advancements in addressing open-world applications. However, we empirically observe that these models {\cheng do not perform very well} when applied to sports, a domain highly related to player movements, exhibiting unique characteristics not present in other video types.

\textbf{(3) Requirement of sufficient supervision labels.} Several existing studies~\cite{DBLP:conf/mmasia/HaruyamaTOH20,DBLP:conf/mm/HaruyamaTOH19} have been developed to learn sports video representations under a supervised learning framework. However, creating labeled video datasets for supervised learning can be a time-consuming and expensive process. For example, in studies on soccer scene retrieval~\cite{DBLP:conf/mmasia/HaruyamaTOH20,DBLP:conf/mm/HaruyamaTOH19}, authors manually annotate video footage to identify soccer events such as shots or corner kicks. The resulting dataset is limited to a set of specific events, and its effectiveness heavily relies on sufficient labeled data. The limitation highlights the need of developing an unsupervised learning framework that utilizes the inherent structure of sports videos and identifies trajectory patterns without explicit labels. Unsupervised learning offers a more flexible way to learning representations.

\textbf{{\chengb Summary of} our solution.} In this paper, we develop a trajectory-based framework for sports video representations, called TrajSV, which avoids the aforementioned three issues.
{\mm \underline{For (1),} TrajSV differs from existing techniques~\cite{DBLP:conf/kdd/WangLCJ19, DBLP:conf/iui/ShaLYCRM16, DBLP:journals/tkdd/DiKSL18, DBLP:journals/tkde/WangLC23} that rely on limited spatial data for training and evaluation. In contrast, TrajSV takes advantage of raw broadcast videos available on the Web, which significantly expands the accessibility and diversity of data. To achieve this, TrajSV incorporates a comprehensive data preprocessing pipeline, including video segmentation, camera calibration, and multi-object tracking.}
{\mm \underline{For (2),} TrajSV focuses on task-agnostic representations for various sports analytic tasks that rely on trajectories, which provides new insight for continued research in this line of studies. To achieve this, TrajSV introduces CRNet and VRNet. In CRNet, the model effectively captures sequential data (e.g., videos), and combines visual and spatial features to derive the clip representations. Additionally, VRNet incorporates an attention mechanism implemented by two attention blocks (called MAB and MSB) via an encoder-decoder architecture. Our solution addresses the issue of uneven clip contribution to video representation by prioritizing the most important clips. This leads to an overall improvement in the video representation's quality. %It is worth noting that while existing methods may employ neural networks for learning sports representations, TrajSV uniquely introduces to derive both clip-level and video-level representations in one framework to support different sports video analyses.
}
{\mm \underline{For (3),} we present a novel approach called triple contrastive learning, which optimizes both video and clip representations in an unsupervised manner by comparing their similarities. The rationale behind this approach is the relative ease of identifying the more similar video among two candidates when a query video is provided. To further enhance the comparison between videos, we introduce a triple contrastive loss that takes into account three aspects: 1) trajectory patterns within clips, 2) dependencies of clips in a video, 3) and mutual information derived from different variants of the same video. 
%The rationale behind using contrastive learning is that it is more feasible to identify the more similar video among two candidates given a query video, compared to directly measuring the similarity between two videos. To achieve this, we design two variants for an anchor video, called intra-clip video and inter-clip video. Then, a triple contrastive loss is introduced by pairwise contrasting the three kinds of videos, which aims to enable the learning of representations by optimizing trajectory patterns inside clips, dependencies of clips in a video, and mutual information between the two variants from the same anchor.}

\textbf{Novelty.} We discuss several novel aspects of the TrajSV design as follows. 
\underline{1)} Utilizing Raw Broadcast Videos: TrajSV is built on raw broadcast videos available on the {\chengb Web}, which overcomes the limitation of sports movement data availability. With a larger and diverse dataset, TrajSV improves the performance of the model. 
\underline{2)} Task-Agnostic Representations: TrajSV focuses on trajectory-based representations in sports videos, allowing their versatile use in various sports analytic tasks that depend on trajectories. This enhances flexibility in sports video analysis.
\underline{3)} The Architecture: TrajSV incorporates three key components, including data preprocessing, CRNet, and VRNet. The incorporation of both clip-level and video-level representations in one framework is a unique feature not found in previous studies. 
%where CRNet is responsible for learning representations of clips, while VRNet further learns a video representation based on the aggregated clip representations. The incorporation of both clip-level and video-level representations in one framework is a unique feature not found in previous studies. 
\underline{4)} Triple Contrastive Learning: it uniquely considers three contrasts of its kind. First, contrasting complex movement patterns in sports videos helps in understanding actions and events within each clip. Second, contrasting dependencies between different clips within a video helps understand temporal continuity and context. Third, contrasting variants of the same video captures diverse cues, enhancing robust representation.
%it uniquely considers three contrasts . First, sports videos often contain complex movement patterns of players and the ball, and contrasting these patterns helps in understanding the actions and events occurring within each clip. Second, it contrasts the dependencies between different clips within a video, which enables the model to understand the temporal continuity and context of the video as a whole, rather than treating each clip individually. Third, by contrasting the information shared between different variants of the same video, it enables the model to capture diverse information, leading to a more robust representation.

Our contributions can be summarized as follows. 

\begin{itemize}
    \item We propose TrajSV, a trajectory-based framework for learning sports video representations. Our model addresses three issues in this field: 1) it utilizes raw broadcast videos, making it widely applicable, 2) it learns task-agnostic representations that can be used for different applications, and 3) it does not require supervision labels. %To the best of our knowledge, TrajSV is the first model of its kind. 
    \item {\indzheng We present a deployed system (used in a real sports video search engine) based on TrajSV, and three applications built on top of the system: 1) sports video retrieval, 2) action spotting, and 3) video captioning, which correspond to using the representations {\cheng at} both video-level (i.e., 1) and clip-level (i.e., 2 and 3).} To the best of our knowledge, this is the first industry system of its kind. 
    \item {\mm We evaluate the TrajSV on \emph{three} broadcast datasets with various baselines, and the results demonstrate its effectiveness across \emph{three} types of sports (i.e., soccer, basketball, and volleyball), for the \emph{three} applications. %This provides a comprehensive evaluation of the pre-trained representations' abilities. Compared to the best baseline methods, TrajSV achieves state-of-the-art performance in sports video retrieval (); and can incorporate with existing models for fine-tuning, which improves the action spotting and video captioning.
    The results demonstrate TrajSV's state-of-the-art performance in sports video retrieval, with an improvement of nearly 70\%. Additionally, it improves baselines in action spotting, achieving state-of-the-art results in 9 out of 17 action categories, and in video captioning, exhibiting an improvement of nearly 20\%.
    }
\end{itemize}

%{\Comment{11.1-28.1\%}} improvement in sports video retrieval; and it can be easily incorporated into existing frameworks for fine-tuning, and improve the action spotting by up to {\Comment{0.6\%}} and video captioning by up to {\Comment{3.52\%}}.

\section{RELATED WORK}
\label{sec:related}

\subsection{Sports Data Analytics}
We review the literature on sports data analytics in terms of different data types, i.e., spatio-temporal data and video data. 
For spatio-temporal data, it is to track the moving objects in a sports game with some professional devices, e.g., optical tracking cameras have been installed in a sports field to track the trajectories of players and the ball in a soccer game~\cite{DBLP:conf/kdd/WangLCJ19}. The data attracts many research efforts~\cite{DBLP:conf/kdd/WangLCJ19, DBLP:journals/tkde/WangLC23, DBLP:conf/iui/ShaLYCRM16, DBLP:journals/tkdd/DiKSL18, DBLP:conf/sdm/ZhangWLY22, DBLP:conf/kdd/DecroosHD18,zhang2025billiards,xu2025sportstraj}. For example, 
play2vec~\cite{DBLP:conf/kdd/WangLCJ19} learns a representation based on spatio-temporal sports trajectories using a Seq2Seq architecture.
Chalkboard~\cite{DBLP:conf/iui/ShaLYCRM16} is also based on trajectory inputs and adopts a role-based pairwise matching strategy to compute the similarity for the retrieval task. {\tcsvtminor{Sports-Traj~\cite{xu2025sportstraj} advances sports analytics by introducing a unified model for trajectory prediction, imputation, and spatio-temporal recovery. It extends Mamba~\cite{gu2023mamba} with a bidirectional temporal design and integrates a Transformer encoder to enhance temporal feature extraction.
}}
\if 0
Wang et al.~\cite{DBLP:conf/kdd/WangLCJ19} study the similarity between two sports plays, where a play corresponds to a set of trajectories of players and the ball. Further, they study a similar play retrieval problem~\cite{DBLP:journals/tkde/WangLC23} of searching plays from a database, which are similar to a given query play. Sha et al.~\cite{DBLP:conf/iui/ShaLYCRM16} also study the similarity of the two plays. They propose a similarity measurement by first aligning trajectories in terms of different roles in a sports game, then pairwise matches trajectory points to compute the similarity within each {\cheng alignment}, and aggregates the similarities as the final similarity. 
%
In addition, Di et al.~\cite{DBLP:journals/tkdd/DiKSL18} use a convolutional autoencoder to extract features from sports tracking data and develop a distributed system for the sports play retrieval task using a learning-to-rank model.
%
Zhang et al.~\cite{DBLP:conf/sdm/ZhangWLY22} collect the layouts (i.e., locations) of billiards balls, and perform the billiards layout prediction, generation, and retrieval tasks based on the dataset. 
%
Decroos et al.~\cite{DBLP:conf/kdd/DecroosHD18} explore a tactics detection problem based on the collected event-stream data from professional soccer games. 
%
\fi
Overall, this line of research is based on spatio-temporal data tracked from professional cameras, and this makes it difficult to apply these studies to sports games where the tracking data is not available. In this work, we explore alternative data sources (i.e., broadcast videos) that have better availability {\cheng on} the Web, to support a wider range of tasks.

For video data, the sports analytic tasks include 
sports video retrieval~\cite{DBLP:conf/mmasia/HaruyamaTOH20, DBLP:conf/mm/HaruyamaTOH19, DBLP:conf/bigdataconf/ProbstRSSR18, DBLP:conf/cikm/KabaryS13, DBLP:conf/sigir/KabaryS14},
action spotting~\cite{DBLP:conf/cvpr/CioppaDGGDGM20,DBLP:conf/cvpr/GiancolaG21,DBLP:conf/icpr/TomeiBCBC20, DBLP:journals/corr/abs-2106-14447,DBLP:journals/tmm/WuWBDLCDX23,li2024contextualized,wang2024knowledge}, 
video captioning~\cite{Mkhallati2023SoccerNetCaption},
%replay grounding~\cite{DBLP:conf/icmcs/ZhaoJHZ06,DBLP:conf/icassp/WangCX05}, 
camera shot segmentation~\cite{DBLP:conf/icmcs/HuHWL07,DBLP:conf/icassp/WangCX05}, 
%field localization~\cite{DBLP:conf/cvpr/HomayounfarFU17}, 
camera calibration~\cite{DBLP:conf/wacv/TheinerE23,DBLP:conf/spieSR/FarinKWE04}, 
tracking, re-identification~\cite{DBLP:journals/pami/LuTLM13,DBLP:conf/eccv/SullivanC06,kim2022cost}, 
and summarization~\cite{DBLP:journals/tmm/Tejero-de-Pablos18}.
%tactic recognition~\cite{kong2022spatio}. 
The work~\cite{DBLP:conf/cvpr/DeliegeCGSDNGMD21} provides a detailed evaluation of these tasks based on their collected datasets. For example,
SimScene~\cite{DBLP:conf/mmasia/HaruyamaTOH20,DBLP:conf/mm/HaruyamaTOH19} learns video clip representations based on human annotations of some common soccer scenes (e.g., shot, corner kick), where it utilizes a BiLSTM-based model and incorporates multimodal features (e.g., images, audio, and text) into the representation.
%
%Probst et al.~\cite{DBLP:conf/bigdataconf/ProbstRSSR18} develop a sports analysis system, which detects sports events and visualizes the statistics in real-time, and it supports 4 types of user queries to retrieve a video clip stored in a database.
%
SportSense~\cite{DBLP:conf/cikm/KabaryS13} is a system that supports interactive sports video retrieval, where a user can freely sketch a path on the soccer field, and it returns video clips that match the sketched path. Further, an auto-suggest feature is deployed into the system, where it suggests potential directions when the user sketches a path, and thus it relaxes the use of system memory and increases the retrieval speed~\cite{DBLP:conf/sigir/KabaryS14}.
The action spotting task~\cite{DBLP:conf/cvpr/GiancolaADG18}, entails pinpointing both the timing and type of a particular action within a video, where each action is annotated with a single timestamp. CALF~\cite{DBLP:conf/cvpr/CioppaDGGDGM20} employs a context-aware loss function that emphasizes the temporal context surrounding each action, as opposed to solely focusing on a single annotated frame for spotting.
LRRCN~\cite{kong2019joint} presents a joint framework for athlete tracking and action recognition in sports videos, featuring a robust tracker for precise athlete localization and a long-term recurrent convolutional network for modeling temporal action cues.
The papers~\cite{DBLP:journals/tmm/WuWBDLCDX23, shih2017survey} provide extensive surveys on video action spotting across over 10 sports, examining datasets, methodologies, and applications.
{\tcsvtminor{Con-RPM~\cite{li2024contextualized} is a self-supervised model for group activity representation that captures evolving interactions in team sports using predictive coding. It integrates individual and scene context via a Transformer-based encoder-decoder, optimized with contrastive and adversarial learning. 
KARI~\cite{wang2024knowledge} enhances group activity recognition by leveraging concretized knowledge from training data. It models action co-occurrence and spatial distributions as structured maps to improve relation inference and individual representation.
}}
The video captioning task~\cite{Mkhallati2023SoccerNetCaption}, centers on generating textual comments anchored with single timestamps. This involves a two-stage process. First, a spotting model combines frame features into a single clip representation, which is then used to generate proposal timestamps. Further, these timestamps are subsequently utilized by the captioning model to produce the anchored comments. 
VCMG~\cite{qi2019sports} introduces a hierarchical neural network with motion representation and group relationship modules to improve sports video captioning.
Our work is different {\cheng from} these studies in two aspects. First, we learn representations based on raw broadcast videos instead of those short clips, where we derive both video-level and clip-level representations to support downstream tasks. Nevertheless, this is not the focus of existing studies.
Second, in contrast to the studies~\cite{DBLP:conf/mmasia/HaruyamaTOH20, DBLP:conf/bigdataconf/ProbstRSSR18}, our model is trained {\cheng with} an unsupervised learning framework (i.e., triple contrastive learning) without human annotations.

\subsection{Video Representation Learning} 
{\mm Video representation learning is a fundamental task that aims to capture informative representations from video data, enabling various downstream applications such as action recognition~\cite{han2019video, sun2019videobert,piergiovanni2020evolving}, and video captioning~\cite{sun2019videobert}. For example, 
Qian et al.~\cite{han2019video} learn video representations for human action recognition, it introduces a Dense Predictive Coding (DPC) framework, which enables the learning of dense spatio-temporal representations from videos by predicting future representations. 
%
%Sun et al.~\cite{sun2019videobert} propose a joint visual-linguistic model called VideoBERT, which utilizes the BERT model to learn high-level features from sequences of visual and linguistic tokens. 
%
%Piergiovanni et al.~\cite{piergiovanni2020evolving} introduce an approach to learn generic video representations from unlabeled data, by leveraging multimodal learning and evolutionary search to optimize loss functions.}
%
Recently, contrastive learning, e.g., video-text alignment~\cite{DBLP:conf/iccv/YangBG21, DBLP:conf/eccv/NiPCZMFXL22, DBLP:journals/ijon/LuoJZCLDL22, DBLP:conf/cvpr/0005XZWGHH22, DBLP:conf/cvpr/QianMG0WBC21} has also been widely used to learn video representations. For example,
%
%TACo~\cite{DBLP:conf/iccv/YangBG21} studies a multimodal representation learning for video-text alignment, where it aligns visual contents with both word-level and sentence-level contrasts.
%
%X-CLIP~\cite{DBLP:conf/mm/MaXSYZJ22} further improves the video-text retrieval task with a multi-grained contrastive learning, it designs three contrasts (i.e., fine-grained, coarse-grained and cross-grained) into the loss function.
\SH{X-CLIP~\cite{DBLP:conf/eccv/NiPCZMFXL22} improves the video-text retrieval task using a cross-frame communication transformer and a multi-frame integration transformer, and {\cheng maximizes} the similarity between the paired video and text.}
%
%CLIP4Clip~\cite{DBLP:journals/ijon/LuoJZCLDL22} studies video-text retrieval based on pre-trained CLIP model~\cite{DBLP:conf/icml/RadfordKHRGASAM21} by transferring the knowledge of the image-text retrieval task, and the CLIP4Clip can be trained in an end-to-end manner.
%
%CACL~\cite{DBLP:conf/cvpr/0005XZWGHH22} incorporates a 3D CNN and a video Transformer to generate strong positive pairs, which enables the model to learn more effective representations.
%
%CVRL~\cite{DBLP:conf/cvpr/QianMG0WBC21} explores what data augmentations make video representations good, and find both spatial and temporal {\cheng ones} are important cues.
%
In this work, we develop a trajectory-based framework \SH{that combines scene-based video representations} to learn sports video representations, which can be trained by comparing similar sports videos with dissimilar ones in an unsupervised way. %To our best knowledge, this is the first of its kind.
\section{PRELIMINARIES AND PROBLEM STATEMENT}
\label{sec:problem}

\if 0
\begin{figure*}
  \centering
  \includegraphics[width=\linewidth]{figures/pipeline6.drawio.pdf}
  %\vspace{-2mm}
  \caption{{\mm The overall framework of TrajSV. In Data Preprocessing, the raw broadcast video $V$ undergoes video segmentation to obtain multiple clips $C_1, ..., C_n$. For each clip, the underlying camera parameters are estimated, and these parameters are then utilized in a multi-object tracking process to extract trajectories. In CRNet, these extracted trajectories are represented as a sequence of segment matrices $\mathcal{M}$, which are further processed through a Transformer encoder to obtain a representation $\mathbf{t}$. The clip representation $\mathbf{c}$ is generated by concatenating (denoted by $\oplus$) the $\mathbf{t}$ with a visual-based representation $\mathbf{y}$. In VRNet, it encodes the clip representations $\mathbf{c}$ into features $\mathbf{E}$, which are then decoded to produce the final video representation $\mathbf{v}$. Finally, a triple contrastive loss $\mathcal{L}$ is introduced to optimize these representations in an unsupervised manner.
  }}\label{fig:framework}
  %\vspace*{-3mm}
\end{figure*}
\fi

%\subsection{Preliminaries}
\noindent\textbf{Sports Video.} Let $V$ denote a sports video, representing an untrimmed broadcast video such as a soccer game on platforms like YouTube. The video $V$ consists of a sequence of images, i.e., $V = \langle I_1, I_2, ..., I_{|V|} \rangle$, where $I_i$ denotes an image at the $i^{th}$ frame.

\smallskip
\noindent\textbf{Sports Clip.} A sports clip $C$ refers to a portion of $V$.
%, and
%$V$ that starts from the $i^{th}$ frame and ends at the $j^{th}$ frame, i.e., $V[i,j]=\langle I_i, I_{i+1}, ..., I_j \rangle$, {\cheng for $1\le i\le j\le |V|$}. any clip from a video corresponds to a (shorter) video. 
Due to the nature of broadcast videos, 
% we denote a clip is that of a sports-related content 
{\cheng we focus clips that are sports-related}
(e.g., a clip {\cheng that} contains offensive movements which lead to a goal), 
% and not that of the sports-unrelated content 
{\cheng but not those that are sports-unrelated}
(e.g., {\cheng a video clip that contains} a dance performance at the opening of a soccer game). 

\smallskip
\noindent\textbf{Trajectory.} A trajectory $T$ corresponds to a sequence of locations, i.e., $(x_1, y_1), (x_2, y_2), ..., (x_{|T|}, y_{|T|})$, that captures the movement of an object (e.g., players or the ball) in a sports clip, where $(x_i, y_i)$ denotes the location tracked from the image at the $i^{th}$ frame. Therefore, a sports clip {\cheng that contains multiple moving objects such as players}
% contained multiple objects 
can be represented by a set of multiple trajectories.

\smallskip
\noindent\textbf{Problem Statement.} Given a sports video $V$, we aim to, {\cheng 1) learn} a vector representation $\mathbf{v}$ for the $V$, and {\cheng 2) derive} the representation $\mathbf{c}_i$ ($1 \le i \le n$) for any sports clip $C_i$ contained in the video $V$, $n$ denotes the number of clips in the $V$.

Note that our goal is to learn representations of sports videos {\cheng at} both the video-level and the clip-level, such that the derived representations could be utilized in various video-based applications (e.g., sports video retrieval~\cite{DBLP:conf/mmasia/HaruyamaTOH20}) and clip-based applications (e.g., action spotting~\cite{DBLP:conf/cvpr/CioppaDGGDGM20}, video captioning~\cite{Mkhallati2023SoccerNetCaption}), respectively.
\section{Model Architecture}
\label{sec:model}

\begin{figure}
  \centering
  \includegraphics[width=\linewidth, trim={5mm 77.6cm 57.5cm 30.6cm}, clip]{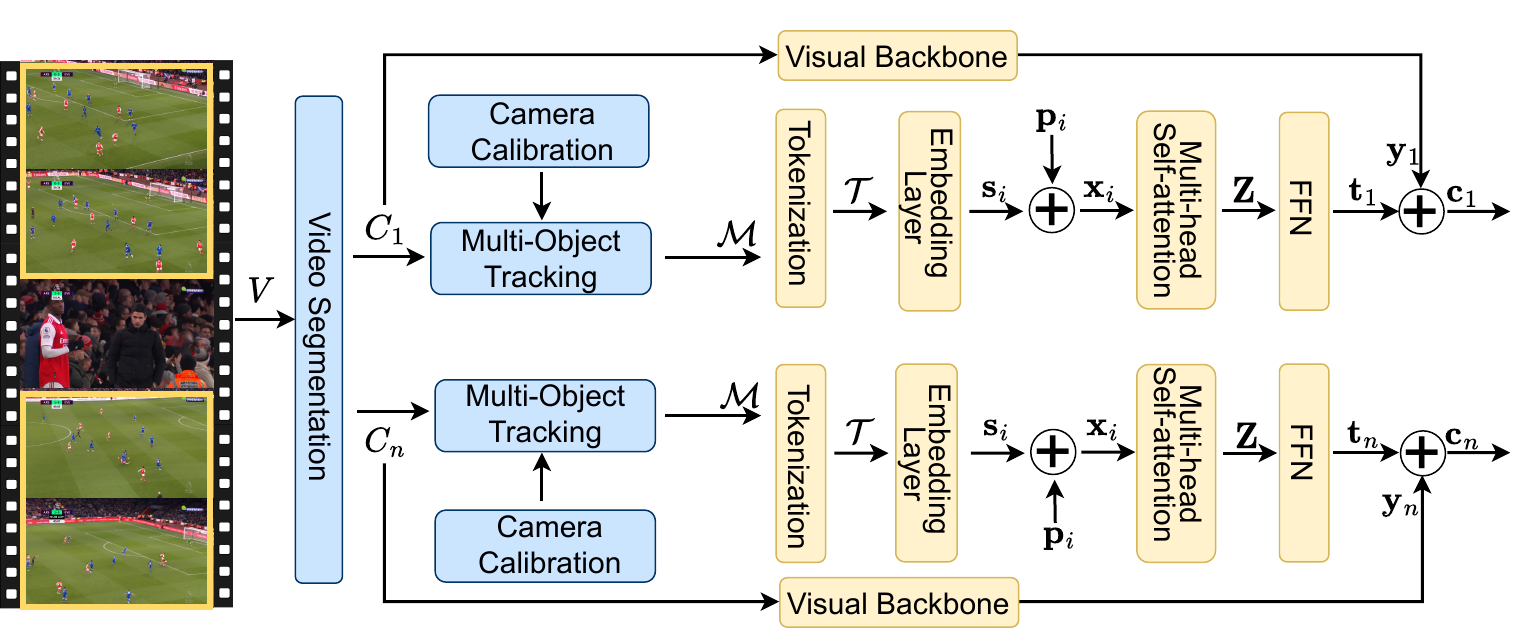}
  \vspace{-3mm}
  \caption{{\mm {\tcsvtminor{The CRNet framework. A raw broadcast video $V$ undergoes video segmentation to obtain multiple clips $C_1, ..., C_n$. For each clip, the underlying camera parameters are estimated, and these parameters are then utilized in a multi-object tracking process to extract trajectories. Then, these extracted trajectories are represented as a sequence of segment matrices $\mathcal{M}$, which are further processed through a Transformer encoder to obtain a representation $\mathbf{t}$. The clip representation $\mathbf{c}$ is generated by concatenating (denoted by $\oplus$) the $\mathbf{t}$ with a visual-based representation $\mathbf{y}$.}}}}\label{fig:framework_crnet}
  \vspace*{-5mm}
\end{figure}

%\subsection{Overview of TrajSV}
%\label{sec:overview}
{\mm TrajSV presents a trajectory-based framework for sports video representations. It consists of three components, namely data preprocessing (Section~\ref{sec:data_preprocessing}), CRNet (Section~\ref{sec:crnet}) and VRNet
(Section~\ref{sec:grnet}). We investigate a triple contrastive learning method to train the TrajSV in an unsupervised manner (Section~\ref{sec:loss}). %Figure~\ref{fig:framework} illustrates an overall framework.

}

\subsection{Data Preprocessing}
\label{sec:data_preprocessing}
Raw sports videos consist of sequences of images, often containing irrelevant content and lacking spatial features crucial for player and ball movement analysis. To address these challenges, the data preprocessing involves three steps: (1) extracting sports clips, (2) extracting camera parameters from video images, and (3) tracking player and ball trajectories by mapping image coordinates to a sports field coordinate system using the obtained camera parameters for each clip.
\if 0
The raw sports video corresponds to a sequence of images. {\cheng The images} carry many sports-unrelated contents in broadcasts, and lack spatial features to describe the movements of players and the ball, which are important to give quantitative analytics for downstream tasks. {\mm To tackle the challenges, the data preprocessing component involves three key steps: (1) extracting sports clips, (2) obtaining camera parameters from the video images, and (3) tracking the trajectories of players and the ball by mapping image coordinates to a sports field coordinate system using the camera parameters for each clip.}
\fi

\smallskip
\noindent \textbf{Video Segmentation.} A key step in raw video editing is segmenting a long video into multiple clips. This task is modeled as a classification problem~\cite{DBLP:conf/cvpr/DeliegeCGSDNGMD21}, where each video frame is classified into predefined camera types (e.g., main camera center, close-up player). A segmented clip consists of consecutive frames with the same type. Our approach adopts a ResNet-based method~\cite{DBLP:conf/cvpr/CioppaDGGDGM20}, sequentially classifying the camera type at each frame based on its extracted ResNet features. Sports clips correspond to classes with specific camera types (e.g., main camera center, left and right). If a clip is highly fragmented (duration below a threshold, which is set to 0.4s based on  empirical findings), it is removed from the outputs.
\if 0
Video segmentation is the process of dividing a long video into multiple clips. This is a crucial step in editing raw video footage. The main idea of the operation is to model the task as a classification problem~\cite{DBLP:conf/cvpr/DeliegeCGSDNGMD21}, where it classifies each video frame among several pre-defined camera types (e.g., main camera center, close-up player), and a segmented clip forms as those consecutive frames with the same type. 
%
We adopt a ResNet-based approach~\cite{DBLP:conf/cvpr/CioppaDGGDGM20} for segmenting sports clips in a video, where it sequentially classifies the camera type at each frame based on its extracted ResNet features, and sports clips correspond to the classes with several certain camera types (e.g., main camera center, left and right).
% 
We remark that if a clip is very fragmented, i.e., it is with a duration smaller than a threshold, it will be removed from the outputs.
\fi

\smallskip
\noindent \textbf{Camera Calibration.} It is a crucial step in mapping trajectory points from pixel coordinates in a video image to world coordinates, such as a sports field. This process involves estimating camera parameters to achieve accurate mapping. Our calibration algorithm utilizes field markings, such as lines and circle segments in a soccer field, as reference points, following~\cite{DBLP:conf/wacv/TheinerE23}. The algorithm minimizes reprojection errors from pixel coordinates to world coordinates, calculated as the Euclidean distance between pixels detected by localization algorithms (e.g., DeepLabV3 \cite{DBLP:journals/corr/ChenPSA17}) and the reprojected field markings with known coordinates in a given sports field.
\if 0
Camera calibration is an essential step in mapping trajectory points from pixel coordinates in an image of a video to world coordinates such as a sports field. The camera calibration algorithm estimates the underlying camera parameters, which are then used to perform this mapping. Specifically, 
we develop a calibration algorithm using field markings such as lines and circle segments in a soccer field as reference points following~\cite{DBLP:conf/wacv/TheinerE23}. The algorithm minimizes the reprojection errors from pixel coordinates to world coordinates, which are calculated as the Euclidean distance between pixels detected by localization algorithms such as DeepLabV3 ResNet \cite{DBLP:journals/corr/ChenPSA17} and the reprojected field markings with known coordinates in a given sports field.
\fi

\smallskip
\noindent \textbf{Multi-Object Tracking (MOT).} MOT involves extracting trajectories of multiple objects in a video, utilizing a multi-task approach~\cite{DBLP:journals/ijcv/ZhangWWZL21} that includes object detection and re-identification. In object detection, the model locates object positions in each video frame using bounding boxes, outputting center coordinates. These coordinates are mapped to sports field coordinates using camera parameters from calibration. The model optimizes three parallel objectives~\cite{DBLP:journals/corr/abs-1904-07850}: heatmaps (for object center locations), object center offsets (for precise localization), and bounding box sizes (for height and width estimation). For re-identification (re-ID), the model recognizes the same object across different camera views. ResNet extracts re-ID features from bounding boxes, and the model classifies each box, determining whether to link the object to an existing track or create a new track.

{\tcsvt{
\smallskip
\noindent \textbf{Real-world Data Processing Challenges.} We discuss the real-world challenges in data processing and our engineering efforts to support the reproduction of TrajSV. (1) Handling Sports-Unrelated Content in Broadcasts: In real-world sports videos, non-sports-related content is often prevalent. To address this, we study video segmentation techniques to classify different camera types (e.g., main camera center, left, and right) to separate relevant sports content from irrelevant footage. (2) Camera Calibration for Sports Videos: Sports videos, especially those captured from various angles, require precise camera calibration. We use field markings, such as lines and circle segments on a soccer field, as reference points for calibration to ensure accurate spatial representation in the video frames. (3) Multi-Object Tracking for Key Sports Elements: Tracking critical elements like the ball is essential for sports analysis. Since the ball plays a central role in many sports, we fine-tune multi-object tracking (MOT) models on a sports-specific dataset. Additionally, we empirically set a weight parameter (i.e., 10) for smaller objects to enhance the tracking performance, particularly for the soccer ball. These engineering efforts are essential for ensuring that TrajSV can be reproduced and applied effectively in real-world scenarios.
}}

\if 0
MOT is a process of extracting trajectories of multiple objects of interest in a video. In this work, we consider a multi-task approach~\cite{DBLP:journals/ijcv/ZhangWWZL21} for the MOT that includes both object detection and re-identification.
%
For object detection, the model detects the {\cheng locations} of objects in each video frame by bounding boxes, which are outputted as the center coordinates of the bounding boxes. These locations are then mapped to the coordinates in a sports field using the underlying camera parameters obtained from camera calibration. To achieve this, the model optimizes three parallel objectives~\cite{DBLP:journals/corr/abs-1904-07850}: heatmaps (for estimating the locations of the object centers), object center offsets (for localizing objects more precisely), and bounding box sizes (for estimating the heights and widths of bounding boxes).
%
For re-identification (re-ID), the model identifies the same object across different camera views. This is done through a classification task, where re-ID features are extracted with ResNet from the bounding boxes in each frame. For each bounding box, the model predicts whether it links the object to one of the existing tracks or creates a new track.
\fi

\subsection{Clip Representation Network (CRNet)}
\label{sec:crnet}
In CRNet, it aims to learn the representations of clips (i.e., a set of tracked trajectories). 
%Motivated by the power of Transformer backbone for sequential data~\cite{DBLP:conf/naacl/DevlinCLT19}, 
{\mm We design a trajectory-enhanced Transformer module, by fusing visual features (extracted from images) and spatial features (extracted from trajectories) together in a clip. {\tcsvtminor{Figure~\ref{fig:framework_crnet} illustrates the architecture.}}} %We next introduce the architecture in a bottom-up manner.

\smallskip
\noindent\textbf{Tokenization.} For each clip, we first break it into a sequence of non-overlapping segments with a fixed duration, and each {\cheng is} called a segment.
To tokenize the trajectories in each segment, we then partition the sports field into a grid with equal cell size, and each segment corresponds to a binary matrix, called segment matrix (denoted by $\mathcal{M}$), where we set the cells (corresponding entries in the matrix) to 1 if they are traveled through by the trajectories, and 0 otherwise. {\cheng This} strategy provides a controllable resolution of sports scenes via the cell size. With a smaller one, it has a higher resolution but reduces the robustness against errors of tracked trajectory locations, and vice versa.
Furthermore, we map the segment matrices into tokens (denoted by $\mathcal{T}$) by following~\cite{DBLP:conf/kdd/WangLCJ19}. The core idea is to scan the segment matrices one by one, and for each matrix, we create a new token to represent it if it is dissimilar (measured by the Jaccard index) from those matrices that have been scanned. 

\smallskip
\noindent\textbf{Input Embedding Layer.} We transform the extracted segment matrices into real vectors via an embedding layer. The Transformer~\cite{DBLP:conf/naacl/DevlinCLT19}, which has received great success in modeling sequential data, is used for this purpose. In contrast to recurrent neural networks (RNN) that receive inputs sequentially, the Transformer-based model in CRNet uses a self-attention mechanism that operates on all input tokens in parallel. Thus, it is insensitive to the order of inputs. To preserve the sequential information of trajectories in a clip, we construct the input representations as follows:
\begin{equation}
    \mathbf{x}_i=\mathbf{s}_i+\mathbf{p}_i,
\end{equation}
where $\mathbf{s}_i$ and $\mathbf{p}_i$ denote an input segment embedding and a learnable positional embedding at the $i^{th}$ segment, respectively. The positional embeddings enable the model to be aware of the order of segments rather than treating them as an unordered set.

\smallskip
\noindent\textbf{Multi-head Self-attention.} To model dependencies between the video segments in a clip, we use a multi-head self-attention mechanism~\cite{DBLP:conf/nips/VaswaniSPUJGKP17} based on the input representations $\mathbf{X}=[\mathbf{x}_1, \mathbf{x}_2, ..., \mathbf{x}_m] \in \mathbb{R}^{m\times d_1}$, where $m$ denotes the number of segments, and each is with the dimension $d_1$.
The multi-head self-attention mechanism maps the input representations to output representations $\mathbf{Z}=[\mathbf{z}_1, \mathbf{z}_2, ..., \mathbf{z}_m]\in \mathbb{R}^{m\times d_2}$, where $d_2$ is the dimension of each output segment. The mechanism is formulated as follows:
\begin{align}
    \label{eq:selfatt}
    \mathbf{Z} &= \text{Multi-head}(\mathbf{X, X, X}) = 
    \text{Concat}(O_1, O_2, ..., O_h ) \mathbf{W}^O, \\
    O_i &=\text{Attention}(\mathbf{X}\mathbf{W}^Q_i, \mathbf{X}\mathbf{W}^K_i, \mathbf{X}\mathbf{W}^V_i) \\
   & = \delta(\frac{(\mathbf{X}\mathbf{W}^Q_i) (\mathbf{X}\mathbf{W}^K_i)^\top}{\sqrt{d_{1}/h}})\mathbf{X}\mathbf{W}^V_i, 
\end{align}
where the three positions in the multi-head correspond to query, key, and value. The input representations are projected into $h$ subspaces (called heads), which allow the model to jointly attend to information from the independent head $O_i$ $(1 \le i \le h)$ by concatenating them together, and $\mathbf{W}^O \in \mathbb{R}^{d_{1}\times d_{2}}$ denotes the weight of the multi-head.
Then, self-attention is applied to each head, where $\mathbf{W}^Q_i$, $\mathbf{W}^K_i$, $\mathbf{W}^V_i \in \mathbb{R}^{d_{1}\times d_{1}/h}$ denote the query, key, and value transformations, respectively. The softmax function $\delta(\cdot)$ is applied. 

\smallskip
\noindent\textbf{Position-wise Feed-forward Network.} The output representation $\mathbf{Z}$ is then fed into a fully connected network with two dense layers, i.e., Position-wise Feed-forward Network. It consists of two linear transformations with a ReLU activation function $\Phi(\cdot)$ in between, that is
\begin{align}
\mathbf{t} = \Phi(\mathbf{Z}\mathbf{W}_1+\mathbf{b}_1)\mathbf{W}_2+\mathbf{b}_2,
\end{align}
where $\mathbf{t} \in \mathbb{R}^{d_3}$ denotes the trajectory-based representation, and $\mathbf{W}_1, \mathbf{W}_2, \mathbf{b}_1, \mathbf{b}_2$ are the parameters of the network.

\smallskip
\SH{\noindent\textbf{Fusion of Trajectory and Visual Representations.}
{\cheng Apart from} capturing the sports trajectory information, we also leverage a video representation backbone~\cite{DBLP:conf/eccv/NiPCZMFXL22} to extract visual information as another modality to further boost representation performance. The video representation model can capture visual information across frames and convert frames into a fixed-length vector, denoted as $\mathbf{y} \in \mathbb{R}^{d_4}$. Finally, the derived trajectory-based representation is concatenated with the visual-based representation to represent a sports clip, which is formulated as:
\begin{align}
\mathbf{c} = \text{Concat}(\mathbf{t}, \mathbf{y}),
\end{align}
where $\mathbf{c} \in \mathbb{R}^{d_5}$ ($d_5=d_3+d_4$) denotes the output clip representation.}

\begin{figure}
  \centering
  \hspace{1.5cm}
  \includegraphics[width=0.7\linewidth, trim={25cm 79.2cm 45.5cm 31.6cm}, clip]{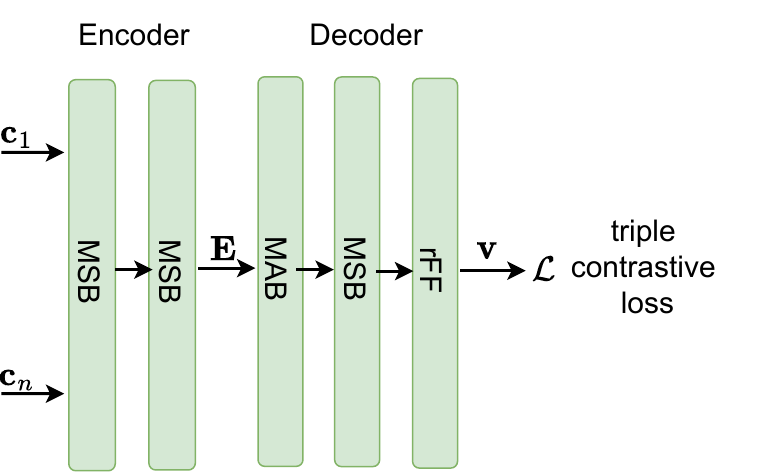}
  \vspace{-3mm}
  \caption{{\mm {\tcsvtminor{The VRNet framework. It encodes the clip representations $\mathbf{c}$ into features $\mathbf{E}$, which are then decoded to produce the final video representation $\mathbf{v}$. Finally, a triple contrastive loss $\mathcal{L}$ is introduced to optimize these representations in an unsupervised manner.}}}}\label{fig:framework_vrnet}
  \vspace*{-5mm}
\end{figure}

\subsection{Video Representation Network (VRNet)}
\label{sec:grnet}
In VRNet, it learns a video representation $\mathbf{v}$ based on the derived clip representations $\mathbf{C}=[\mathbf{c}_1, \mathbf{c}_2, ..., \mathbf{c}_n] \in \mathbb{R}^{n\times d_5}$, where $n$ represents the number of clips contained in the video. To do this, one immediate idea is to aggregate the clip representations as the video representation. However, this design assumes that each clip contributes equally to the final representation, whereas it is more intuitive to assign more weight to those highlighted clips (e.g., a clip that contains a goal). Inspired by this, we design two attention blocks (called MAB and MSB) on those clips in VRNet, and the video representation is outputted via an encoder-decoder-based architecture. {\tcsvtminor{Figure~\ref{fig:framework_vrnet} illustrates the architecture.}}

\smallskip
\noindent\textbf{Encoder.} We first define a Multihead Attention Block (MAB) as a building block in the encoder, as follows:
\begin{align}
    &\text{MAB}(\mathbf{X},\mathbf{Y}) = \text{LayerNorm}(\mathbf{H}+\text{rFF}(\mathbf{H})),\\
    &\mathbf{H} = \text{LayerNorm}(\mathbf{X}+\text{Multi-head}(\mathbf{X}, \mathbf{Y}, \mathbf{Y})),  
\end{align}
where $\mathbf{X}$ and $\mathbf{Y}$ denote two input representations to be pairwise attended by multi-head. $\text{rFF}(\cdot)$ denotes a row-wise feed-forward layer that operates each instance independently and returns the same shape as the input. $\text{LayerNorm}(\cdot)$ denotes a layer normalization operation~\cite{DBLP:journals/corr/BaKH16}. Further, when $\mathbf{X} = \mathbf{Y}$, the MAB reduces to a self-attention building block called Multihead Self-attention Block (MSB), that is
\begin{align}
\text{MSB}(\mathbf{X}) = \text{MAB} (\mathbf{X},\mathbf{X})
\end{align}
Based on MSB, we construct an encoder as follows:
\begin{align}
\mathbf{E} = \text{Encoder}(\mathbf{C}) = \text{MSB}(\text{MSB}(\mathbf{C})).
\end{align}
{\mm The objective of MSB is to transform clip representations $\mathbf{C} \in \mathbb{R}^{n \times d_5}$ into features $\mathbf{E} \in \mathbb{R}^{n \times d_6}$ and capture the correlation of those clips by self-attention in between.}

\smallskip
\noindent\textbf{Decoder.} The decoder further aggregates the features $\mathbf{E}$ into a single video representation $\mathbf{v} \in \mathbb{R}^{d}$, which is formulated as
\begin{align}
    \mathbf{v} = \text{Decoder}(\mathbf{E}) = \text{rFF}(\text{MSB}(\text{MAB}(\mathbf{s}, \mathbf{E}))),
\end{align}
{\mm The objective of MAB is to utilize a trainable seed vector $\mathbf{s} \in \mathbb{R}^{d}$ to reduce the $\mathbf{E}$ into a single representation by attention,} and then feeds into a feed-forward network to output the final video representation with the same shape as $\mathbf{s}$.

{\mm 
\smallskip
\subsection{Optimization}
\label{sec:loss}
To learn the video representations, we present triple contrastive learning. It helps to train the model in an unsupervised manner by introducing comparison. Specifically, let $\mathcal{V}^{(1)}$ denote a batch of sports videos and for each video denoted by $V^{(1)}$ contained in $\mathcal{V}^{(1)}$, it corresponds to a vector representation $\mathbf{v}^{(1)}_i$ ($1 \le i \le n$), where $n$ denotes the batch size. Then, we introduce two variants of $V^{(1)}$ (resp. $\mathcal{V}^{(1)}$, $\mathbf{v}^{(1)}$), denoted by $V^{(2)}$ (resp. $\mathcal{V}^{(2)}$, $\mathbf{v}^{(2)}$) representing intra-clip videos and $V^{(3)}$ (resp. $\mathcal{V}^{(3)}$, $\mathbf{v}^{(3)}$) representing inter-clip videos, as follows. 

\smallskip
\noindent\textbf{Intra-clip Video.} Recall that a clip corresponds to a set of multiple trajectories, where we randomly replace some of the trajectories with other trajectories from the whole training set. Here, we control the replacement with a noise rate denoted by $\delta$ from 0 to 1, e.g., by setting a larger ratio, it tends to replace more trajectories, and thus the variant $V^{(2)}$ will be more dissimilar to the original $V^{(1)}$ in terms of the trajectory patterns within the clips. 

\smallskip
\noindent\textbf{Inter-clip Video.} Each video corresponds to a set of clips after data prepossessing. Similarly, we replace some of the clips with other clips sampling from the training set controlled by $\delta$. In this way, the variant $V^{(3)}$ is dissimilar to the original $V^{(1)}$ in terms of the clip dependencies in the video.

\smallskip
\noindent\textbf{Triple Contrastive Loss.} We further introduce a triple contrastive loss $\mathcal{L}$ based on $V^{(1)}$, $V^{(2)}$, and $V^{(3)}$, where it contains three contrasts: $\mathcal{L}(V^{(1)},V^{(2)})$, $\mathcal{L}(V^{(1)},V^{(3)})$ and $\mathcal{L}(V^{(2)},V^{(3)})$. To calculate $\mathcal{L}(V^{(1)}, V^{(2)})$, we denote $V^{(2)}$ as the positive of $V^{(1)}$, and we consider the negatives as those videos in the same batch with $V^{(1)}$. Here, we note that the training data is randomly shuffled, and the negatives within a batch are generally quite dissimilar to the query. We denote a symmetric InfoNCE loss for $\mathcal{L}(V^{(1)},V^{(2)})$, which is composed of two terms $\mathcal{L}_{1,2}$ and $\mathcal{L}_{2,1}$. For $\mathcal{L}_{1,2}$, it is defined as:

\begin{align}
\label{eq:clLoss}
\tiny
\mathcal{L}_{1,2} = \sum\nolimits_{V_i^{(1)} \in \mathcal{V}^{(1)}} -\log \frac{\exp\Bigr({\mathbf{v}_i^{(1)}\cdot\mathbf{v}_i^{(2)}/\tau}\Bigr)}{ \sum\nolimits_{V_j^{(2)}\in\mathcal{V}^{(2)},j\neq i} \exp\Bigr({\mathbf{v}_i^{(1)}\cdot\mathbf{v}_j^{(2)}/\tau}\Bigr)},
\end{align}
where $\tau$ denotes a temperature parameter in contrastive learning. Symmetrically, we can get $\mathcal{L}_{2,1}$ by anchoring at $V^{(2)}$, then $\mathcal{L}(V^{(1)},V^{(2)})$ is defined as 
\begin{align}
\mathcal{L}(V^{(1)},V^{(2)}) = \mathcal{L}_{1,2} + \mathcal{L}_{2,1}.
\end{align}
Similarly, we can calculate $\mathcal{L}(V^{(1)},V^{(3)})$ and $\mathcal{L}(V^{(2)},V^{(3)})$, and the triple contrastive loss $\mathcal{L}$ is defined as
\begin{align}
\mathcal{L} = &\alpha\mathcal{L}(V^{(1)},V^{(2)}) + \beta\mathcal{L}(V^{(1)},V^{(3)})\\ 
&+ (1-\alpha-\beta)\mathcal{L}(V^{(2)},V^{(3)}), \nonumber
\end{align}
where $0 \le \alpha, \beta \le 1$ are hyperparameters that are used to balance the importance of each contrast.

We discuss the rationale behind the design. First, inspired by the fact that discerning the similarity between two videos can be intricate. However, given a query video and two candidate videos, it is easier to tell the model which one (i.e., positive sample) is more similar than the other one (i.e., negative sample) to the query, and therefore we can optimize the representations by comparing the similarities of videos in this unsupervised manner. %it can be challenging to determine whether two videos are similar or not. 
Second, we enhance the representation by contrasting the videos $V^{(2)}$ and $V^{(3)}$ with the original $V^{(1)}$. The contrasts help the representation to capture both trajectory patterns within the clips and their dependencies in a video. Additionally, contrasting $V^{(2)}$ and $V^{(3)}$ further enhances the representation by maximizing the mutual information derived from the two variants of the same video $V^{(1)}$.
}

\section{Deployment and Applications}
\label{sec:applications}

{\indzheng 
\subsection{Deployment} 
A sports system, built upon TrajSV, is integrated into a practical sports video search engine. The system architecture is illustrated in Figure~\ref{fig:system}. For offline use, the TrajSV model is employed to extract representations from all database videos, forming an Approximate Nearest Neighbor (ANN) index (specifically, HNSW~\cite{DBLP:journals/pami/MalkovY20}). This index facilitates rapid retrieval by computing vector similarities. To optimize memory usage and efficiency, compression and quantization techniques~\cite{DBLP:books/daglib/0015804} are applied, trading off some accuracy. For online use, a query video undergoes TrajSV to derive clip representations from CRNet and a video representation from VRNet. After compression and quantization, the video representation is used for similarity retrieval through the ANN index. By specifying a Top-$K$ parameter, result videos from the database are retrieved and recommended to users. The system is extended to enhance action spotting and video captioning by utilizing clip representations from CRNet. %Below, we introduce three applications: sports video retrieval, action spotting, and video captioning.

\if 0
A sports system based on TrajSV is deployed into a real search engine product for sports video search. Figure~\ref{fig:system} illustrates the system architecture. For offline use, we employ the trained TrajSV model to extract representations from all videos in a database for retrieval. Based on this, we construct an Approximate Nearest Neighbor (ANN) index, i.e., HNSW~\cite{DBLP:journals/pami/MalkovY20}, using the learned representations to enable a fast retrieval process by calculating vector similarities. It further incorporates compression and quantization techniques~\cite{DBLP:books/daglib/0015804} to reduce memory usage and enhance efficiency while sacrificing some accuracy. For online use, we initially process a query video with the TrajSV model to obtain clip representations from CRNet and a video representation from VRNet for the query. After compression and quantization, the video representation is used for similarity retrieval with the ANN index. By setting a Top-$K$ parameter, the result videos from the database are returned and recommended to users. Utilizing the clip representations from CRNet, we extend the system to improve action spotting and video captioning.}
\fi

\subsection{Applications}
We introduce the following three applications developed on top of the system: sports video retrieval, action spotting, and video captioning.
% For sports video retrieval, we describe how the representations work in a retrieval system.
% %
% For action spotting and video captioning, the clip representations will be used to concatenate the embedded features of inputs on any existing models~\cite{DBLP:conf/cvpr/CioppaDGGDGM20,DBLP:conf/cvpr/GiancolaG21,DBLP:journals/corr/abs-2106-14447,Mkhallati2023SoccerNetCaption}, serving as a plugin functionality.

\smallskip
\noindent\textbf{Sports Video Retrieval.} Sports video retrieval is an emerging application used in several sports broadcast platforms, such as ESPN, to recommend videos to sports fans based on their interests. The retrieval process involves the offline stage and online stage as introduced in the deployed system, where it uses the video representation for the retrieval.

\smallskip
\noindent\textbf{Action Spotting.} The action spotting task, as defined in~\cite{DBLP:conf/cvpr/GiancolaADG18}, involves localizing when and which a given action (e.g., goal, shots on target) occurs in a video, with each action annotated with a single timestamp. To support this task, we utilize clip representations as a plug-in built on top of existing spotting models. We achieve this by concatenating the representations with the embedded features of the model, creating a new input. We then unfreeze the representations, allowing them to be further optimized with the model parameters. This is because the clip representations embed spatial features from extracted trajectories that are potentially associated with the actions to be detected. For example, the CALF model~\cite{DBLP:conf/cvpr/CioppaDGGDGM20} leverages ResNet to extract per-frame feature representations. We concatenate the representation of a clip where the frame falls with the embedded features, which creates a new input representation that is then fed into the model for fine-tuning. 
%Through empirical findings, we have found that this approach improves the action spotting baselines and boosts training efficiency.

\begin{figure}
  \centering
  \includegraphics[width=\linewidth]{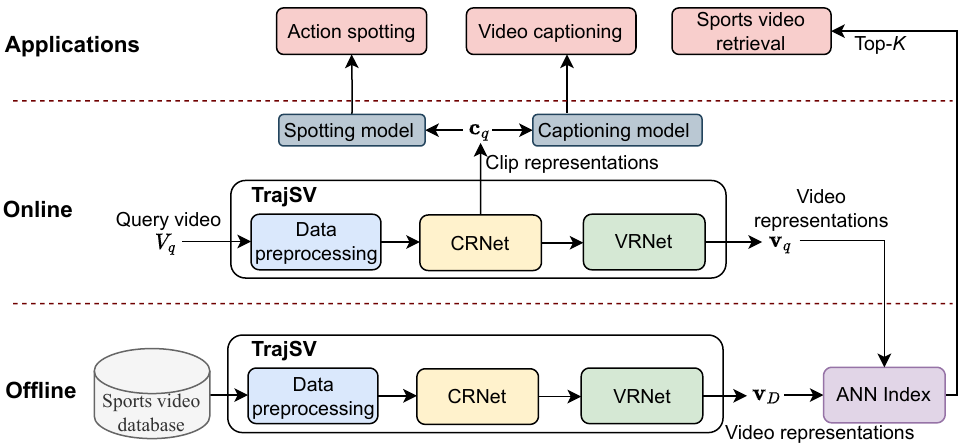}
  \vspace{-6mm}
  \caption{{\indzheng System deployment on TrajSV.}}
  \label{fig:system}
  \vspace*{-6mm}
\end{figure}

\smallskip
\noindent\textbf{Video Captioning.} We consider two video captioning tasks that are introduced in Baidu-VC~\cite{Mkhallati2023SoccerNetCaption}, called Single-anchored Dense Video Captioning (SDVC) and Dense Video Captioning (DVC). For SDVC, the task involves analyzing a video, identifying specific moments that require captions (Commentary Spotting), and generating natural language sentences that describe the events taking place at those moments. To do this, we follow a similar approach in the action spotting task. We concatenate the clip representation with the embedded moment feature, creating a new input representation to fine-tune the baseline models (e.g., Baidu-VC) for video captioning. For DVC, the task involves temporally localizing a caption with the start and end frames instead of anchoring a single moment.
\section{EXPERIMENTS}
\label{sec:experiment}

\subsection{Experimental Setup}
\label{sec:setup}

\begin{table*}[htbp]
\caption{Effectiveness of sports video retrieval (HR@1 and MRR) on YouTube, SoccerNet, and SportsMOT (soccer, basketball, and volleyball). The best results are marked in bold, and the second-best results are underlined\protect\footnotemark.}
\vspace{-3mm}
\small
\setlength{\tabcolsep}{1.7pt}
\resizebox{\linewidth}{!}{%
    \begin{tabular}{c|ccc|ccc|ccc|ccc|ccc|ccc}
    \hline
    \multirow{2}[6]{*}{Method} & \multicolumn{6}{c|}{YouTube} & \multicolumn{6}{c|}{SoccerNet} & \multicolumn{6}{c}{SportsMOT} \\
\cline{2-19}    \multicolumn{1}{c|}{} & \multicolumn{3}{c|}{HR@1} & \multicolumn{3}{c|}{MRR} & \multicolumn{3}{c|}{HR@1} & \multicolumn{3}{c|}{MRR} & \multicolumn{3}{c|}{HR@1} & \multicolumn{3}{c}{MRR} \\
\cline{2-19}    \multicolumn{1}{c|}{} & 0.5 & 0.55 & 0.6 & 0.5 & 0.55 & 0.6 & 0.5 & 0.55 & 0.6 & 0.5 & 0.55 & 0.6 & 0.5 & 0.55 & 0.6 & 0.5 & 0.55 & 0.6 \\
    \hline
    play2vec(Mean)~\cite{DBLP:conf/kdd/WangLCJ19} & 0.069 & 0.075 & 0.031 & 0.105 & 0.111 & 0.058 & 0.078 & 0.031 & 0.058 & 0.133 & 0.090 & 0.052 & 0.097 & 0.106 & 0.057 & 0.071 & 0.063 & 0.031 \\
    play2vec(MLP)~\cite{DBLP:conf/kdd/WangLCJ19} & 0.223 & 0.205 & 0.088 & 0.331 & 0.299 & 0.156 & 0.312 & 0.203 & 0.156 & 0.443 & 0.322 & 0.221 & 0.328 & 0.301 & 0.154 & 0.210 & 0.232 & 0.084 \\
    Chalkboard~\cite{DBLP:conf/iui/ShaLYCRM16} & 0.077 & 0.000 & 0.000 & 0.109 & 0.019 & 0.010 & 0.300 & 0.280 & 0.140 & 0.372 & 0.357 & 0.242 & 0.326 & 0.218 & 0.224 & 0.167 & 0.250 & {\ul 0.194} \\
    SimScene (Mean)~\cite{DBLP:conf/mmasia/HaruyamaTOH20} & 0.129 & 0.141 & 0.038 & 0.199 & 0.209 & 0.087 & 0.180 & 0.080 & 0.080 & 0.285 & 0.138 & 0.168 & 0.276 & 0.254 & 0.121 & 0.170 & 0.191 & 0.061 \\
    SimScene (MLP)~\cite{DBLP:conf/mmasia/HaruyamaTOH20} & 0.191 & 0.170 & 0.061 & 0.276 & 0.254 & 0.121 & 0.120 & 0.060 & 0.040 & 0.219 & 0.102 & 0.098 & 0.007 & 0.011 & 0.004 & 0.003 & 0.007 & 0.001 \\
    ResNet (Mean)~\cite{he2016deep} & 0.347 & 0.364 & 0.130 & 0.452 & 0.467 & 0.238 & 0.438 & 0.156 & 0.078 & 0.571 & 0.328 & 0.222 & 0.454 & 0.379 & 0.099 & 0.250 & 0.281 & 0.031 \\
    ResNet (MLP)~\cite{he2016deep} & {\ul 0.659} & {\ul 0.658} & {\ul 0.231} & {\ul 0.779} & {\ul 0.779} & {\ul 0.443} & 0.609 & 0.281 & 0.109 & {\ul 0.762} & 0.500 & 0.325 & {\ul 0.711} & 0.672 & {\ul 0.347} & 0.438 & {\ul 0.531} & 0.125 \\
    X-CLIP (Mean)~\cite{DBLP:conf/eccv/NiPCZMFXL22} & 0.358 & 0.380 & 0.119 & 0.506 & 0.521 & 0.237 & 0.562 & 0.234 & 0.094 & 0.699 & 0.431 & 0.302 & 0.665 & {\ul 0.680} & 0.282 & {\ul 0.469} & 0.500 & 0.156 \\
    X-CLIP (MLP)~\cite{DBLP:conf/eccv/NiPCZMFXL22} & 0.611 & 0.611 & 0.216 & 0.768 & 0.764 & 0.426 & {\ul 0.641} & {\ul 0.312} & {\ul 0.141} & {\ul 0.762} & {\ul 0.538} & {\ul 0.332} & 0.604 & 0.527 & 0.117 & 0.281 & 0.406 & 0.062 \\
    TrajSV & \textbf{0.791} & \textbf{0.802} & \textbf{0.475} & \textbf{0.881} & \textbf{0.887} & \textbf{0.680} & \textbf{0.719} & \textbf{0.484} & \textbf{0.250} & \textbf{0.846} & \textbf{0.689} & \textbf{0.551} & \textbf{0.748} & \textbf{0.810} & \textbf{0.614} & \textbf{0.712} & \textbf{0.650} & \textbf{0.475} \\
    \hline
    \end{tabular}%
  }
\label{tab:retrieval_all_datasets}%
\vspace*{-4mm}
\end{table*}%

\footnotetext{All results are statistically significant (t-test with $p<0.05$).\label{sigtest}}

\newcolumntype{P}[1]{>{\centering\arraybackslash}p{#1}}
\newcolumntype{M}[1]{>{\centering\arraybackslash}c{#1}}
\begin{table*}[]
\caption{Effectiveness of action spotting (Avg-mAP\%) on SoccerNet\textsuperscript{\ref{sigtest}}.}%, where the actions are (1) Ball out, (2) Throw-in, (3) Foul, (4) Ind free-kick, (5) Clearance, (6) Shots on tar, (7) Shots off tar, (8) Corner, (9) Substitution, (10) Kick-off, (11) Yellow card, (12) Offside, (13) Dir free-kick, (14) Goal, (15) Penalty, (16) Yel $\rightarrow$Red, (17) Red card.}
\vspace{-3mm}
\small
\setlength{\tabcolsep}{2pt}
\resizebox{\textwidth}{!}{%
    \begin{tabular}{M|c|ccccccccccccccccc}
    \hline
       \multirow{-3}[2]{*}{Method} & \begin{sideways}\footnotesize{Overall}\end{sideways} & \begin{sideways}\footnotesize{Ball out}\end{sideways} & \begin{sideways}\footnotesize{Throw-in}\end{sideways} & \begin{sideways}\footnotesize{Foul}\end{sideways} & \begin{sideways}\footnotesize{Ind. free-kick}\end{sideways} & \begin{sideways}\footnotesize{Clearance}\end{sideways} & \begin{sideways}\footnotesize{Shots on tar.}\end{sideways} & \begin{sideways}\footnotesize{Shots off tar.}\end{sideways} & \begin{sideways}\footnotesize{Corner}\end{sideways} & \begin{sideways}\footnotesize{Substitution}\end{sideways} & \begin{sideways}\footnotesize{Kick-off}\end{sideways} & \begin{sideways}\footnotesize{Yellow card}\end{sideways} & \begin{sideways}\footnotesize{Offside}\end{sideways} & \begin{sideways}\footnotesize{Dir. free-kick}\end{sideways} & \begin{sideways}\footnotesize{Goal}\end{sideways} & \begin{sideways}\footnotesize{Penalty}\end{sideways} & \begin{sideways}\footnotesize{Yel. → Red}\end{sideways} & \begin{sideways}\footnotesize{Red card}\end{sideways} \\
    \hline
   \# of instances & 22551 & 6460 & 3809 & 2414 & 2283 & 1631 & 1175 & 1058 & 999 & 579 & 514 & 431 & 416 & 382 & 337 & 41 & 14 & 8 \\ \hline
    CALF~\cite{DBLP:conf/cvpr/CioppaDGGDGM20} & 40.7 & 63.9 & 56.4 & 53.0 & 41.5 & 51.6 & 26.6 & 27.3 & 71.8 & 47.3 & 37.2 & 41.7 & 25.7 & 43.5 & 72.2 & 30.6 & 0.7 & 0.7 \\
    NetVLAD++~\cite{DBLP:conf/cvpr/GiancolaG21} & 53.4 & 70.3 & 69.0 & 64.2 & 44.4 & 57.0 & 39.3 & 41.0 & 79.7 & 68.7 & 62.1 & 56.7 & 39.3 & 57.8 & 71.6 & 79.3 & 3.7 & 4.0 \\
    Baidu-AS~\cite{DBLP:journals/corr/abs-2106-14447} & 73.2 & {\ul 87.5} & 87.5 & {\ul 84.1} & 70.9 & {\ul 76.7} & \textbf{66.3} & \textbf{66.7} & 89.0 & 76.3 & {\ul 82.7} & \textbf{77.9} & {\ul 68.8} & 74.9 & 87.3 & {\ul 82.7} & 39.8 & \textbf{25.9} \\
    \hline
    Baidu-AS+play2vec & {\ul 73.5} & 87.4 & \textbf{87.7} & 84.0 & {\ul 71.2} & 76.6 & {\ul 65.9} & 66.3 & {\ul 89.0} & {\ul 78.4} & 82.4 & 77.3 & \textbf{69.0} & \textbf{75.2} & {\ul 87.5} & 82.4 & \textbf{44.5} & 24.1 \\
    Baidu-AS+TrajSV & \textbf{73.7} & \textbf{87.6} & {\ul 87.6} & \textbf{84.2} & \textbf{71.8} & \textbf{76.7} & 65.3 & {\ul 66.3} & \textbf{89.0} & \textbf{79.4} & \textbf{82.8} & {\ul 77.6} & 68.5 & {\ul 75.1} & \textbf{87.6} & \textbf{82.8} & {\ul 44.2} & {\ul 25.8} \\
    \hline
\end{tabular}
}
\label{tab:spotting}
\vspace*{-4mm}
\end{table*}

\noindent\textbf{Datasets and {\indzheng Ground Truth}.} We conduct experiments on three broadcast sports video datasets, i.e., YouTube, SoccerNet~\cite{DBLP:conf/cvpr/DeliegeCGSDNGMD21}, and SportsMOT (it is employed for retrieval and qualitative experiments, lacking labels necessary for evaluating action spotting and video captioning tasks)~\cite{cui2023sportsmot}. For YouTube, we crawl 3,261 soccer videos from YouTube sports channels with durations varying from 13 to 962 seconds. The SoccerNet dataset contains 550 complete broadcast soccer games from the major European leagues and provides annotations for action spotting in 17 common classes (e.g., goal, offside, shots on target). Additionally, the dataset contains an average of 78.33 comments per game that are temporally localized, corresponding to a total of 36,894 captions for the entire dataset. {\mm For SportsMOT, it consists of 240 videos across three types of sports (soccer, basketball, and volleyball), where the videos are collected from some main sports events like the Olympic Games, NCAA Championship, and NBA.}
To track trajectories, we set the sports field coordinates to a range of $[-52.5, +52.5]$ meters along the x-axis and $[-34, +34]$ meters along the y-axis, following the setting described in~\cite{DBLP:conf/kdd/WangLCJ19}, the setting is also used for non-soccer data (i.e., basketball and volleyball).

{\indzheng We then discuss the ground truth for evaluation. For (1) sports video retrieval, we introduce variations in the original videos contained in the datasets by manipulating them with different noise rates $\delta$. This manipulation involves replacing both trajectories and clips in the videos. Subsequently, we use the manipulated videos as queries and aim to retrieve their corresponding original versions (i.e., ground truth) from the database. For (2) action spotting and (3) video captioning, we take the common classes and the comments provided by the SoccerNet dataset as the ground truth, respectively.}

\smallskip
\noindent\textbf{Baselines.} 
\if 0
{\Comment We compare TrajSV with the baseline methods in terms of 
%
\underline{sports video retrieval} (i.e., play2vec~\cite{DBLP:conf/kdd/WangLCJ19}, Chalkboard~\cite{DBLP:conf/iui/ShaLYCRM16}, SimScene~\cite{DBLP:conf/mmasia/HaruyamaTOH20}, ResNet~\cite{he2016deep}, X-CLIP~\cite{DBLP:conf/eccv/NiPCZMFXL22}), 
%
\underline{action spotting} (i.e., CALF~\cite{DBLP:conf/cvpr/CioppaDGGDGM20}, NetVLAD++~\cite{DBLP:conf/cvpr/GiancolaG21}, Baidu-AS~\cite{DBLP:journals/corr/abs-2106-14447}), 
%
and \underline{video captioning} (i.e., Baidu-VC~\cite{Mkhallati2023SoccerNetCaption}). A detailed description of the baselines is included in the supplementary material due to the page limit.}
\fi 
% {\indzheng We carefully examine the literature including a recent challenge paper~\cite{cioppa2023soccernet}, and identify the following baselines in terms of sports video retrieval, action spotting, and video captioning. Notably, these models are open-sourced, and we tune their embedding dimensions to the best on a development set as shown in Table~\ref{tab:baseline} for the comparison.
% }
{\indshihao We carefully examine the literature including a recent challenge paper~\cite{cioppa2023soccernet}, and identify recent baselines in terms of (1) sports video retrieval, i.e., play2vec~\cite{DBLP:conf/kdd/WangLCJ19}, Chalkboard~\cite{DBLP:conf/iui/ShaLYCRM16}, SimScene~\cite{DBLP:conf/mmasia/HaruyamaTOH20}, ResNet~\cite{he2016deep}, X-CLIP~\cite{DBLP:conf/eccv/NiPCZMFXL22}, (2) action spotting, i.e., CALF~\cite{DBLP:conf/cvpr/CioppaDGGDGM20}, NetVLAD++~\cite{DBLP:conf/cvpr/GiancolaG21}, Baidu-AS~\cite{DBLP:journals/corr/abs-2106-14447}, and (3) video captioning, i.e., Baidu-VC~\cite{Mkhallati2023SoccerNetCaption}. The descriptions of these methods can be found in Section~\ref{sec:related}. Notably, these models are open-sourced, and we tune their parameters to the best for the comparison.
}

\smallskip
\noindent\textbf{Implementation Details.} We conduct the experiments using Python 3.7 and PyTorch 1.8.0. {\indzheng We describe the implementation details with open-sourced codes below.}
%on a single Nvidia V100 32G GPU 

{\tcsvt{
(1) For data preprocessing, in video segmentation, a 1D CNN with three layers and a 21-frame kernel is trained for binary classification to identify the main-camera view. The model extracts ResNet 512-dimensional PCA-reduced features\footnote{{\indshihao https://github.com/SoccerNet/sn-spotting}}, aiding in segmenting long videos into shorter clips. We adopt morphological opening and closure techniques~\cite{soille2013morphological} with a kernel size of three to enhance binary prediction quality. In camera calibration, following TVCalib~\cite{DBLP:conf/wacv/TheinerE23}\footnote{{\indshihao https://github.com/mm4spa/tvcalib}}, a segment localization model is trained for instance segmentation (e.g., lines and circle segments in a soccer field). {\tcsvtminor{TVCalib predicts camera and distortion parameters by minimizing the reprojection loss. Following the original hyperparameter setup, we use AdamW~\cite{loshchilov2017decoupled} with a learning rate of 0.05 and weight decay of 0.01 to train the camera parameters over 2000 fixed steps, employing the one-cycle learning rate schedule~\cite{smith2019super} with $pct_{start} = 0.5$.}} 
Batch size and downsample Frames per Second (FPS) are set to 512 and 5, respectively. {\tcsvtminor{In multi-object tracking, we fine-tune the FairMOT model~\cite{DBLP:journals/ijcv/ZhangWWZL21}\footnote{{\indshihao https://github.com/ifzhang/FairMOT}} on the SoccerNet-Tracking dataset~\cite{DBLP:conf/cvpr/CioppaGDKZCGD22} for 40 epochs. The model uses a variant of DLA-34~\cite{zhou2019objects} as the backbone. Training is conducted using the Adam optimizer~\cite{kingma2014adam} with an initial learning rate of $10^{-4}$, which decays to $10^{-5}$ after 20 epochs. We use a batch size of 12.}} 
A weight parameter of 10 is assigned to small objects (bounding box size $< 500$ pixels) to improve soccer ball tracking. The fine-tuned FairMOT model is then used to track players and balls in all main-camera video clips, maintaining the original sampling frequency by default. After this, we transform the tracking result from the image coordinate into the real-world coordinate (sports field) by using the result of the camera calibration and MOT. Meanwhile, we filter out the objects that are tracked outside the sports field.}}

(2) For CRNet, following~\cite{DBLP:conf/kdd/WangLCJ19}\footnote{\indshihao https://github.com/zhengwang125/play2vec}, we segment the sports field with a three-meter cell size and a one-second time segment length to tokenize trajectories. We set the number of video segments in a clip to 16 with zero padding ($m = 16$), and randomly choose 16 consecutive clips for each video ($n = 16$). %For SoccerNet, where videos are longer, $n$ is set to 64. If the number of clips is less than $n$, repeat padding is applied to reach $n$ clips. 
The segment matrices are then tokenized with a Jaccard threshold of 0.3. The dimensions of the embedding layer, segment embedding, and transformer encoder are all set to 128 ($d_1 = d_2 = d_3 = 128$). The dropout value in the positional encoding layer is 0.3, and the transformer encoder employs two layers and two heads. For visual representation, X-CLIP~\cite{DBLP:conf/eccv/NiPCZMFXL22} is used as the backbone (pre-trained on Kinetics-600 dataset)\footnote{\indshihao https://github.com/microsoft/VideoX/tree/master/X-CLIP}, representing each video clip as a 512-dimensional vector ($d_4 = 512$). Visual and trajectory vectors are concatenated into a single blend vector, representing the video clip with a dimension of 640 ($d_5 = d_3 + d_4 = 640$). 

(3) For VRNet, the encoder has two MSB layers that transform the input dimension from \SH{640 to 1280}, and the decoder subsequently reduces the encoded embedding to the output dimension \SH{(${d_6}=128$)}. The attention model employs two heads.

(4) For training, the dataset is divided into 80\% for training and 20\% for testing. The model is trained for 100 epochs using the triple contrastive loss. A noise rate, randomly sampled between 0 and 0.2, generates intra-clip and inter-clip videos. {\tcsvtminor{The rationale is twofold. 1) We adopt random sampling within a controlled range to generate diverse intra-clip and inter-clip variants, enriching the training signal for contrastive learning. 2) Empirical tuning shows that setting $\delta > 0.2$ introduces excessive noise, likely disrupting trajectory or clip coherence and degrading the model’s ability to learn consistent representations. The chosen range thus achieves a favorable trade-off between sample diversity and training stability.
}}
\SH{\mm The hyperparameters for $\alpha$ and $\beta$ in the loss function are set to 0.5 and 0.3, respectively.} The early stopping with a patience step is set to 10. The learning rate is 0.01, and stochastic gradient descent (SGD) with a momentum factor of 0.7 serves as the optimizer. The temperature parameter is adjusted to 0.1.

\if 0
For training, the dataset is split into 80\% for training and 20\% for testing. We train the model for 100 epochs with the triple contrastive loss, where we randomly sample a noise rate ranging from 0 to 0.2 to generate the intra-clip videos and inter-clip videos. \SH{\mm By default, the hyperparameter $\alpha$ and $\beta$ in the loss function are set to 0.5 and 0.3, respectively.} The early stopping with a patience step is set to 10, and the batch sizes are set to 128 for both YouTube and SoccerNet, and 32 for SportsMOT. The learning rate is set to 0.01. We utilize stochastic gradient descent (SGD) with a momentum factor of 0.7 as the optimizer. The temperature parameter is adjusted to 0.1. %{\indshihao On average, the training time is around 25 minutes.}%\SH{\mm When generating a noisy version of the original video for contrastive learning, we randomly select a replacement ratio between 0 and 0.2 to replace the trajectory and clip.} the preprocessing time is about five times the video duration, and 
\fi

\if 0
{\Comment We conduct the experiments on a single Nvidia V100 32G GPU using Python 3.7 and PyTorch 1.8.0. 
%
The implementation details of the model and training process can be found in the supplementary material due to the page limit.
}
\fi

\smallskip
\noindent\textbf{Evaluation Metrics.} 
We evaluate the effectiveness of TrajSV in terms of different tasks. (1) For sports video retrieval, Hitting Ratio for Top-1 (denoted by HR@1) and Mean Reciprocal Rank (denoted by MRR) are used by following~\cite{DBLP:conf/mm/MaXSYZJ22,DBLP:conf/kdd/LiuZC18}. (2) For action spotting, Avg-mAP ($\delta$ is varied from 5 to 60 seconds) is used by following~\cite{DBLP:conf/cvpr/CioppaDGGDGM20,DBLP:conf/cvpr/GiancolaG21,DBLP:conf/icpr/TomeiBCBC20,DBLP:conf/cvpr/GiancolaADG18,DBLP:conf/cvpr/DeliegeCGSDNGMD21}. (3) For video captioning, METEOR, BLEU, ROUGE, CIDEr, and SODA\_c are used by following~\cite{Mkhallati2023SoccerNetCaption}. Overall, a higher evaluation metric (i.e., HR@1, MRR, Avg-mAP, METEOR, BLEU, ROUGE, CIDEr, SODA\_c) indicates a better result.

\subsection{Experimental Results}
\label{sec:results}

\begin{table*}[t]
\centering
\caption{Effectiveness of video captioning (BLEU, METEOR, ROUGE, CIDEr, SODA\_c) on SoccerNet\textsuperscript{\ref{sigtest}}.} 
\vspace{-3mm}
\resizebox{\textwidth}{!}{%
\begin{tabular}{c|ccc|cccc|ccccc}
\hline
\multirow{2}[2]{*}{Method} & \multicolumn{3}{c|}{\makecell{Commentary Spotting\\(mAP@ (\%))}} & \multicolumn{4}{c|}{Dense Video Captioning (DVC)} & \multicolumn{5}{c}{Single-anchored Dense Video Captioning (SDVC)} \\ \cline{2-13}  
 & 5 & 30 & 60 & B@4 & M & R & C & B@4@30 & M@30 & R@30 & C@30 & SODA\_C \\ \hline
Baidu-VC~\cite{Mkhallati2023SoccerNetCaption} & {\ul 5.27} & {\ul 49.40} & {\ul 63.10} & {\ul 6.62} & {\ul 23.84} & {\ul 24.53} & {\ul 21.45} & \textbf{21.63} & \textbf{29.44} & 22.09 & 27.20 & 7.79 \\
Baidu-VC+Play2Vec & \textbf{6.07} & 47.70 & 60.97 & 6.48 & 23.83 & 24.26 & 20.58 & {\ul 20.94} & 21.93 & {\ul 26.40} & {\ul 27.11} & {\ul 7.73} \\
Baidu-VC+TrajSV & 4.00 & \textbf{53.07} & \textbf{66.64} & \textbf{6.78} & \textbf{24.26} & \textbf{24.80} & \textbf{22.34} & 20.35 & {\ul 21.97} & \textbf{26.61} & \textbf{27.38} & \textbf{7.90} \\ \hline
\end{tabular}%
}
\label{tab:caption}%
\vspace*{-4mm}
\end{table*}%

\begin{table}[t]
\centering
\caption{Ablation study on YouTube\textsuperscript{\ref{sigtest}}.}
\vspace{-3mm}
\begin{tabular}{lll}
\hline
Components       & HR@1 & MRR \\ \hline
CRNet + VRNet  &   \textbf{0.475}  &  \textbf{0.680}   \\ \hline
w/o CRNet (BiLSTM)  &  0.419     &   0.641  \\
w/o CRNet (LSTM) &  0.384     &   0.625  \\
w/o VRNet (Mean) &   0.239    &   0.147  \\ 
w/o VRNet (MLP)  &   0.335    &  0.197   \\ \hline
Trajectory + VRNet &    0.344  &   0.444  \\ 
Visual (X-CLIP) + VRNet  &   0.242    &   0.461  \\
Visual (ResNet) + VRNet  &   0.386    &   0.603  \\ \hline
w/o $\mathcal{V}^{(1)}$ in loss  &  0.448     &   0.662  \\
w/o $\mathcal{V}^{(2)}$ in loss  &  0.433     &   0.646  \\
w/o $\mathcal{V}^{(3)}$ in loss  &  0.441     &   0.637  \\ \hline
\end{tabular}
\label{tab:ablation}
\vspace*{-3mm}
\end{table}

\begin{table}[t]
\small
  \centering
  \caption{{\indshihao Parameter study of batch size on YouTube.}}
  \vspace{-3mm}
    \begin{tabular}{c|c|ccc}
    \hline
    Parameter & Value & Training time (min) & HR@1 & MRR \\
    \hline
    \multirow{4}[2]{*}{Batch size} & 32 & 60 & 0.442 & 0.647 \\
      & 64 & 42 & 0.440 & 0.646 \\
      & 96 & 36 & 0.456 & 0.658 \\
      & 128 & 25 & \textbf{0.475} & \textbf{0.680} \\
    \hline
    \end{tabular}%
    \vspace{-4mm}
  \label{tab:paramter_study_batch_size}%
\end{table}%

\begin{table}[t]
\small
\vspace{-2mm}
  \centering
  \caption{{\indshihao Parameter study of cell size on YouTube.}}
  \vspace{-3mm}
    \begin{tabular}{c|c|cc}
    \hline
    Parameter & Value & HR@1 & MRR \\
    \hline
    \multirow{5}[2]{*}{Cell size} & 1 & 0.469 & 0.678 \\
      & 3 & \textbf{0.475} & \textbf{0.680} \\
      & 5 & 0.467 & 0.673 \\
      & 7 & 0.458 & 0.662 \\
      & 9 & 0.433 & 0.653 \\
    \hline
    \end{tabular}%
    \vspace{-3mm}
  \label{tab:paramter_study_cell_size}%
\end{table}%

\begin{table}[t]
\small
  \centering
  \caption{Parameter study of embedding dimension on YouTube.}
  \vspace{-3mm}
    \begin{tabular}{c|c|cc}
    \hline
    Parameter & Value & HR@1 & MRR \\
    \hline
    \multirow{4}[2]{*}{Embedding dimension} & 32 & 0.436 & 0.665 \\
      & 64 & 0.464 & 0.682 \\
      & 128 & 0.497 & 0.697 \\
      & 256 & \textbf{0.564} & \textbf{0.738} \\
    \hline
    \end{tabular}%
    \vspace{-3mm}
  \label{tab:paramter_embedding}%
\end{table}%

\begin{table}[t]
\small
\centering
  \caption{Transferability on YouTube (YT) and SoccerNet (SN).}
\vspace{-3mm}
\resizebox{\linewidth}{!}{%
\setlength{\tabcolsep}{2.5pt}
    \begin{tabular}{cccccccc}
    \hline
    \multirow{2}[2]{*}{Method} & \multirow{2}[2]{*}{Train $\rightarrow$ Test} & \multirow{2}[2]{*}{Mode} & \multicolumn{5}{c}{HR@1} \\
\cline{4-8}      &   &   & 0.4 & 0.45 & 0.5 & 0.55 & 0.6 \\
    \hline
    \multirow{2}[2]{*}{ResNet (MLP)} & SN $\rightarrow$ YT & zero-shot & 0.981 & 0.938 & 0.659 & 0.658 & 0.231 \\
      & SN $\rightarrow$ YT & fine-tuning & 0.953 & 0.889 & 0.672 & 0.670 & 0.377 \\
    \hline
    \multirow{3}[2]{*}{TrajSV} & SN $\rightarrow$ YT & zero-shot & 0.956 & 0.886 & 0.591 & 0.248 & 0.139 \\
      & SN $\rightarrow$ YT & fine-tuning & 0.980 & 0.931 & 0.719 & 0.723 & 0.386 \\
      & YT $\rightarrow$ YT & training & \textbf{0.984} & \textbf{0.956} & \textbf{0.791} & \textbf{0.802} & \textbf{0.475} \\
    \hline
    \end{tabular}%
}
\label{tab:Transferability}%
\vspace{-4mm}
\end{table}%

\begin{table}[t]
\caption{{\tcsvt{Retrieval scalability on YouTube.}}}
\centering
\vspace{-2mm}
\label{tab:scalability}
\begin{tabular}{l|cccccc}
\hline
Database size       &500 &1,000 &1,500 &2,000 &2,500 &3,000 \\ \hline
Retrieval time (s)  &10.70 &11.87 &12.58 &13.03 &13.44 &13.75 \\ \hline
\end{tabular}
\vspace{-5mm}
\end{table}

\noindent\textbf{(1) Sports Video Retrieval.} 
%{\mm We conduct the sports video retrieval task on three datasets: YouTube, SoccerNet, and SportsMOT. To evaluate the performance, we introduce variations in the original videos by manipulating them with different noise rates $\delta$ ranging from 0.4 to 0.6. This manipulation involves replacing both trajectories and clips in the videos. Subsequently, we use the manipulated videos as queries and aim to retrieve their corresponding original versions from the database.}
%
To evaluate performance, we introduce variations ($V_{+}$) in original videos ($V$) by applying noise rates from 0.5 to 0.6. If $V_{+}$ is the top-1 retrieved video, HR@1=1; otherwise, HR@1=0. MRR is defined as MRR $= \frac{1}{rank}$, where the $rank$ denotes the rank of $V_{+}$ in the database. The average results of HR@1 and MRR are reported in \cref{tab:retrieval_all_datasets}. We observe that our TrajSV outperforms the best baseline methods significantly. For example, on YouTube (resp. SoccerNet and SportsMOT) with a noise rate of 0.6, TrajSV outperforms ResNet (MLP) (resp. X-CLIP (MLP) and ResNet (MLP)) by 105.6\% (resp. 77.3\% and 76.9\%), because the fusion of visual and trajectory information enhances retrieval performance, surpassing single-modality approaches. TrajSV shows a similar trend on SportsMOT (covering soccer, basketball, and volleyball), highlighting its effectiveness in handling diverse sports videos.

\if 0
We observe that our proposed TrajSV architecture exhibits significant improvements compared to the baseline methods. Specifically, on the SoccerNet dataset, the Mean Reciprocal Rank (MRR) and Hit Ratio at 1 (HR@1) improve by 21.9\% and 11.1\%, respectively, compared to using X-CLIP~\cite{DBLP:conf/eccv/NiPCZMFXL22}, when the noise rate is 0.6. Moreover, on the YouTube dataset, the MRR and HR@1 values improve by 25.4\% and 28.1\%, respectively. The fusion of visual and trajectory information contributes to the enhanced retrieval performance, outperforming single-modality approaches.
%
Furthermore, for all baseline methods, the contrastive learning approach outperforms directly averaging the representations with mean pooling. Contrastive learning's ability to minimize distances between similar representations and maximize distances between different representations improves retrieval performance.
%
When the noise rate increases to 0.5 and above, trajectory representations show more significant improvements compared to using visual cues alone. This indicates that trajectory-based video representations have inherent noise-resistant performance, particularly when sports videos' visual features in the background scenes are highly similar.
%
{\mm In addition, a similar trend is demonstrated on the SportsMOT dataset, which encompasses sports including soccer, basketball, and volleyball. The results on this dataset reflect the effectiveness of the TrajSV in handling diverse sports videos.}
\fi

\smallskip
\noindent\textbf{(2) Action Spotting.}
As shown in \cref{tab:spotting}, we incorporate our clip-level representations with the Baidu-AS embeddings to enhance action spotting on SoccerNet. We observe an overall improvement in action spotting Avg-mAP by 0.5\%, showcasing the effectiveness of including self-trained embeddings in identifying sports actions. Compared to Baidu-AS and Baidu-AS+play2vec (merging Baidu-AS embeddings with play2vec embeddings), our approach achieves state-of-the-art results in 9 out of 17 action categories. Notable improvements include 11.1\% for yellow to red cards, 4.1\% for substitution, and 1.3\% for indirect free-kicks. {\tcsvt{We note that although the improvement over Baidu-AS+Play2Vec is small, it is still appreciated, as Baidu-AS has already reached the dataset’s bottleneck performance by effectively capturing action spotting with vision-based features. The extra gain from incorporating trajectory features (e.g., TrajSV) suggests that vision cues dominate the task, but the added trajectory information remains valuable for refining performance.
}}

\smallskip
\noindent\textbf{(3) Video Captioning.}
Following~\cite{Mkhallati2023SoccerNetCaption}, we evaluate video captioning using our clip-level representations combined with the Baidu-VC encoder on SoccerNet, and the results are reported in \cref{tab:caption}. TrajSV achieves state-of-the-art performance on most metrics, with significant improvements of 7.4\% and 5.6\% in mAP@30 and mAP@60 for the commentary spotting task, respectively. The video captioning performance also shows relative improvements up to 20.5\% (R@30) on the DVC and SDVC tasks. Additionally, our results outperform the implementation that concatenates Baidu-VC embeddings with play2vec embeddings. These findings indicate that incorporating trajectory-informative representations enhances performance for commentary spotting and SDVC.

\smallskip
\noindent\textbf{(4) Ablation Study.} 
To evaluate the effectiveness of different components in TrajSV, we conduct an ablation study for sports video retrieval on YouTube as shown in \cref{tab:ablation}, where we (1) replace the CRNet with LSTM and BiLSTM layers, and the VRNet with mean pooling and a two-layer MLP; (2) test the performance of utilizing different features, including trajectory-based, X-CLIP, and ResNet features; (3) evaluate the effect of the triple contrastive loss. For (1), it demonstrates that both the CRNet and VRNet components contribute to improving the performance. When the CRNet (resp. VRNet) is replaced with BiLSTM and LSTM (resp. Mean and MLP), the HR@1 decreases by 11.8\% and 19.2\% (resp. 49.7\% and 29.5\%), respectively. For (2), our observations reveal that the integration of both trajectory and visual representations yields superior results compared to using either representation individually. {\tcsvtminor{Specifically, excluding trajectory features leads to a 49.1\% drop in HR@1 with X-CLIP features and 18.7\% with ResNet features, highlighting the importance of trajectory-based representations in sports video retrieval. When combined with visual features, the trajectory-visual fusion further enhances fine-grained visual information across frames, leading to a 27.6\% improvement in HR@1.
}}
For (3), we compare the performance when excluding each of the three components in the loss calculation, which shows that using triple contrastive loss is better than using either of them individually. % ($\mathcal{V}^{(1)}$, $\mathcal{V}^{(2)}$, and $\mathcal{V}^{(3)}$) 

\begin{figure*}[t]
    \centering
    \small
    \setlength{\tabcolsep}{1pt}
    \begin{tabular}{m{0.1\textwidth}|m{0.15\textwidth}m{0.15\textwidth}m{0.15\textwidth}m{0.15\textwidth}m{0.15\textwidth}}
        \hline
    
        \makecell{Query\\(Corner)} & \includegraphics[width=0.15\textwidth]{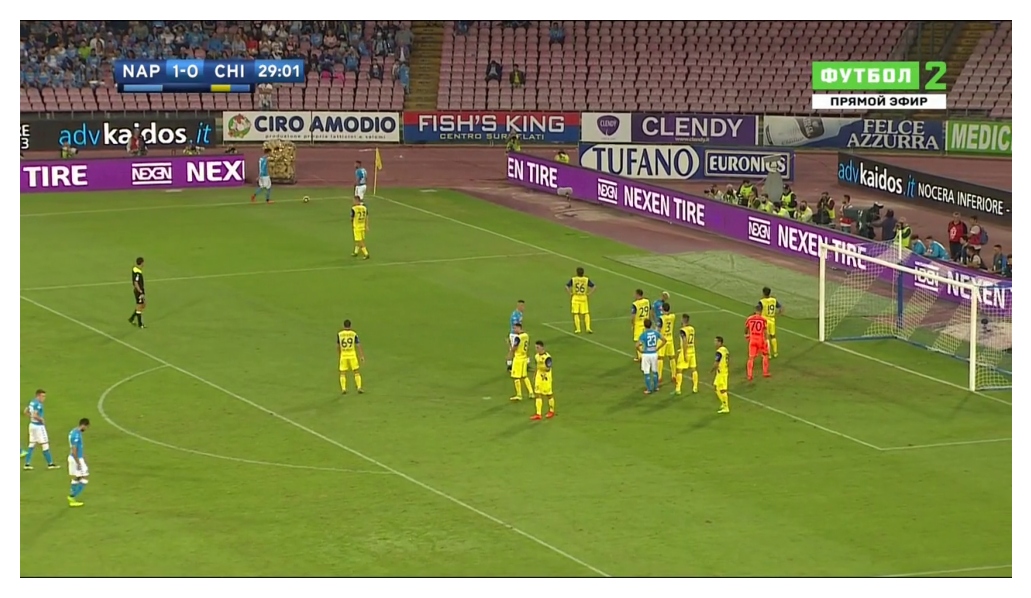} & \includegraphics[width=0.15\textwidth]{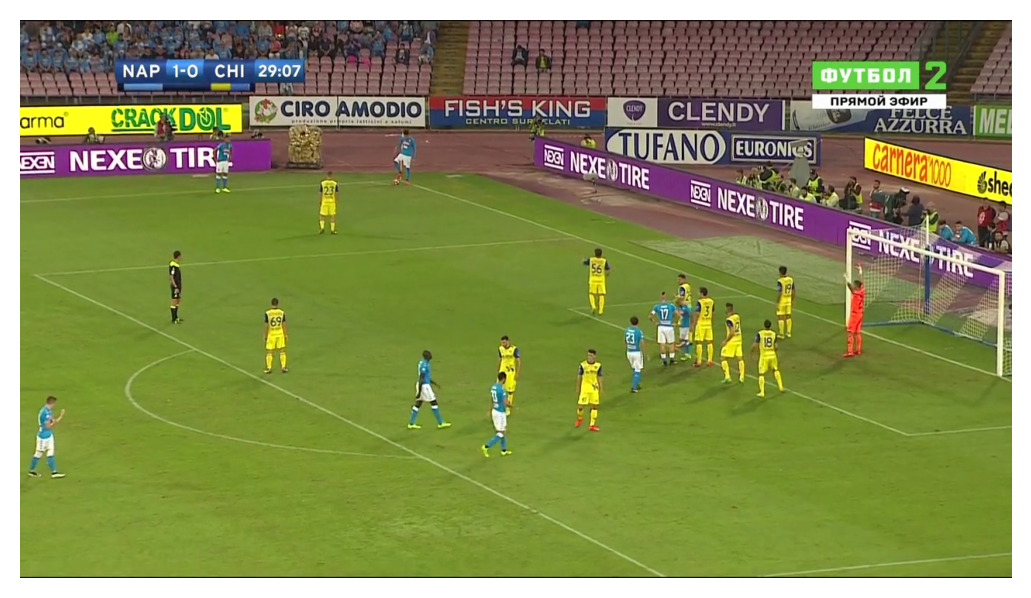} & \includegraphics[width=0.15\textwidth]{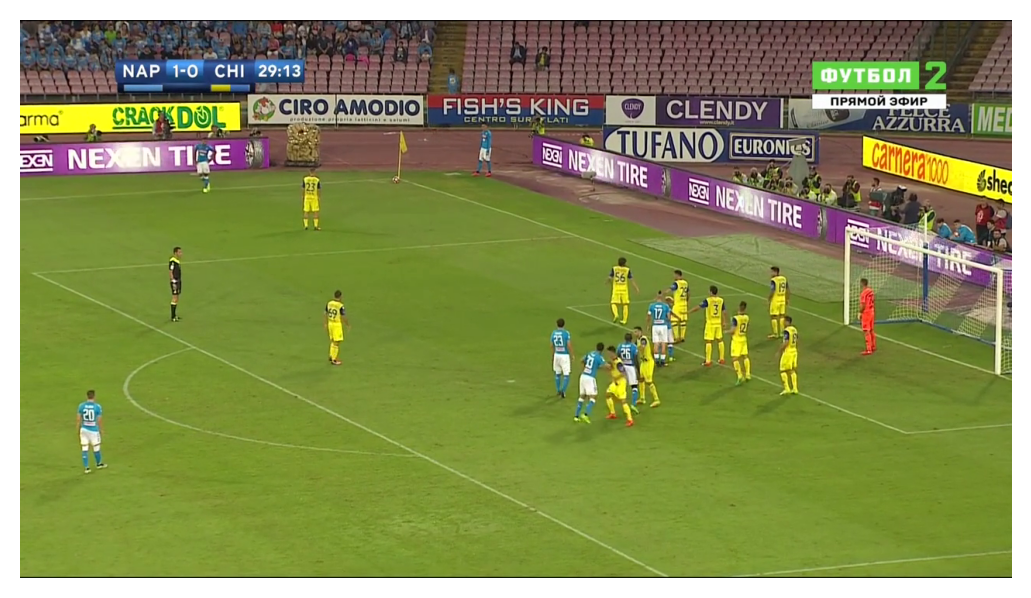} & \includegraphics[width=0.15\textwidth]{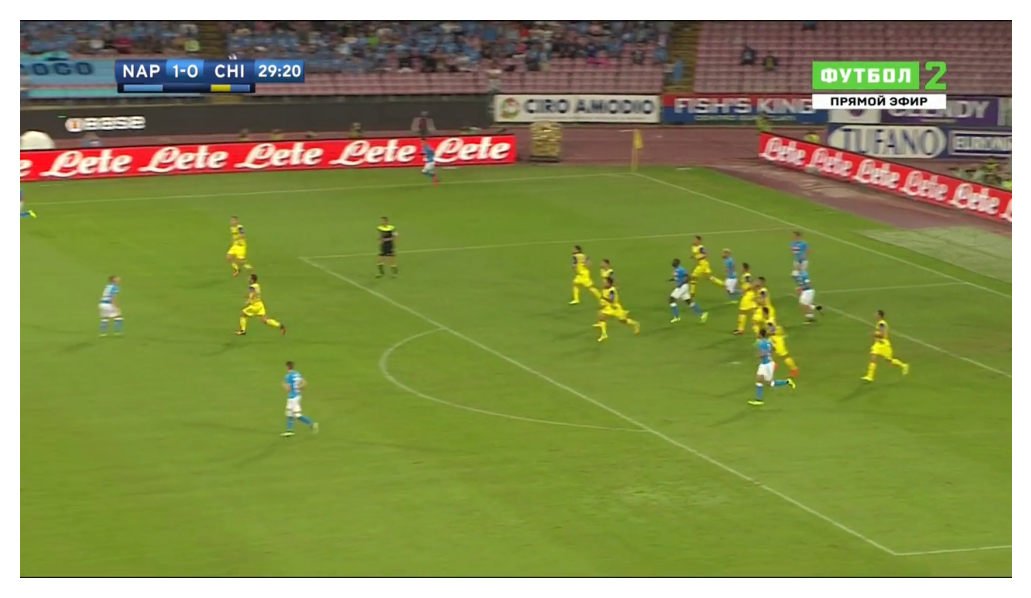} & \includegraphics[width=0.15\textwidth]{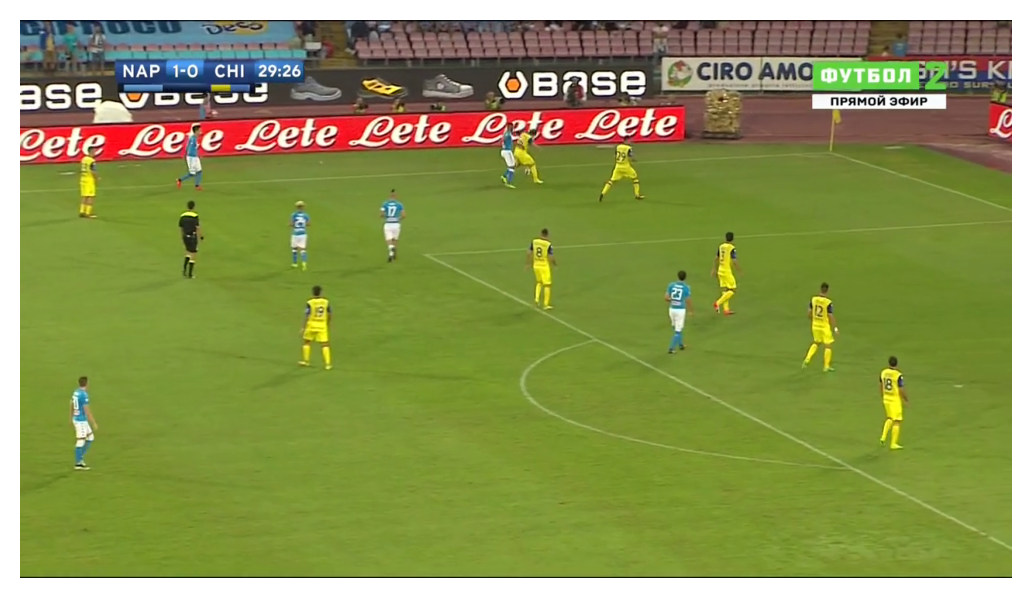}  \\   
        \makecell{Top-1\\(TrajSV)} & \includegraphics[width=0.15\textwidth]{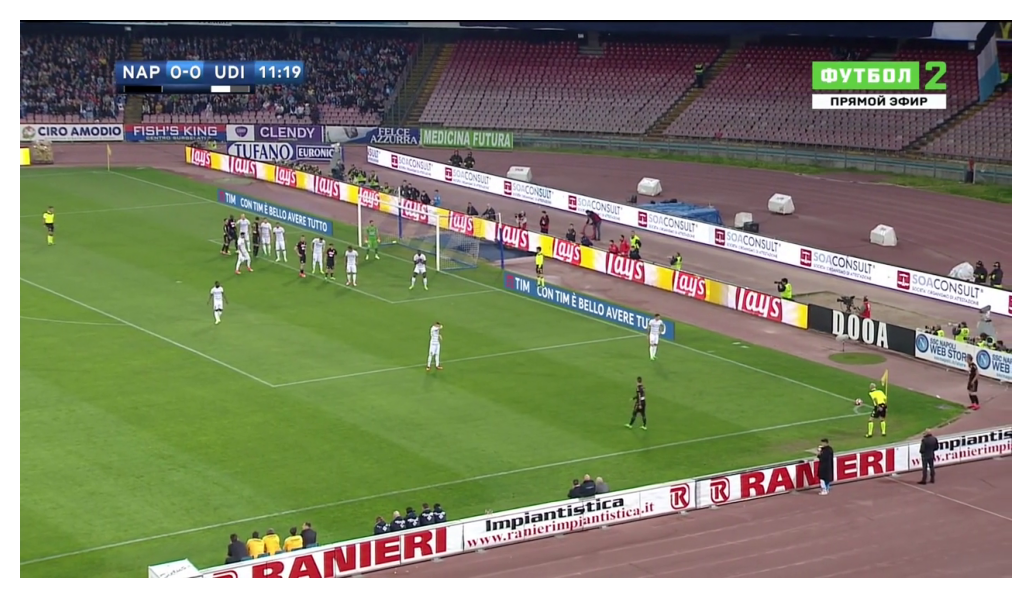} & \includegraphics[width=0.15\textwidth]{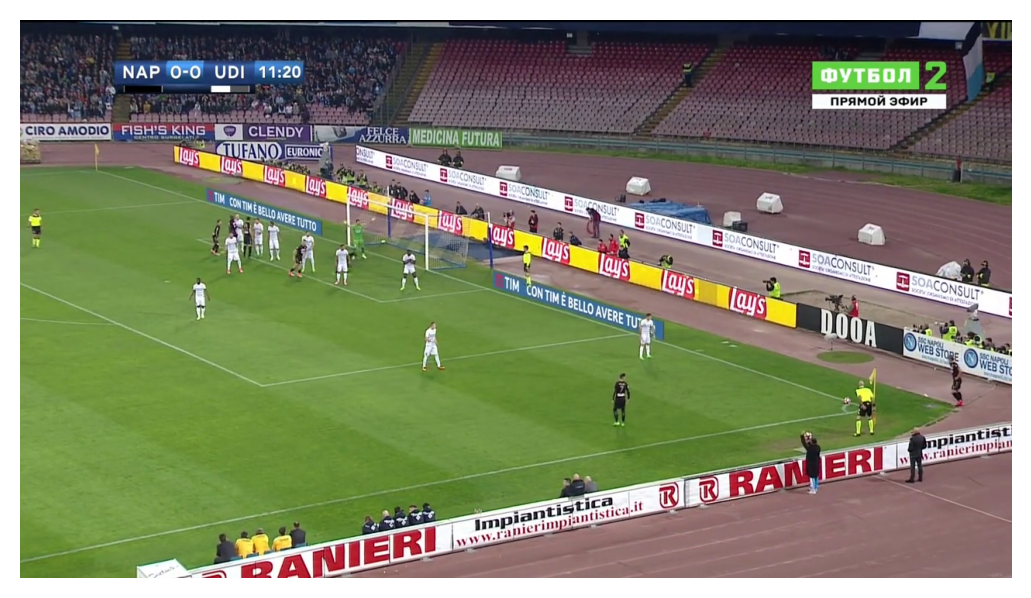} & \includegraphics[width=0.15\textwidth]{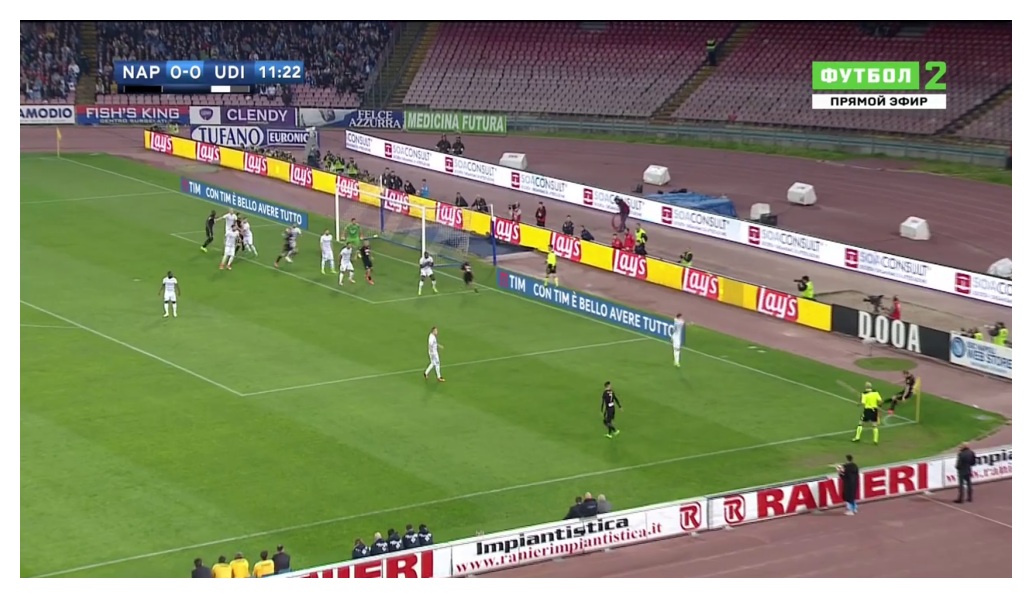} & \includegraphics[width=0.15\textwidth]{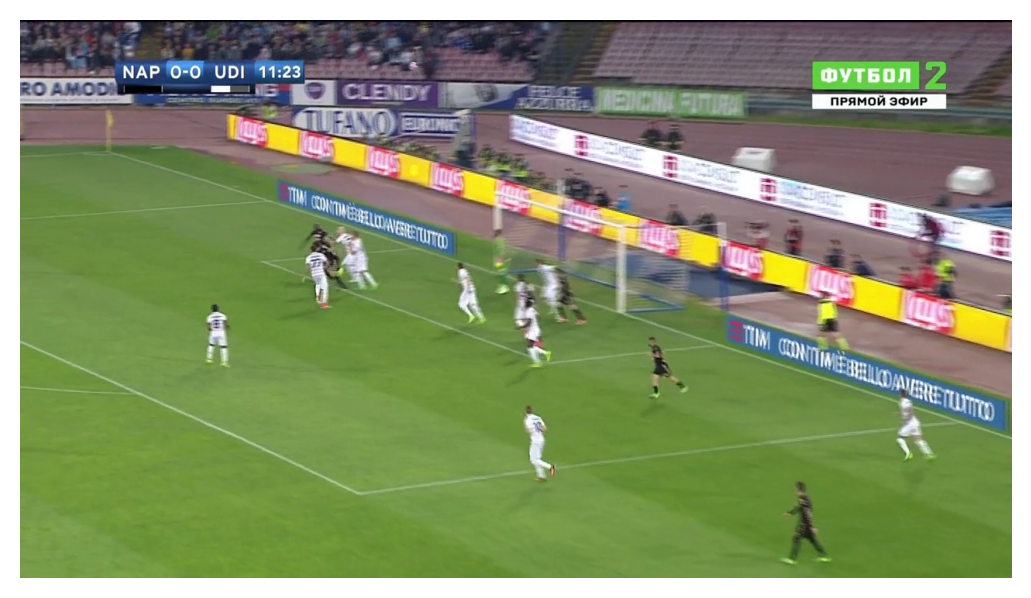} & \includegraphics[width=0.15\textwidth]{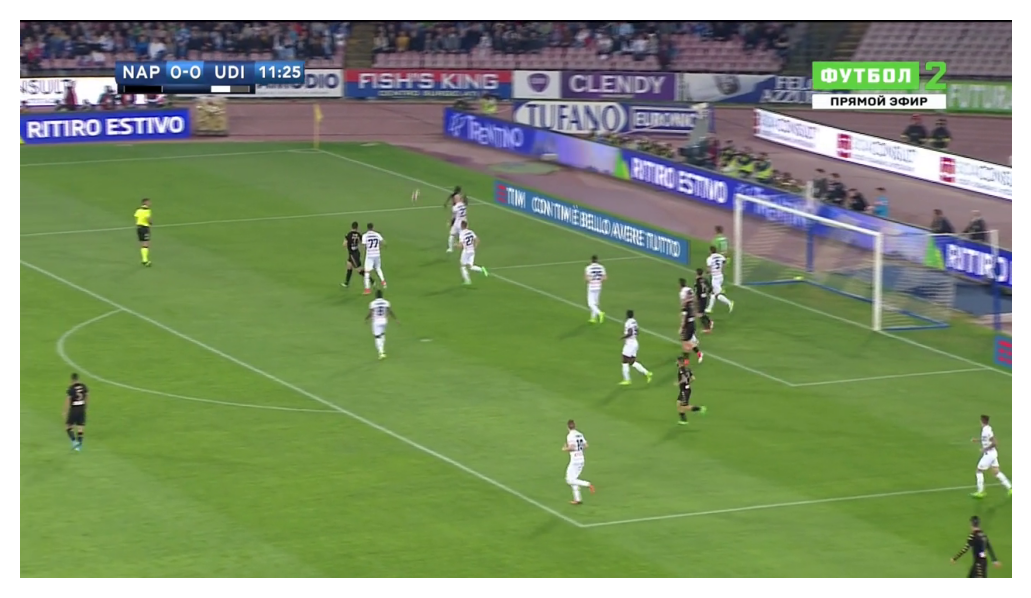} \\ 
        \makecell{{\tcsvt{Top-1}}\\{\tcsvt{(X-CLIP)}}} & \includegraphics[width=0.15\textwidth]{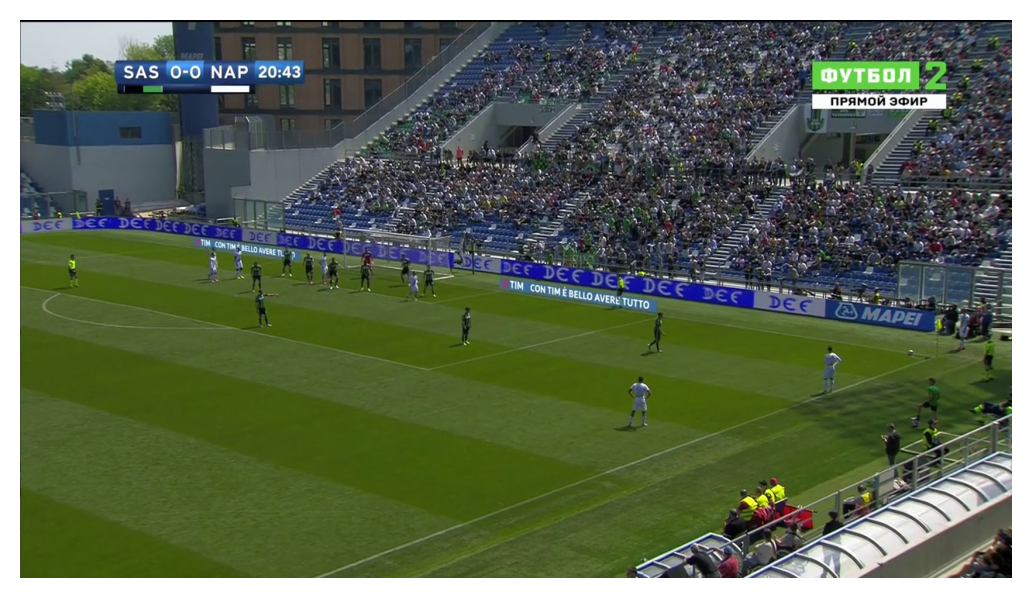} & \includegraphics[width=0.15\textwidth]{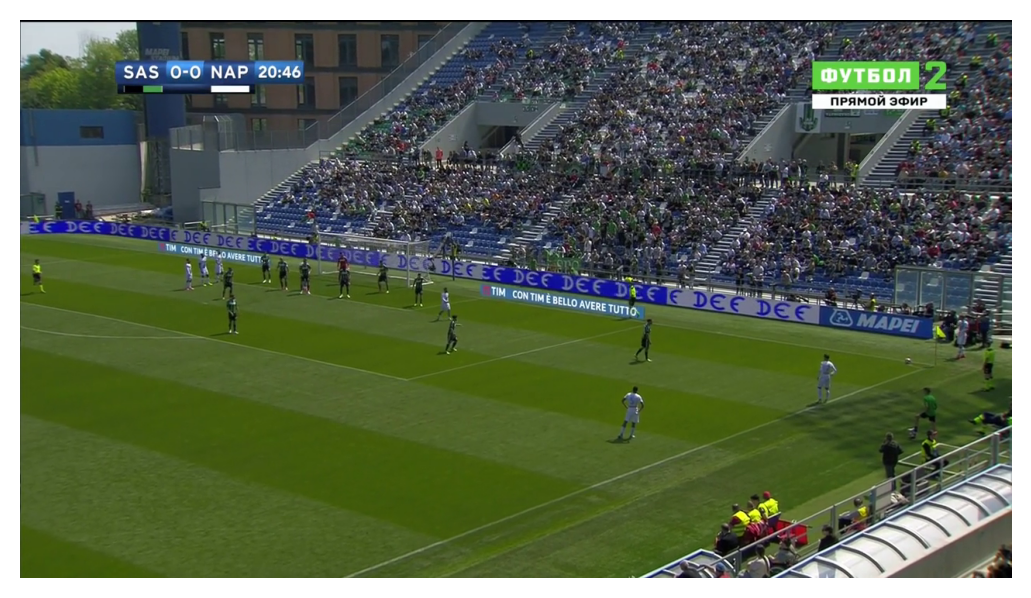} & \includegraphics[width=0.15\textwidth]{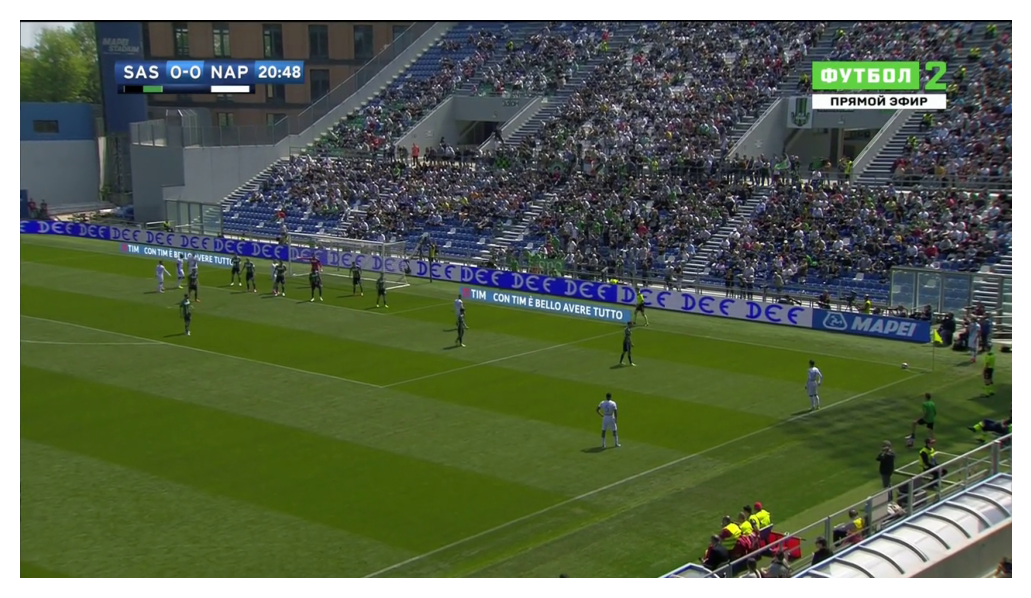} & \includegraphics[width=0.15\textwidth]{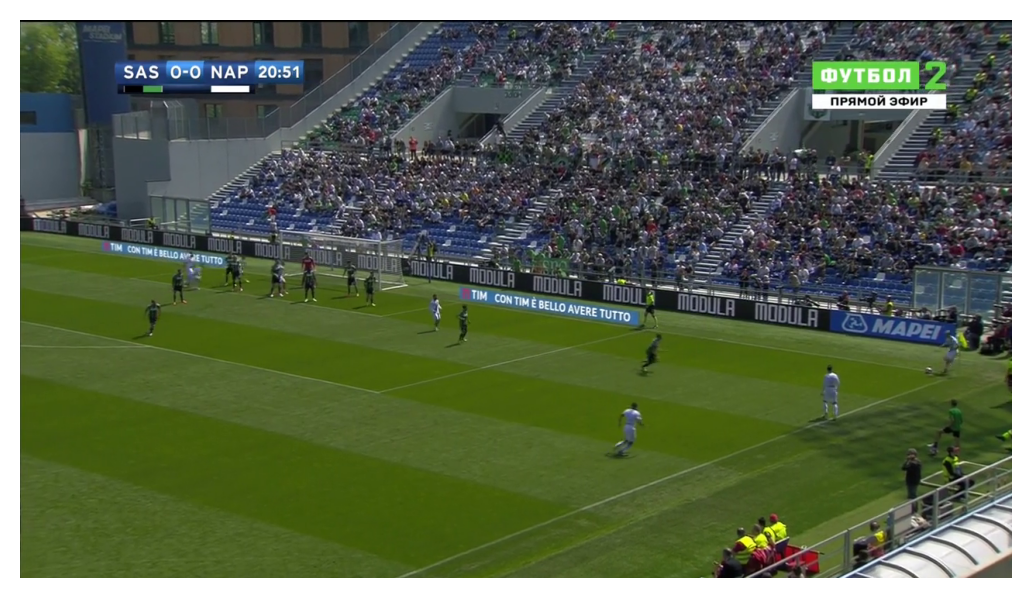} & \includegraphics[width=0.15\textwidth]{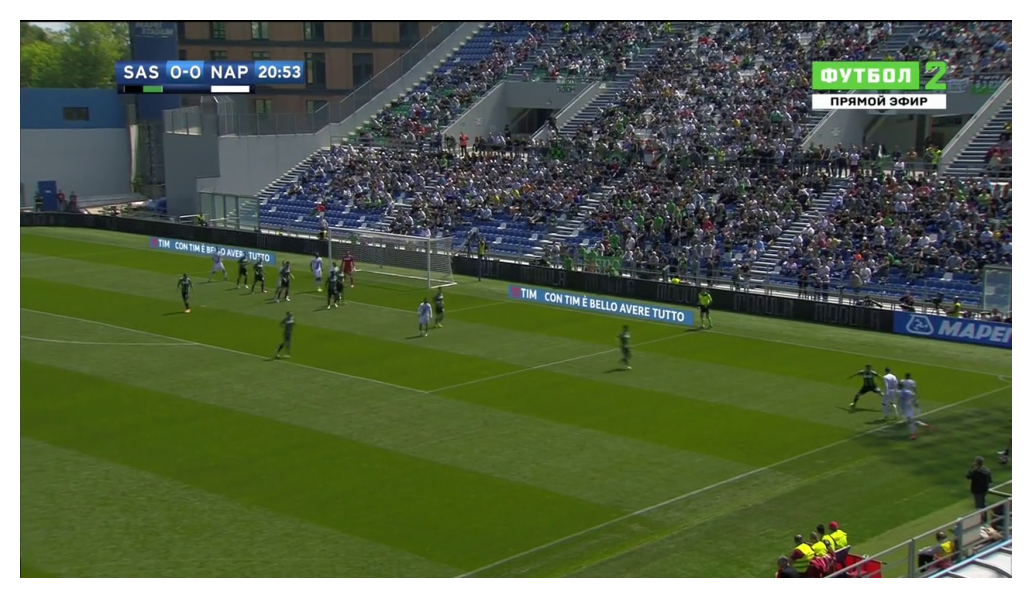}  \\ \hline   
               
        \makecell{Query\\(Direct\\free-kick)} & \includegraphics[width=0.15\textwidth]{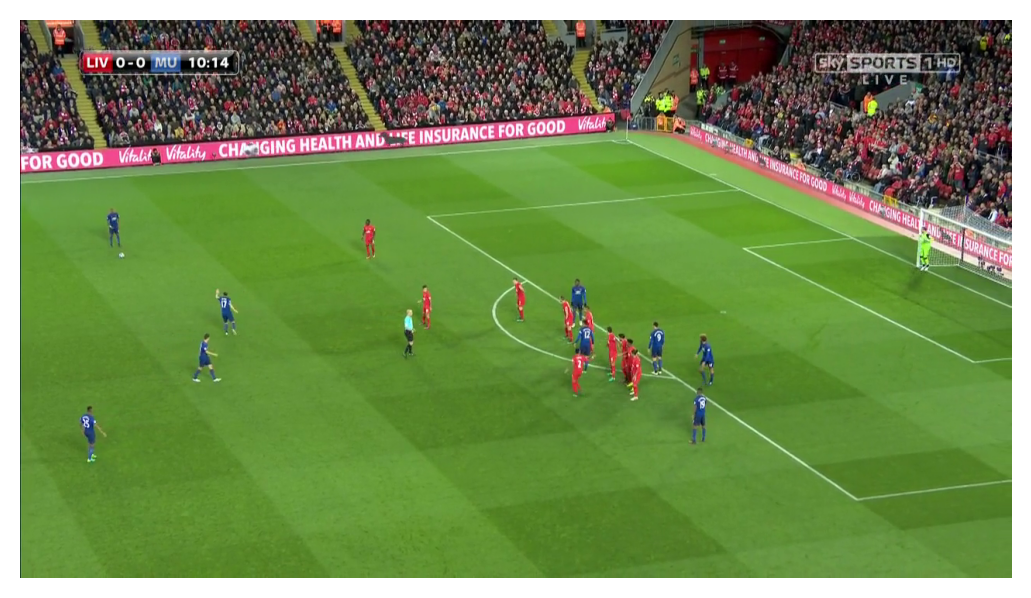} & \includegraphics[width=0.15\textwidth]{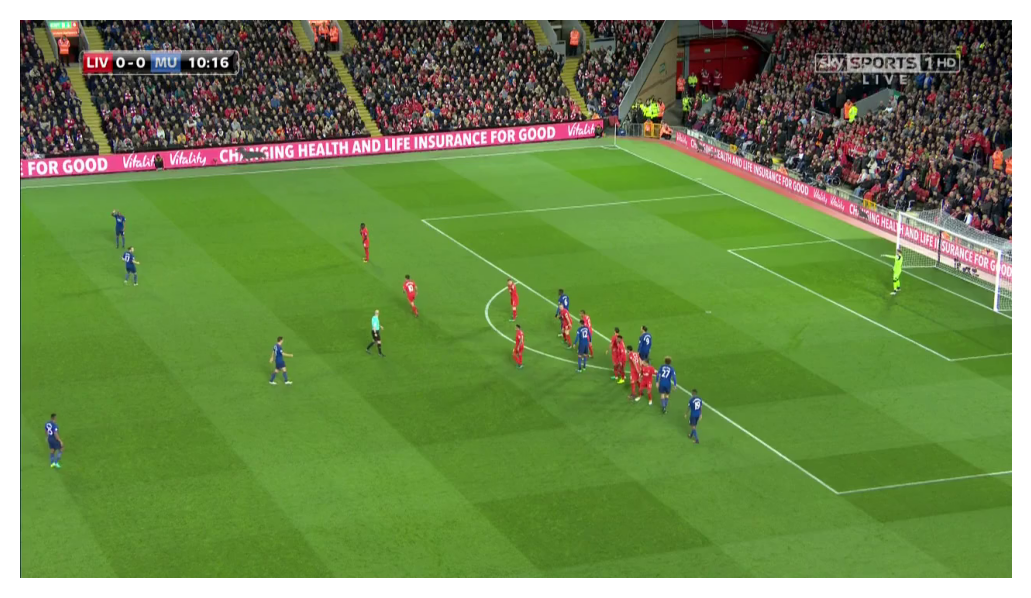} & \includegraphics[width=0.15\textwidth]{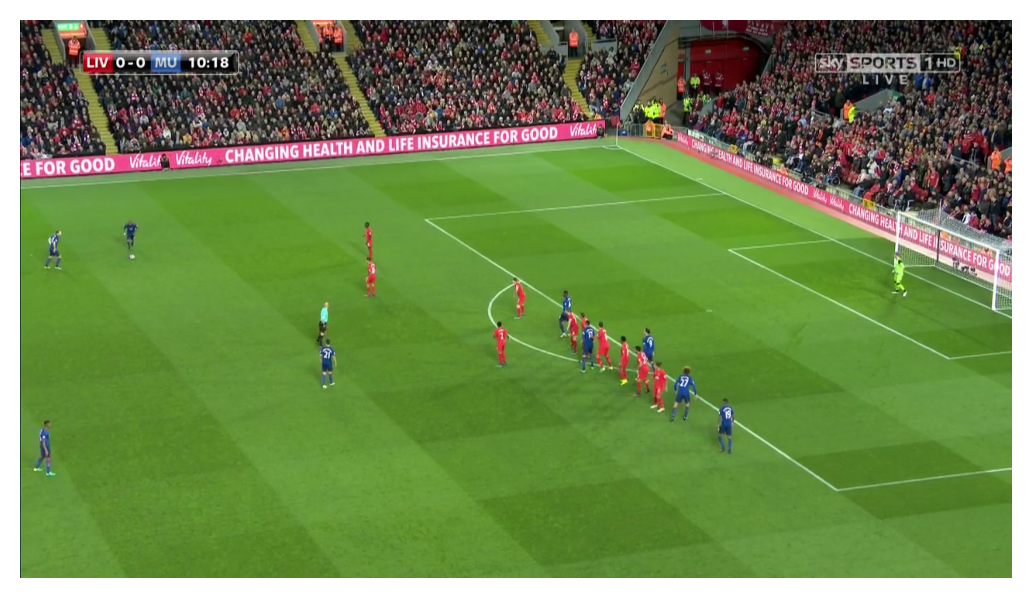} & \includegraphics[width=0.15\textwidth]{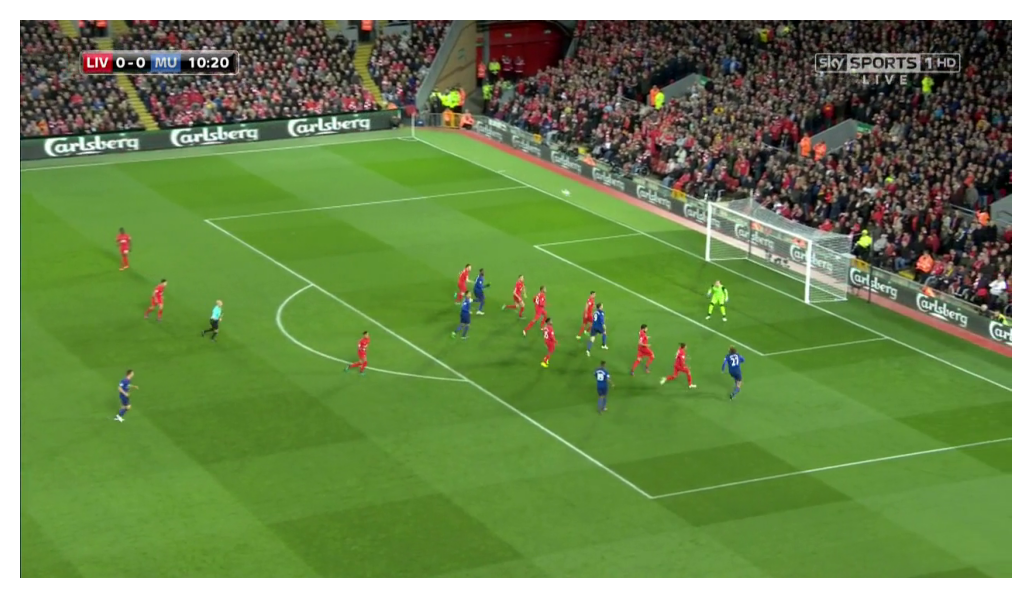} & \includegraphics[width=0.15\textwidth]{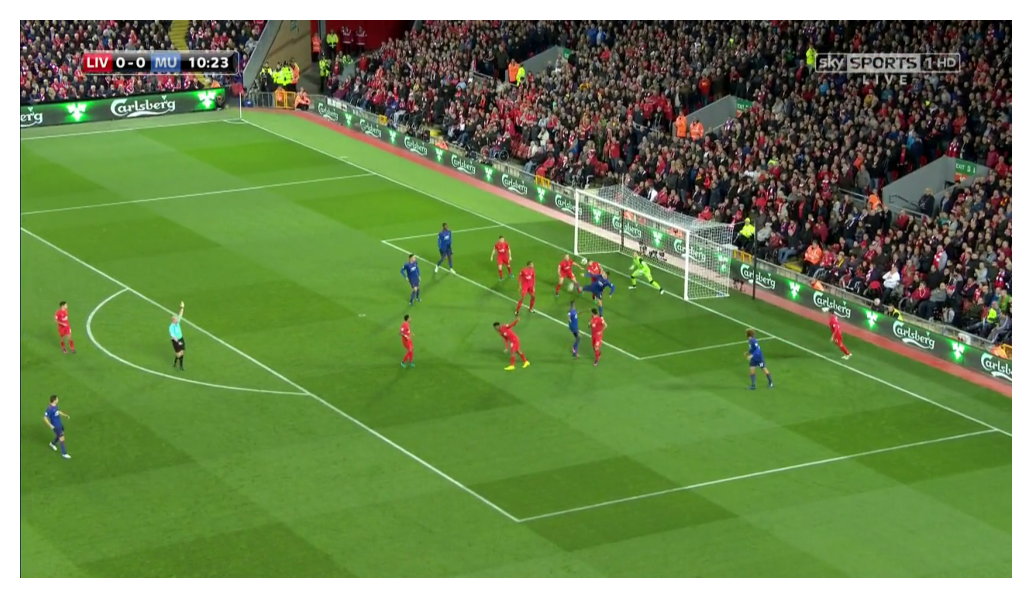}  \\   
        \makecell{Top-1\\(TrajSV)} & \includegraphics[width=0.15\textwidth]{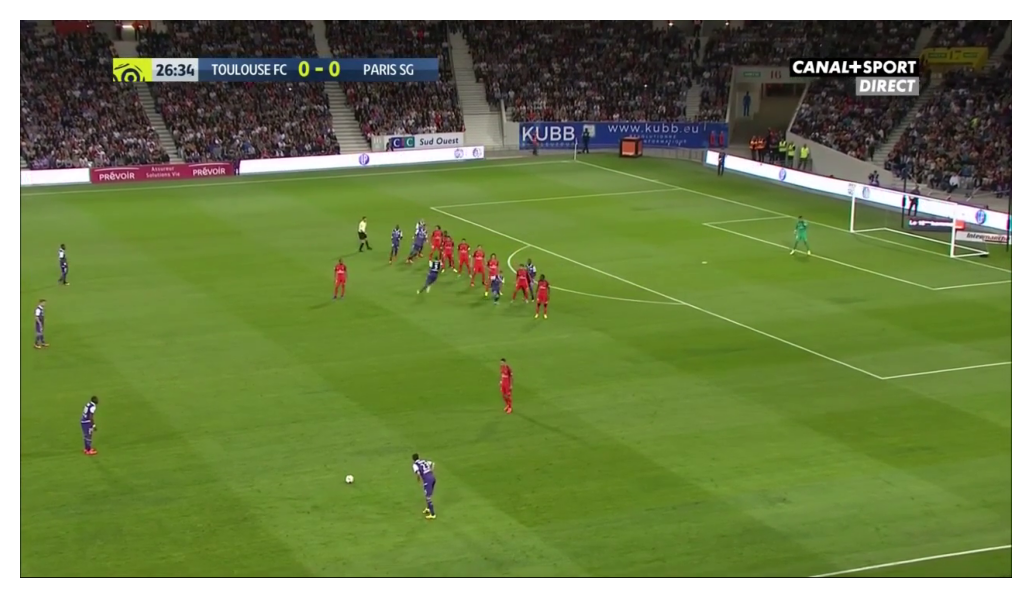} & \includegraphics[width=0.15\textwidth]{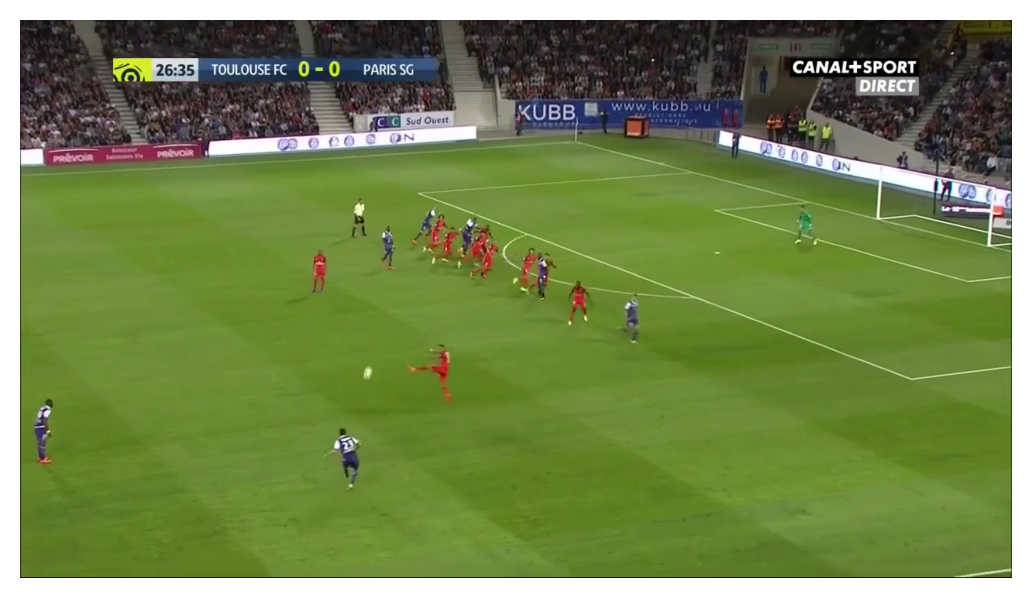} & \includegraphics[width=0.15\textwidth]{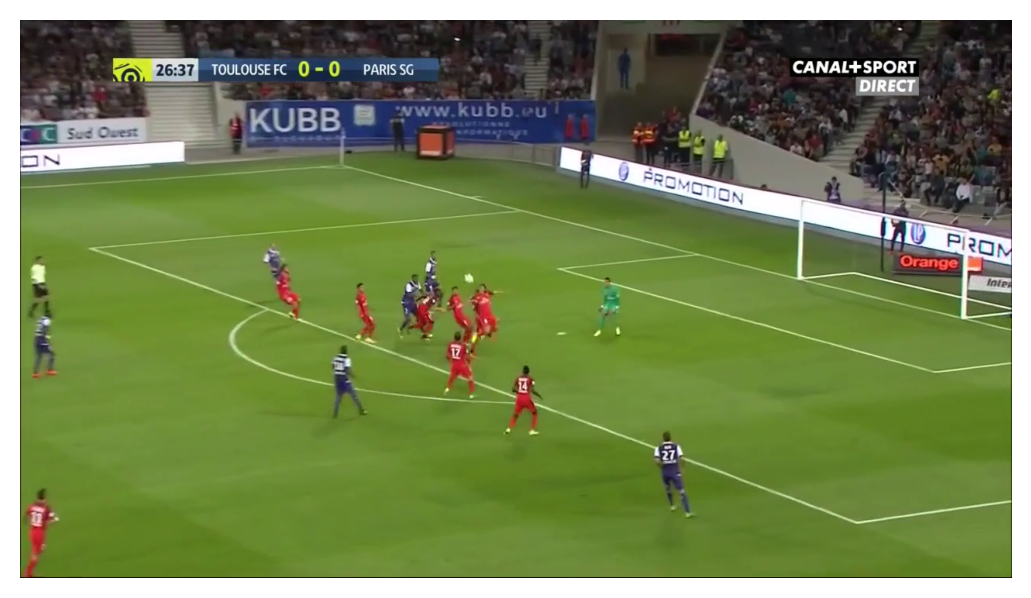} & \includegraphics[width=0.15\textwidth]{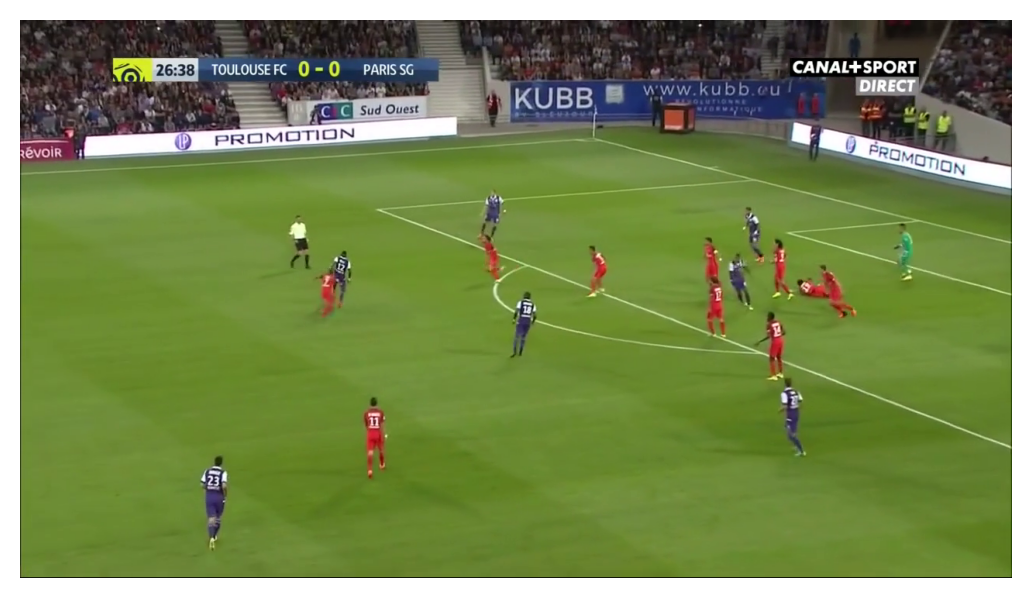} & \includegraphics[width=0.15\textwidth]{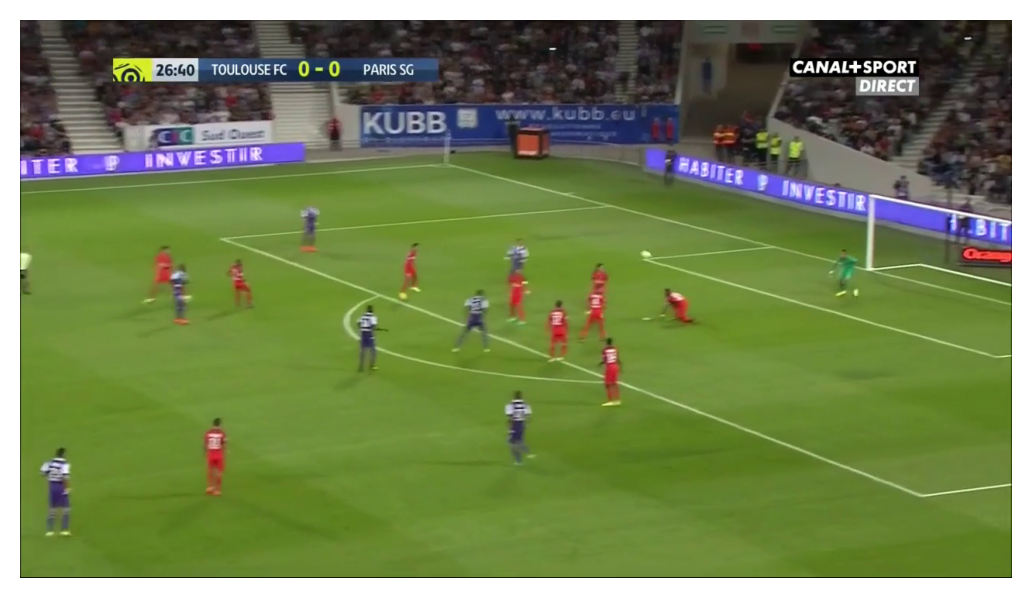}  \\   
        \makecell{{\tcsvt{Top-1}}\\{\tcsvt{(X-CLIP)}}} & \includegraphics[width=0.15\textwidth]{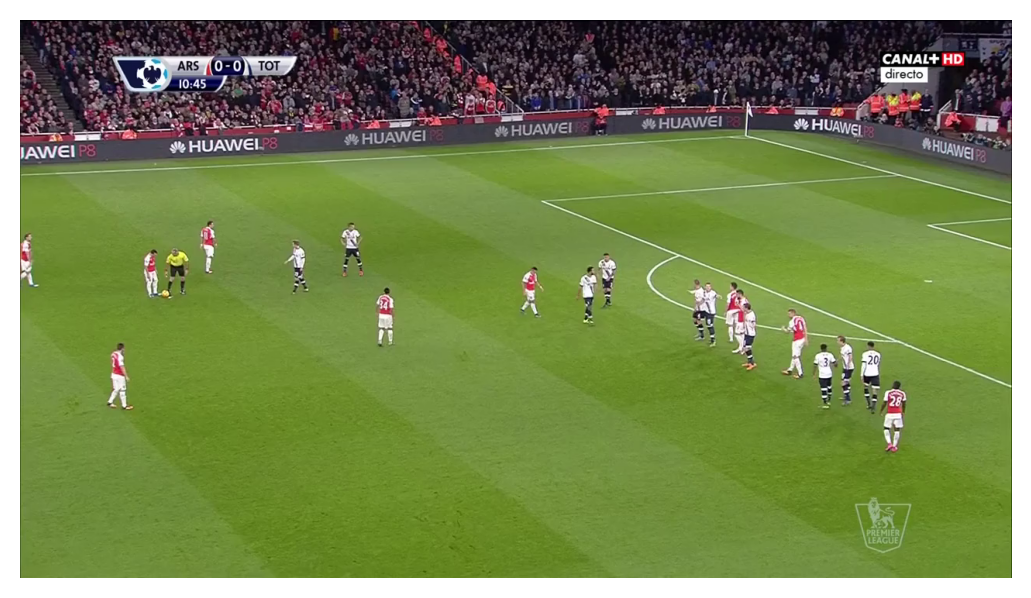} & \includegraphics[width=0.15\textwidth]{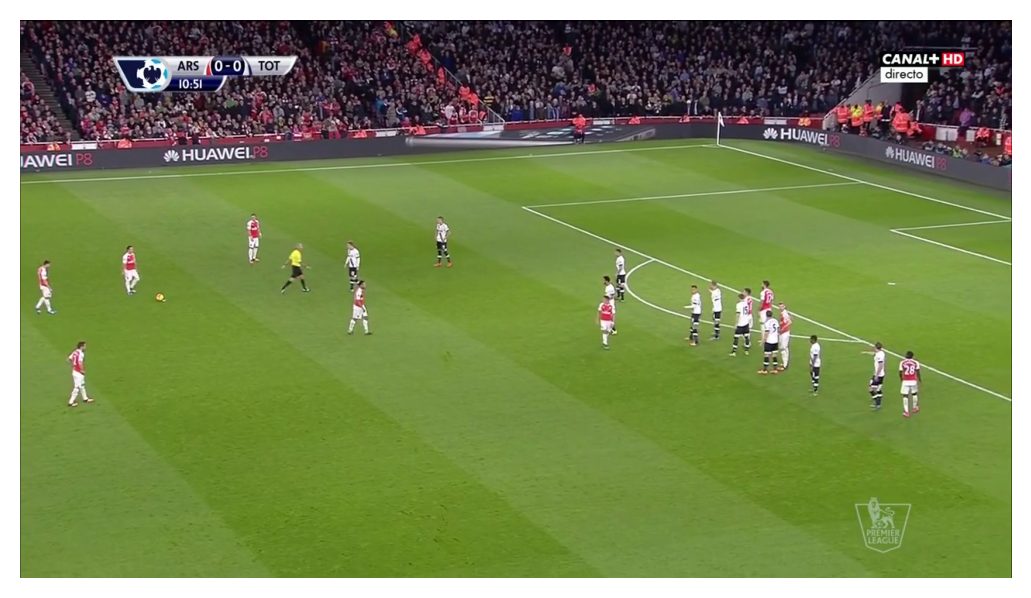} & \includegraphics[width=0.15\textwidth]{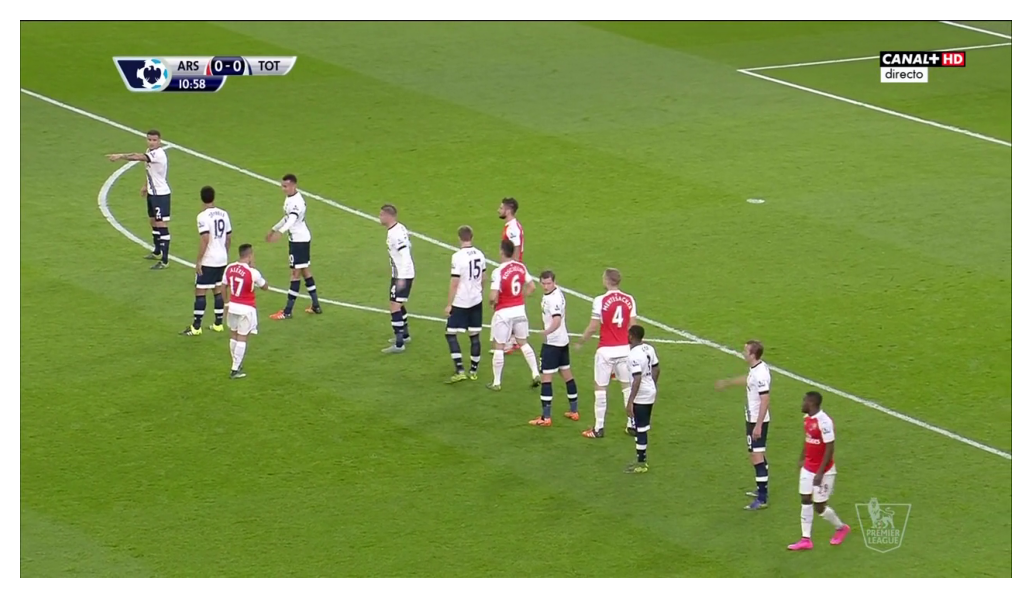} & \includegraphics[width=0.15\textwidth]{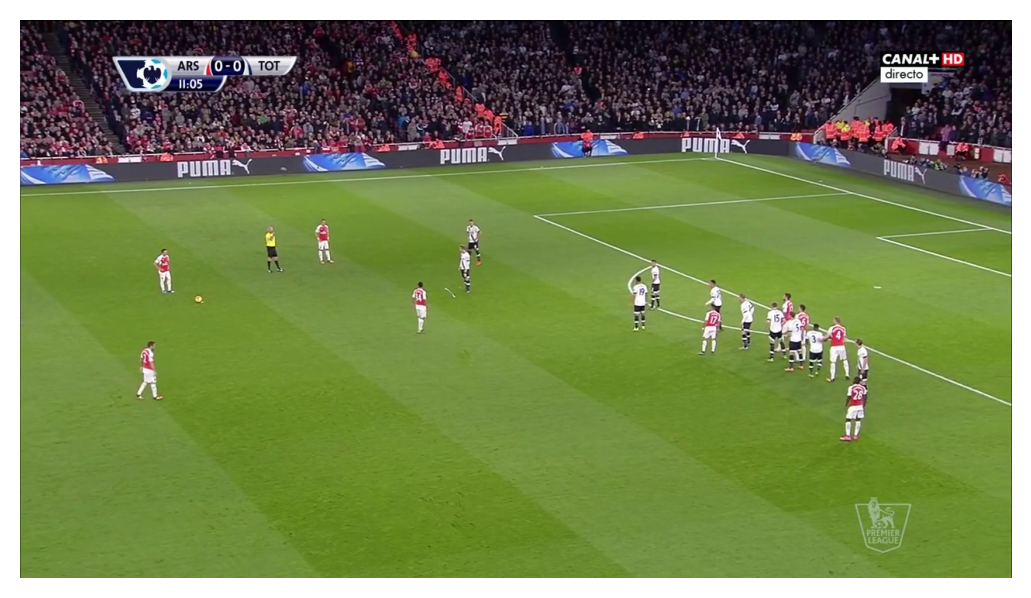} & \includegraphics[width=0.15\textwidth]{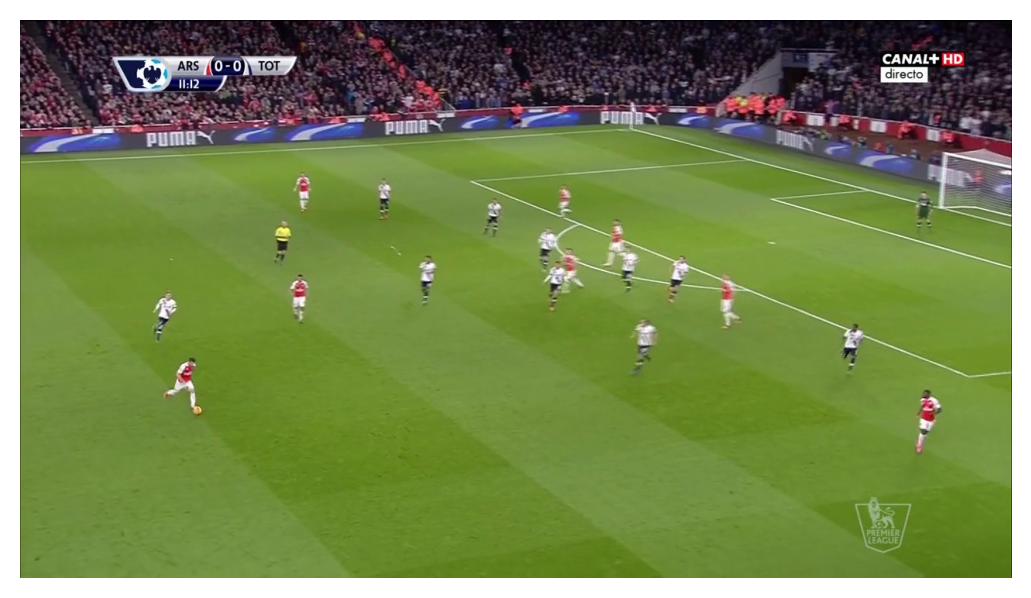} \\   
        % \makecell{Top3\\(0.796)} & \includegraphics[width=0.15\textwidth]{figures/retrieval/Direct free-kick/top3/2017-04-02 - 21-45 Granada CF 1 - 4 Barcelona_1 - 34-40_2073.3.png} & \includegraphics[width=0.15\textwidth]{figures/retrieval/Direct free-kick/top3/2017-04-02 - 21-45 Granada CF 1 - 4 Barcelona_1 - 34-40_2076.8.png} & \includegraphics[width=0.15\textwidth]{figures/retrieval/Direct free-kick/top3/2017-04-02 - 21-45 Granada CF 1 - 4 Barcelona_1 - 34-40_2079.25.png} & \includegraphics[width=0.15\textwidth]{figures/retrieval/Direct free-kick/top3/2017-04-02 - 21-45 Granada CF 1 - 4 Barcelona_1 - 34-40_2082.75.png} & \includegraphics[width=0.15\textwidth]{figures/retrieval/Direct free-kick/top3/2017-04-02 - 21-45 Granada CF 1 - 4 Barcelona_1 - 34-40_2085.55.png} & \includegraphics[width=0.15\textwidth]{figures/retrieval/Direct free-kick/top3/2017-04-02 - 21-45 Granada CF 1 - 4 Barcelona.png}  \\   
        \hline

        \makecell{Query\\(Volleyball)} & \includegraphics[width=0.15\textwidth]{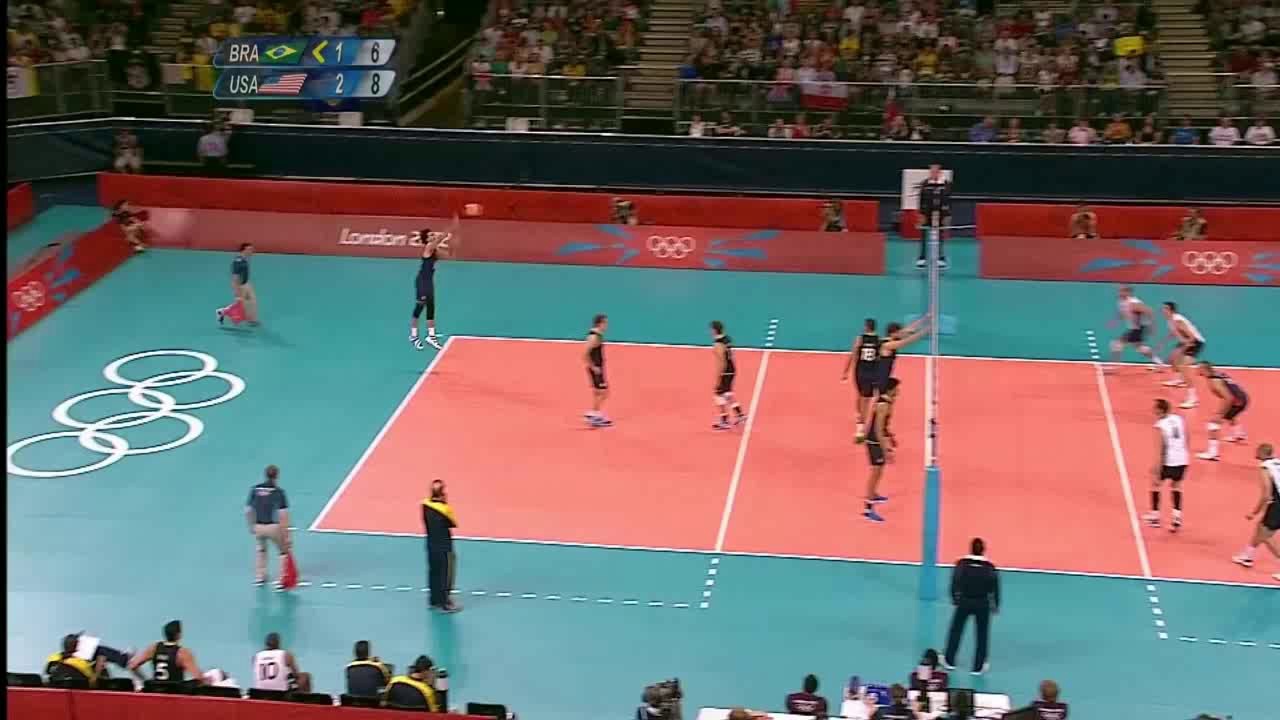} & \includegraphics[width=0.15\textwidth]{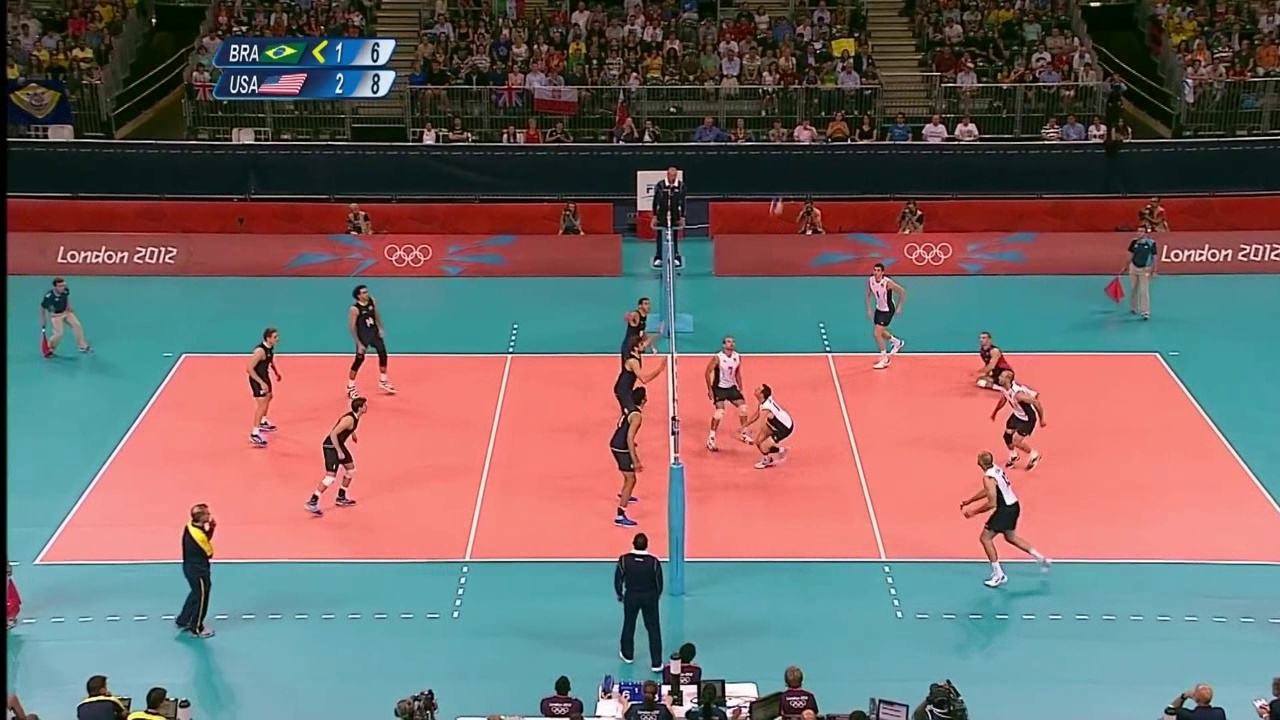} & \includegraphics[width=0.15\textwidth]{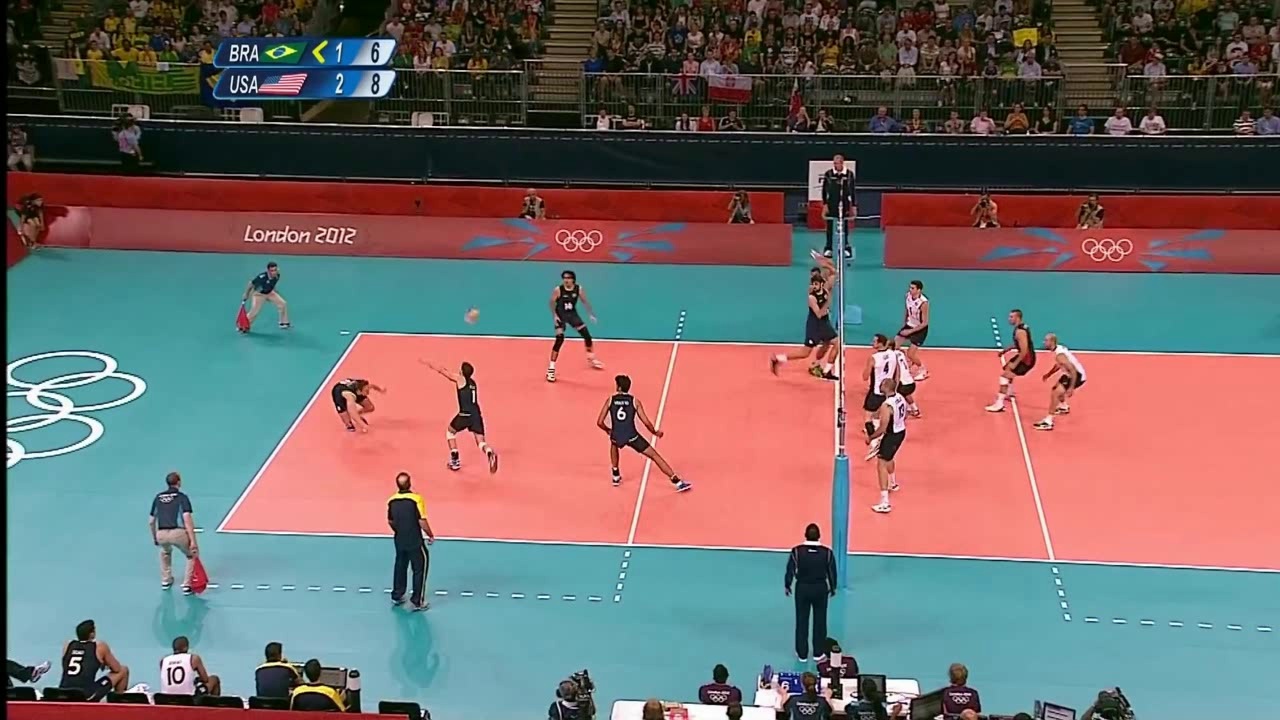} & \includegraphics[width=0.15\textwidth]{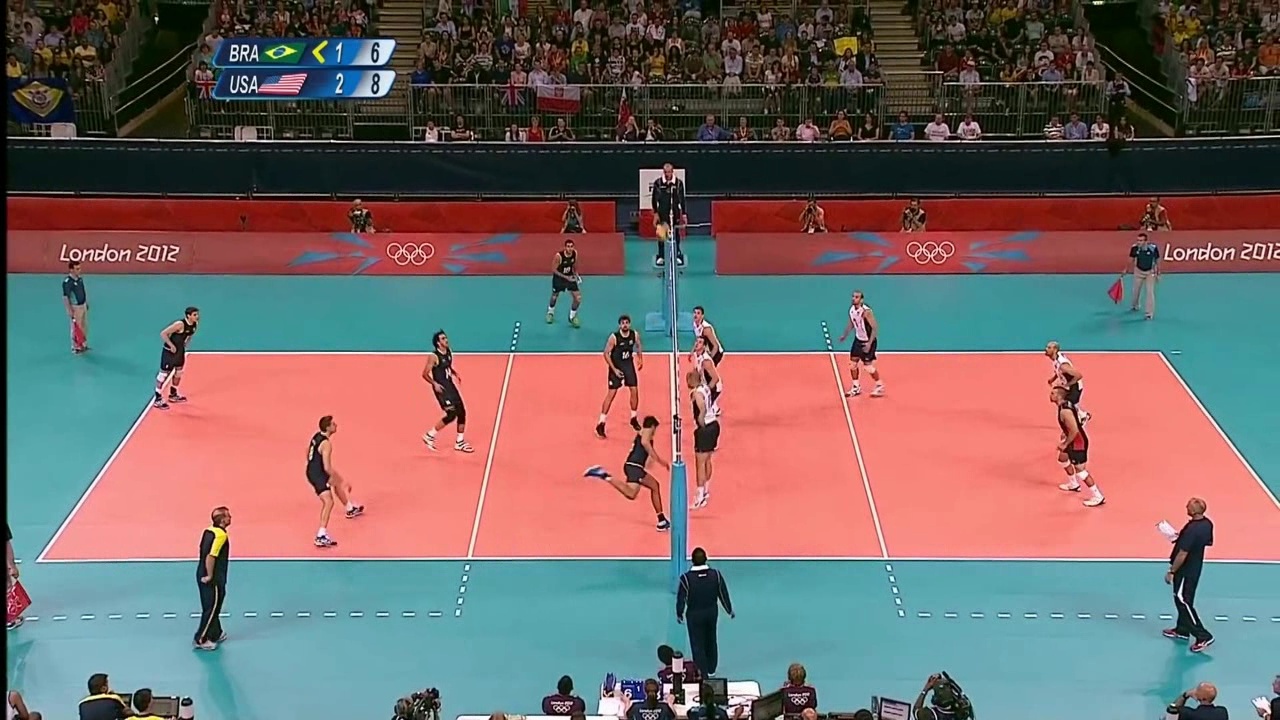} & \includegraphics[width=0.15\textwidth]{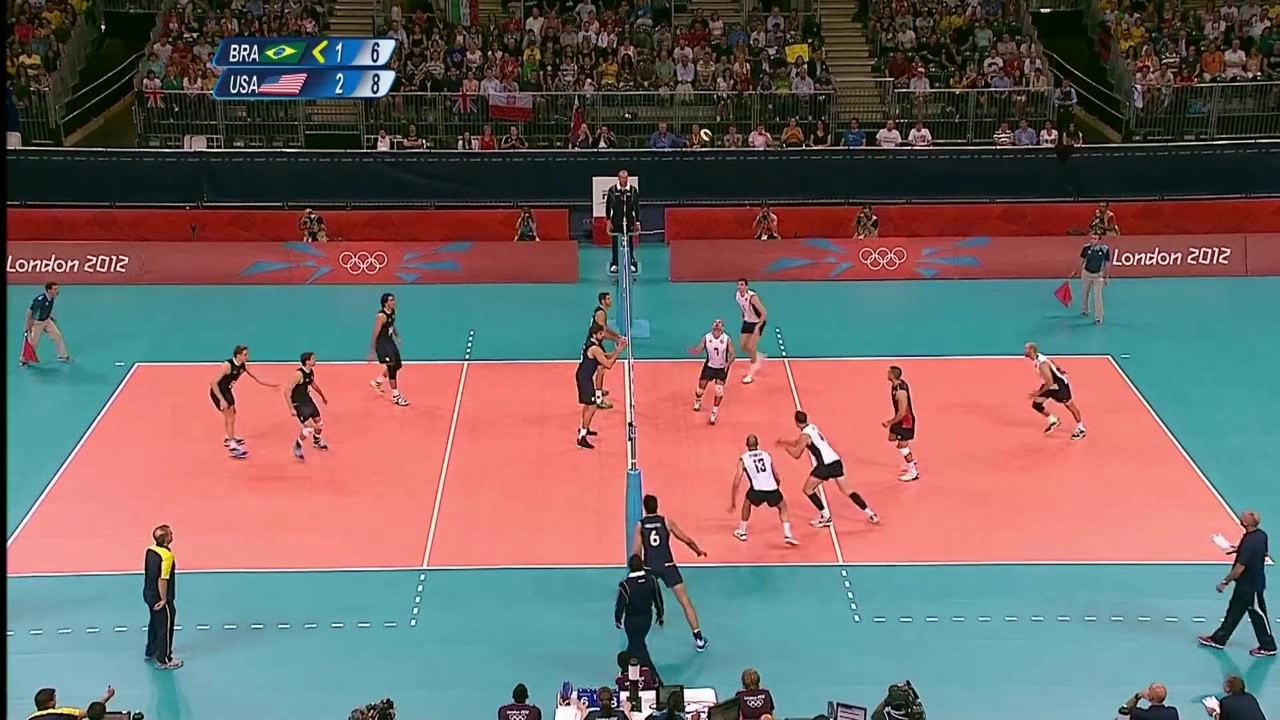}  \\   
        \makecell{Top-1\\(TrajSV)} & \includegraphics[width=0.15\textwidth]{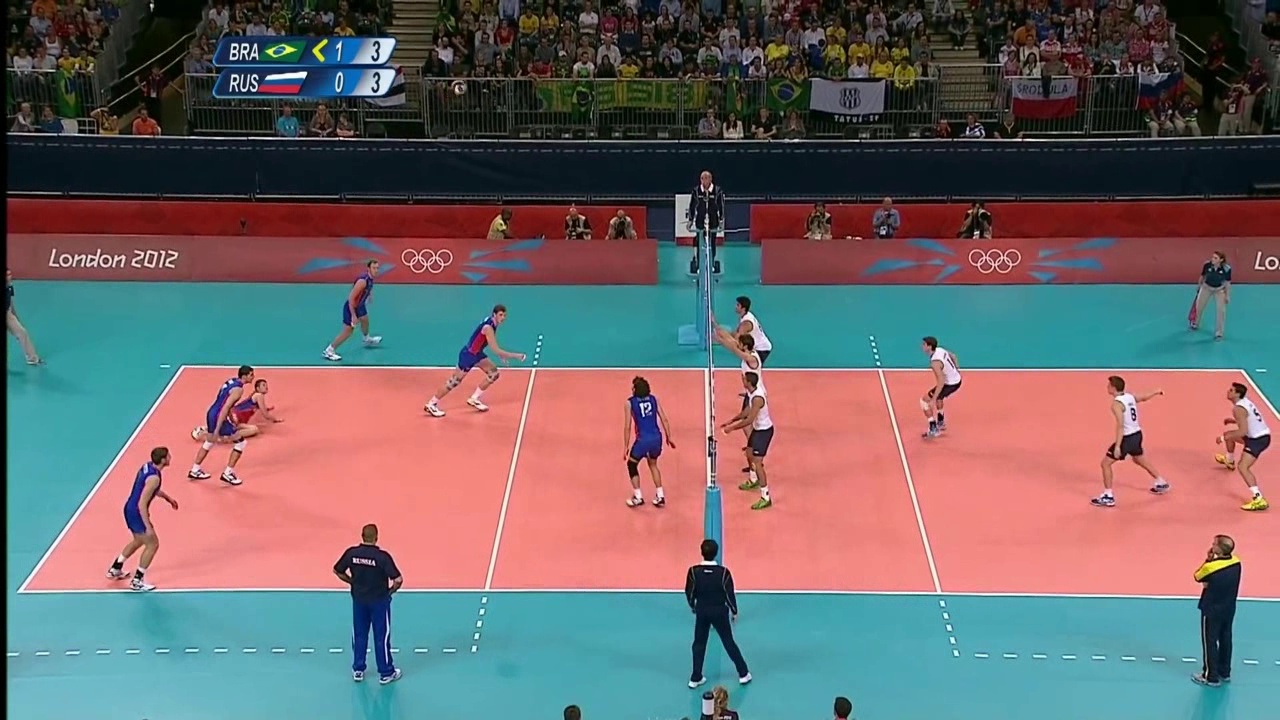} & \includegraphics[width=0.15\textwidth]{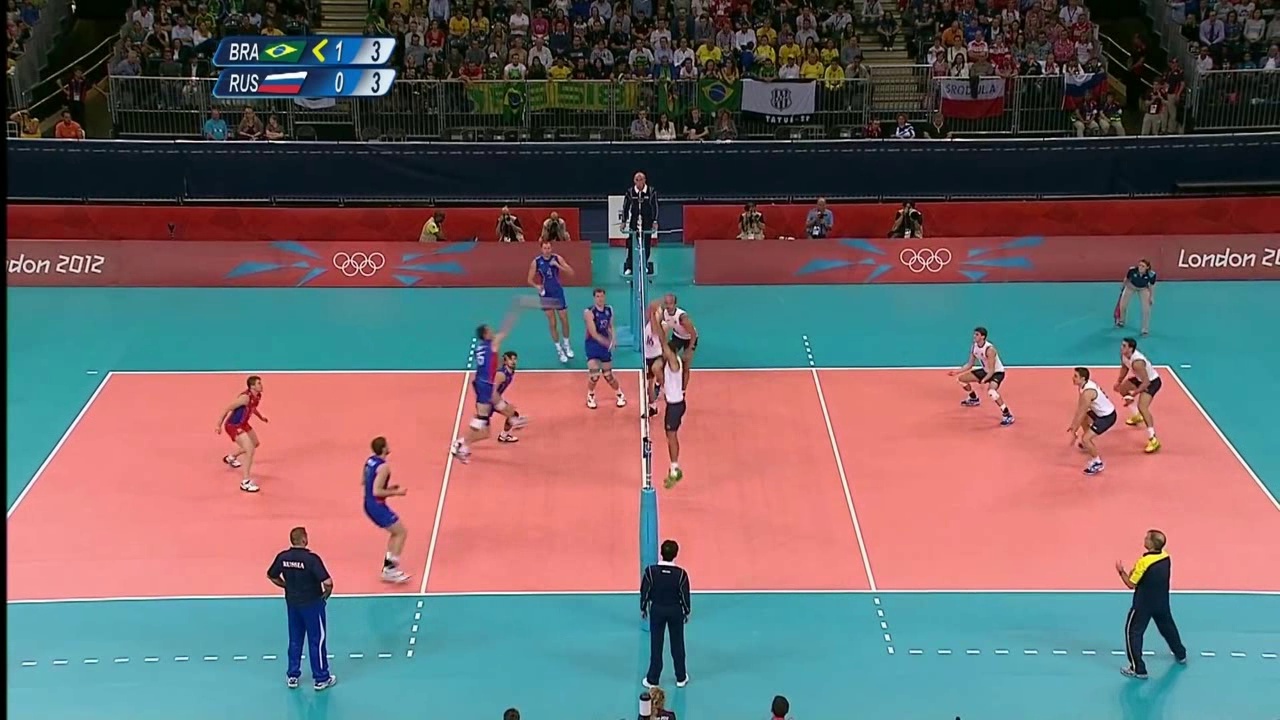} & \includegraphics[width=0.15\textwidth]{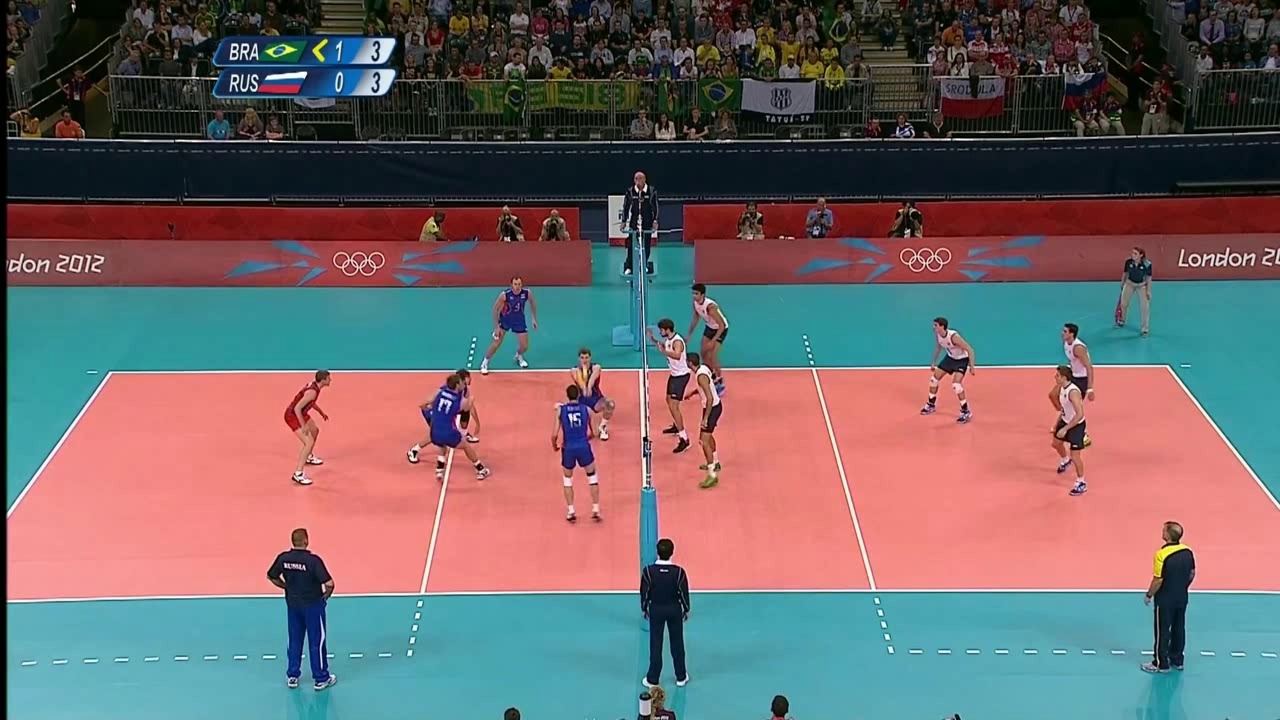} & \includegraphics[width=0.15\textwidth]{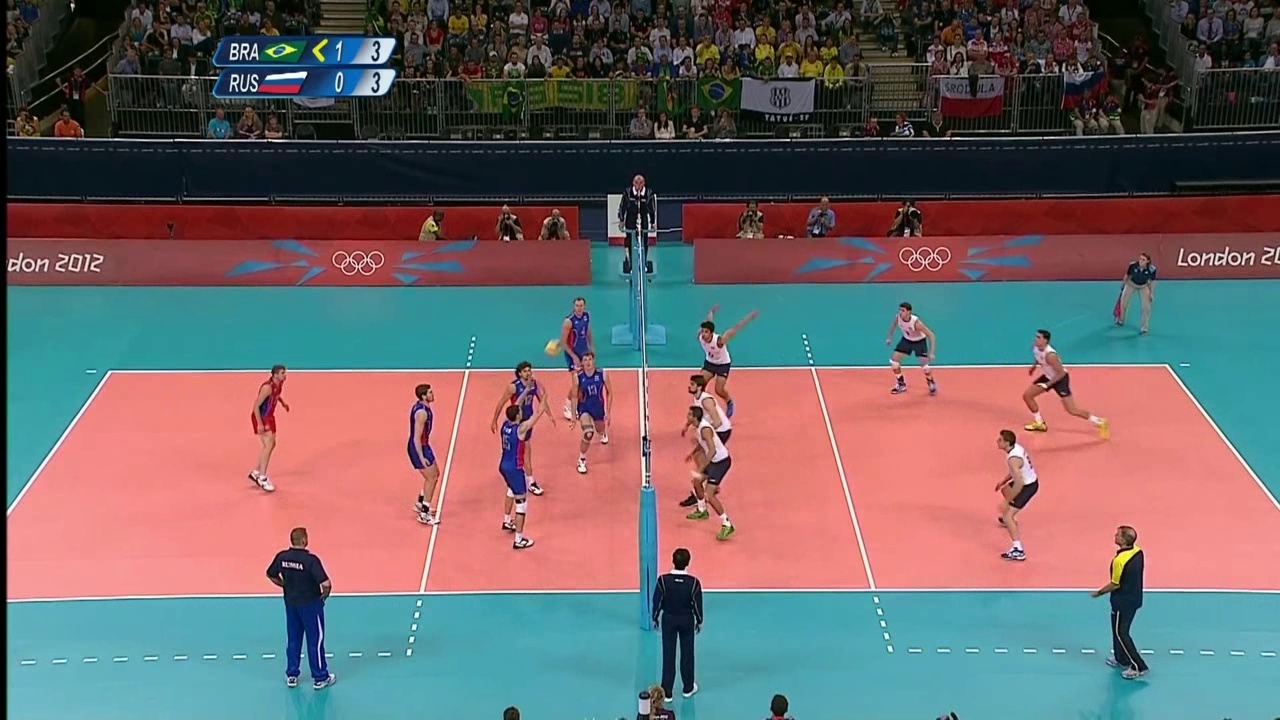} & \includegraphics[width=0.15\textwidth]{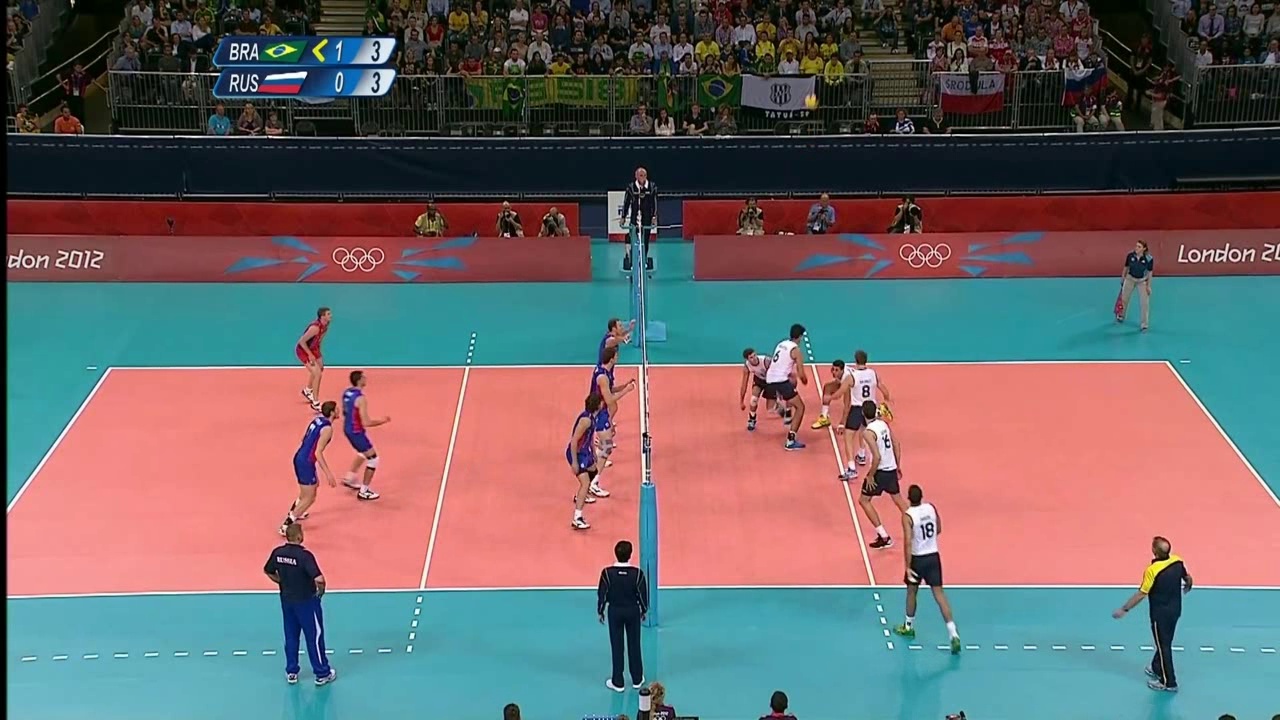}   \\  
        \hline
        
        \makecell{Query\\(Basketball)} & \includegraphics[width=0.15\textwidth]{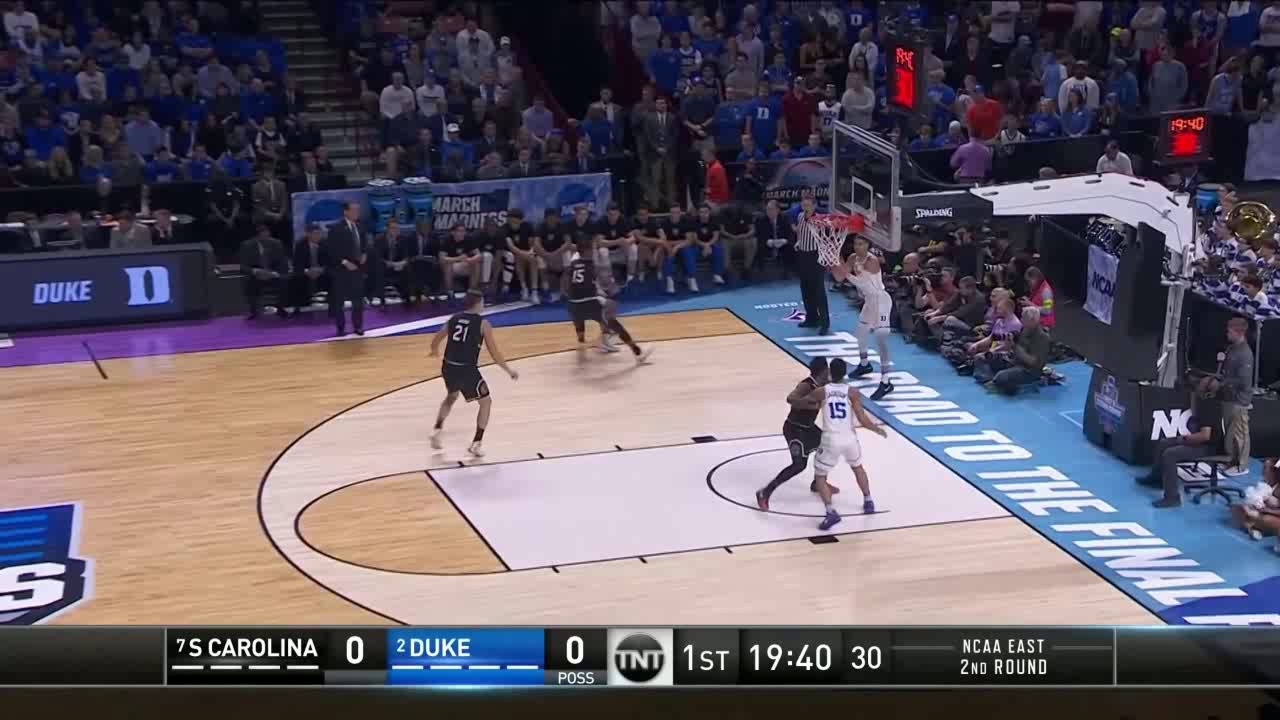} & \includegraphics[width=0.15\textwidth]{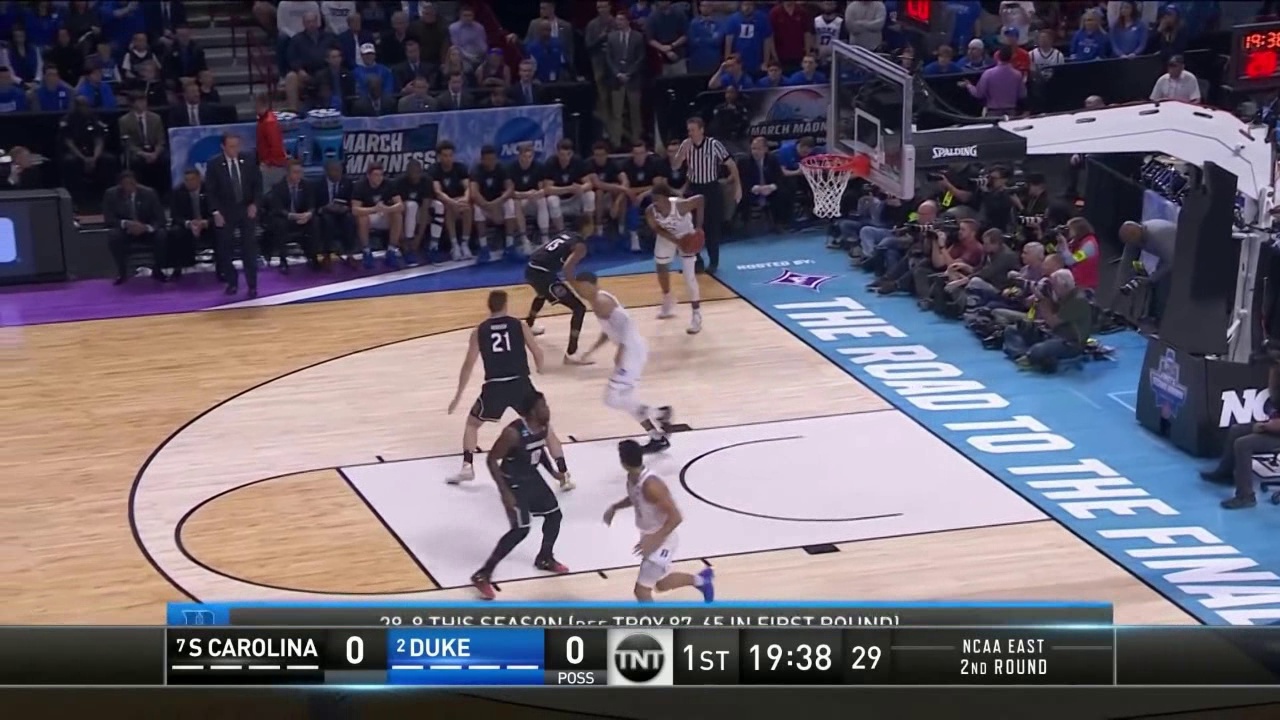} & \includegraphics[width=0.15\textwidth]{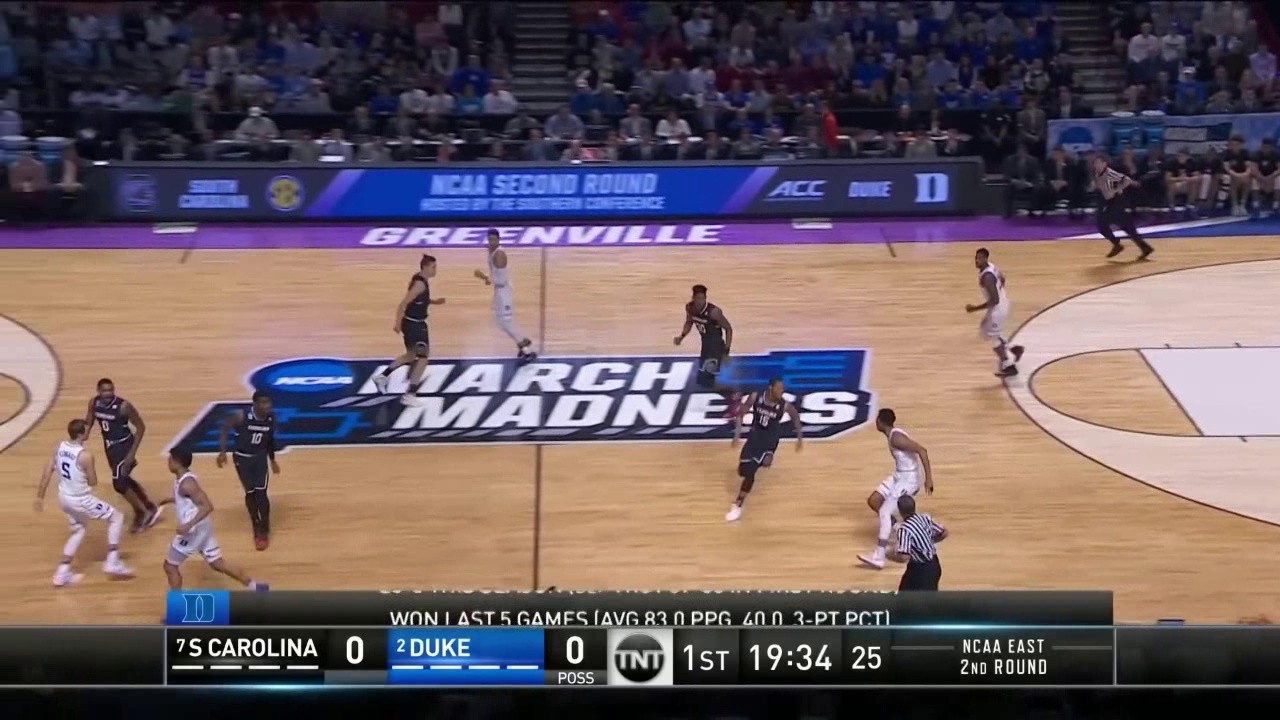} & \includegraphics[width=0.15\textwidth]{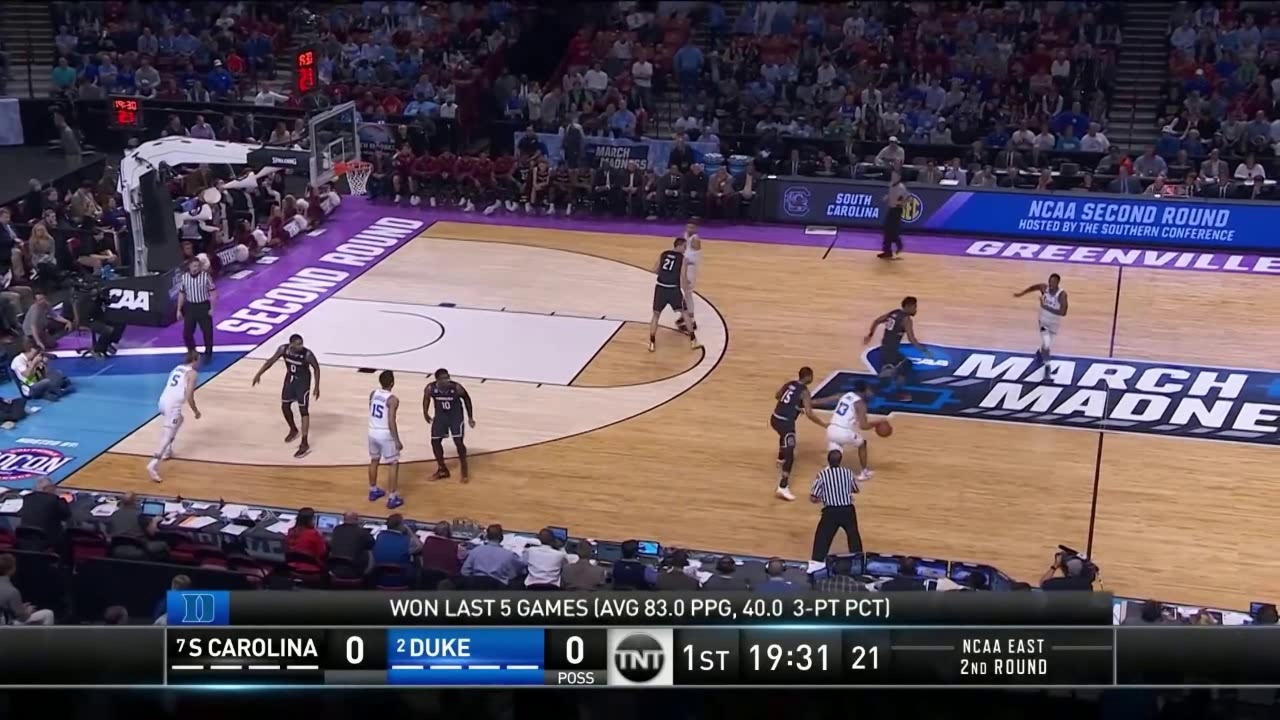} & \includegraphics[width=0.15\textwidth]{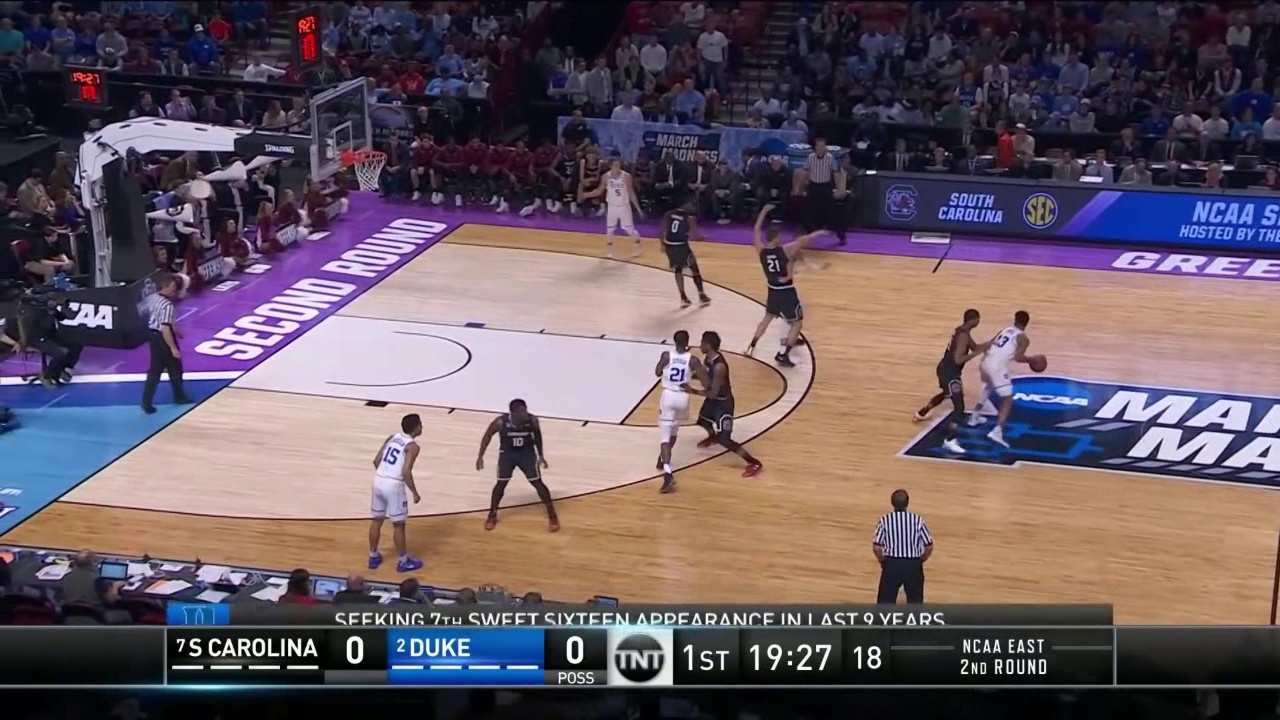}  \\   
        \makecell{Top-1\\(TrajSV)} & \includegraphics[width=0.15\textwidth]{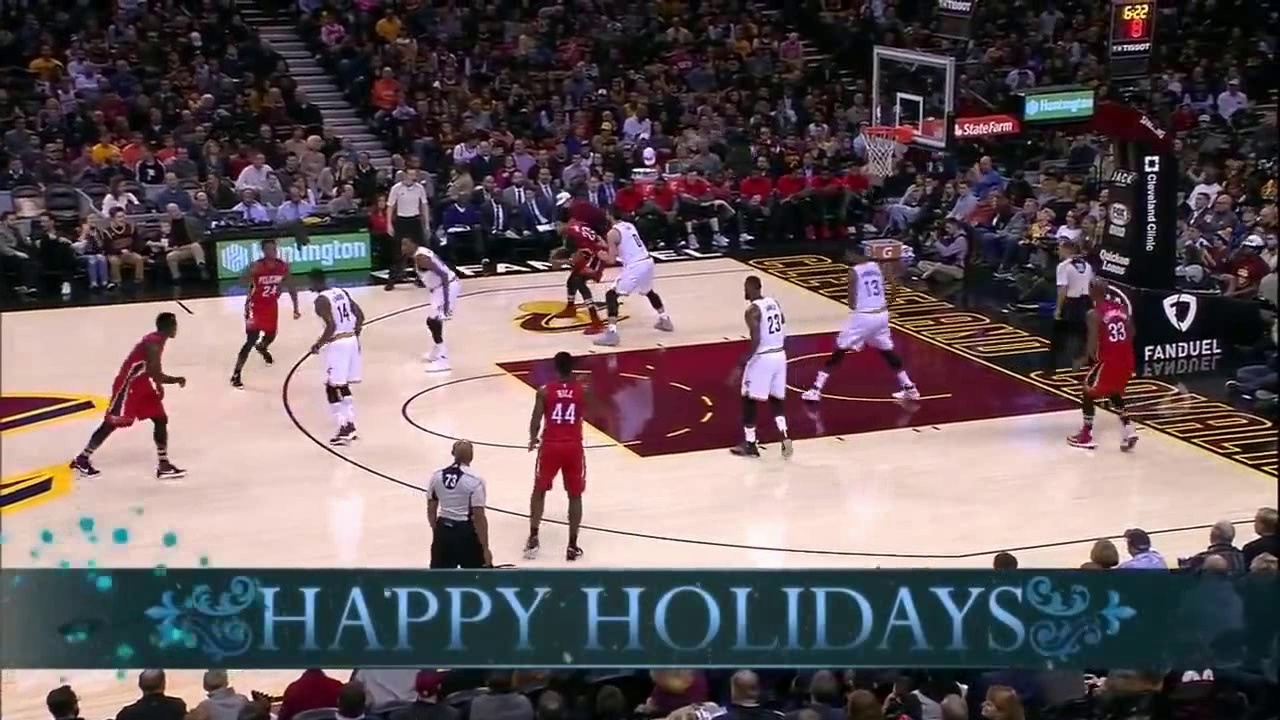} & \includegraphics[width=0.15\textwidth]{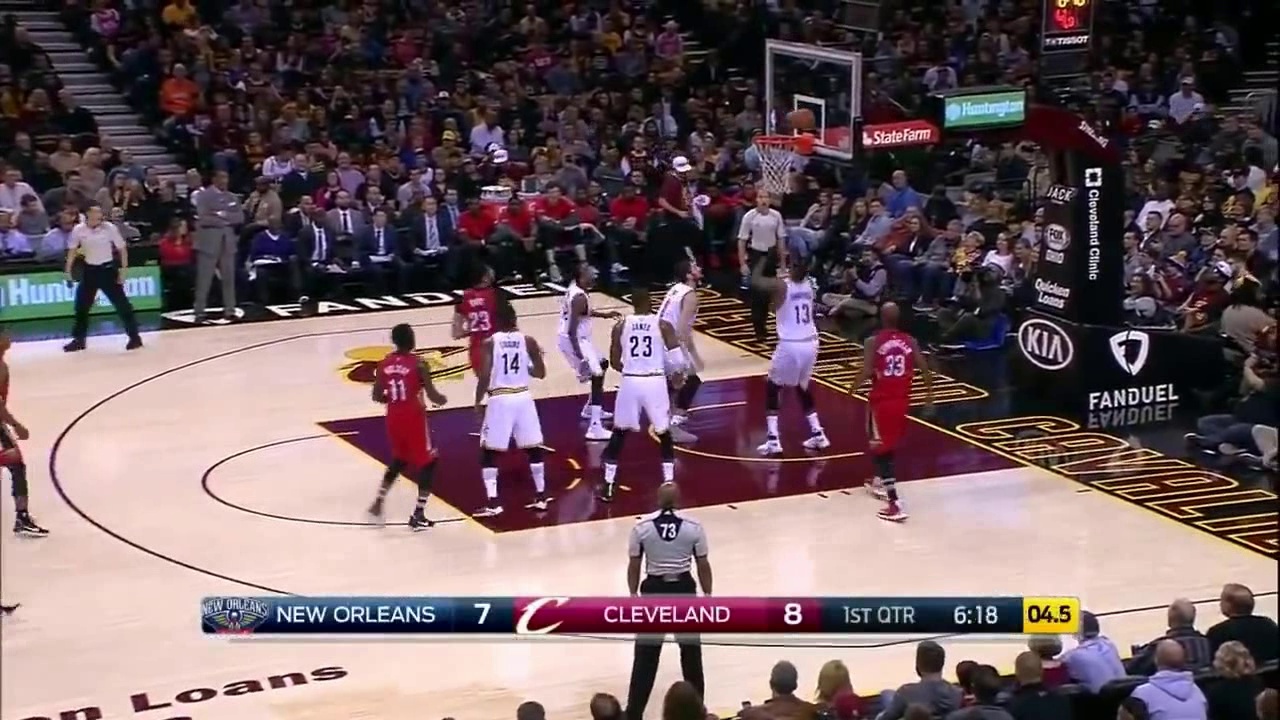} & \includegraphics[width=0.15\textwidth]{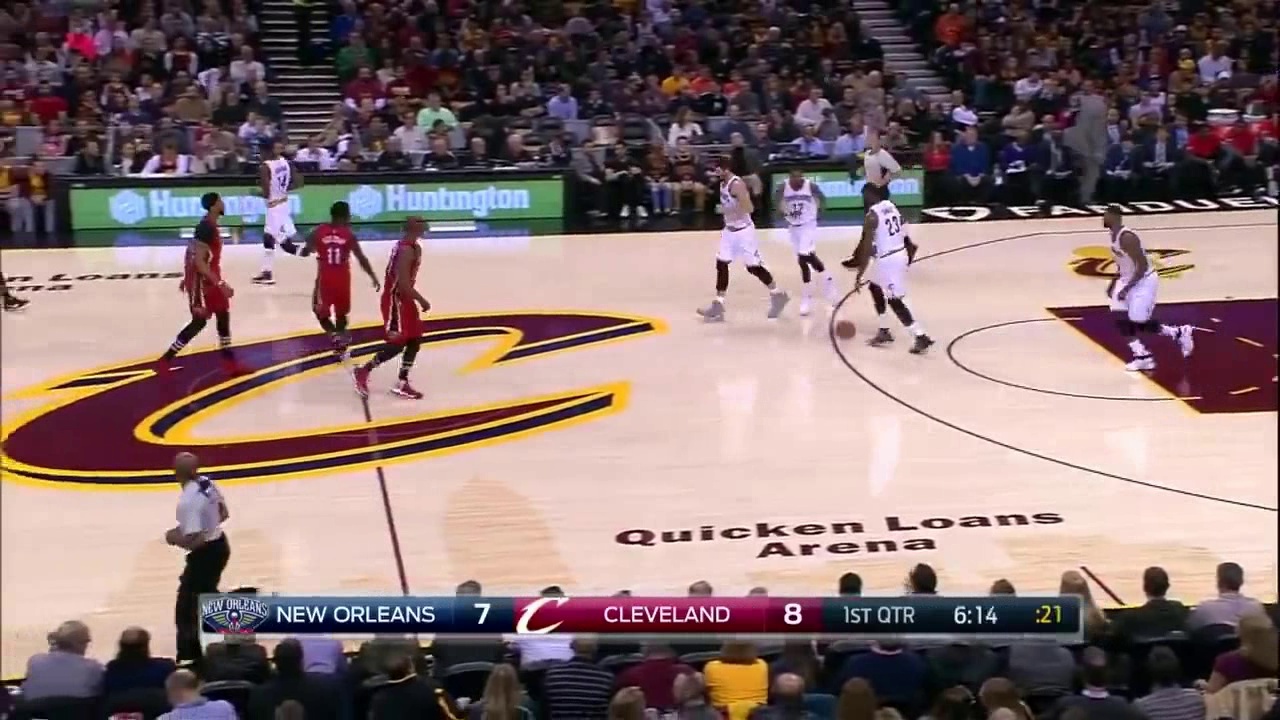} & \includegraphics[width=0.15\textwidth]{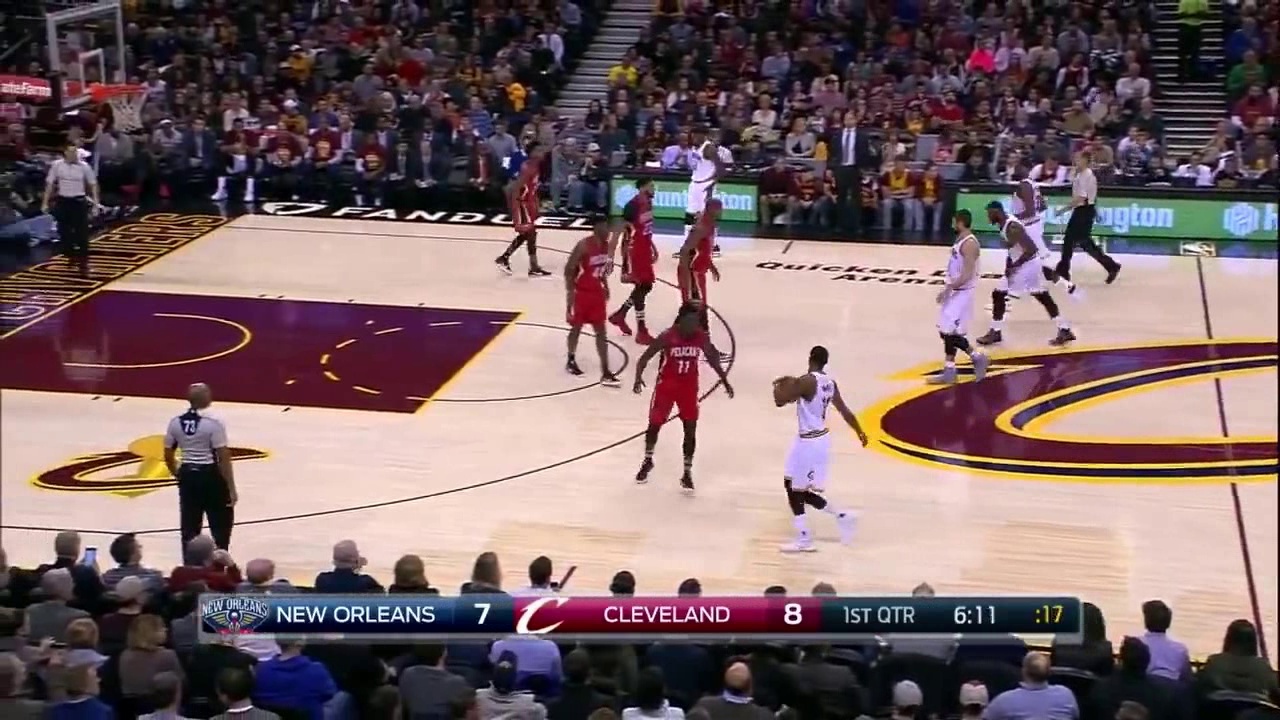} & \includegraphics[width=0.15\textwidth]{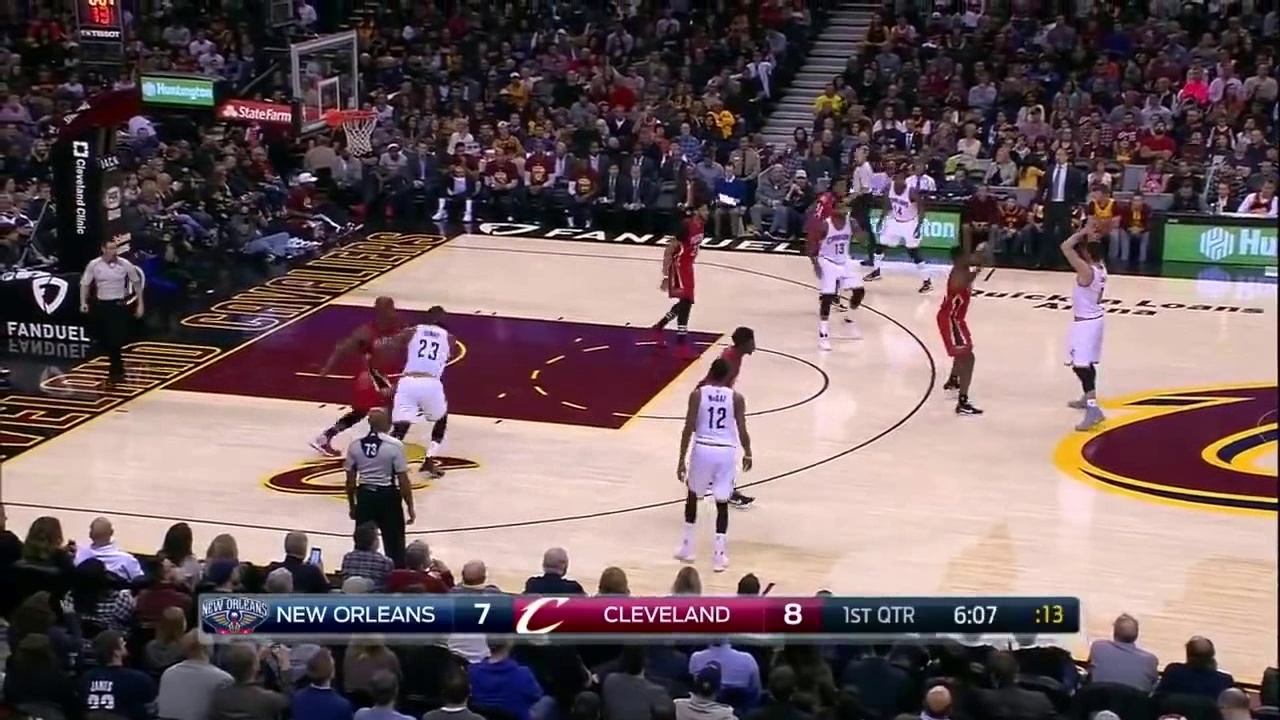}   \\  
        \hline
\end{tabular}
%\vspace{-4mm}
\caption{{\tcsvt{Qualitative results of sports video retrieval, illustrating improvements from incorporating trajectory information in corner and direct
free-kick queries.}}}
\vspace{-4mm}
\label{fig:case_study_retrieval}
\end{figure*}

\begin{figure*}[t]
  \centering
  \footnotesize
  \setlength{\tabcolsep}{1pt}
  \begin{tabular}{m{\textwidth}}
  \includegraphics[width=\textwidth]{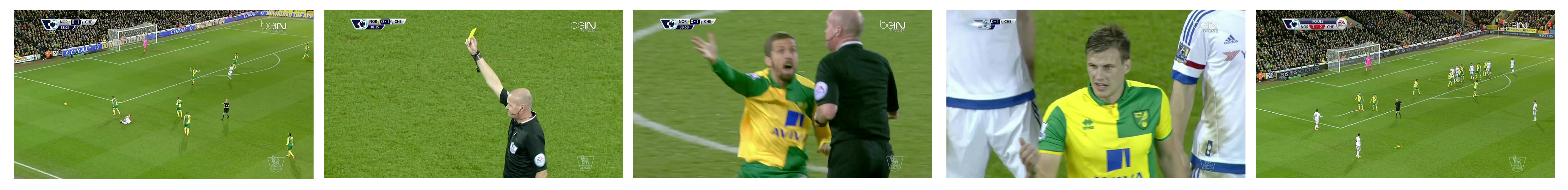} \\
    \textbf{(1) Ground truth} $\{$38:40$\}$[1.0]: PLAYER TEAM is awarded a yellow card for his tackle. He doesn't seem to agree with the decision but REFEREE ignores the protests.
    
    \textbf{(2) Baidu-VC} $\{$38:36$\}$[0.0]: PLAYER TEAM is booked after bringing down an opponent REFEREE had an easy decision to make.
    
    \textbf{(3) Baidu-VC+play2vec} $\{$38:31$\}$[0.0]: PLAYER TEAM is booked after bringing down an opponent REFEREE had an easy decision to make.
    
    \textbf{(4) Baidu-VC+TrajSV} $\{$38:36$\}$[1.0]: PLAYER TEAM is awarded a yellow card for his tackle. He doesn't seem to agree with the decision but REFEREE ignores the protests.
  \end{tabular}
  %\vspace{-4mm}
  \caption{Qualitative results of SDVC, where the spotting time and BLEU@4 are shown in $\{$$\}$ and [], respectively.}\label{fig:case_study}
  \vspace*{-4mm}
\end{figure*}

\begin{table}[t]
\centering
\caption{{\tcsvtminor{Comparison with general video models on SoccerNet.}}}
\vspace{-2mm}
\begin{tabular}{lll}
\hline
Components       & HR@1 & MRR \\ \hline
TrajSV  &   \textbf{0.250}  &  \textbf{0.551}  \\ 
w/o Trajectory  &   0.208    &   0.489  \\
StreamMind & 0.223      &0.514  \\ \hline
\end{tabular}
\label{tab:sota}
\vspace*{-6mm}
\end{table}

\begin{table*}[]
\centering
\caption{{\tcsvtminor{Effectiveness across diverse sports scenarios on SoccerNet.}}}
\vspace*{-2mm}
\label{tab:scenarios}
\begin{tabular}{c|ccccc|ccc|ccc}
\hline
\multirow{2}{*}{Method} & \multicolumn{5}{c|}{Concrete Events}                                               & \multicolumn{3}{c|}{Abstract Patterns}           & \multicolumn{3}{c}{Variable-length Actions}      \\ \cline{2-12} 
                        & Free-kick      & Corner         & Goal           & Penalty        & Overall        & Throw-in       & Kick-off       & Overall        & Second-level   & Minute-level   & Overall        \\ \hline
X-CLIP (MLP)            & 0.392          & 0.312          & 0.461          & 0.402          & 0.417          & 0.327          & 0.254          & 0.314          & 0.363          & 0.225          & 0.280          \\
TrajSV                  & \textbf{0.538} & \textbf{0.521} & \textbf{0.611} & \textbf{0.572} & \textbf{0.549} & \textbf{0.615} & \textbf{0.509} & \textbf{0.538} & \textbf{0.633} & \textbf{0.577} & \textbf{0.612} \\ \hline
\end{tabular}
\vspace*{-5mm}
\end{table*}

\smallskip
\noindent\textbf{(5) Parameter Study (batch size, cell size, and embedding dimension).}
%Retrieval effectiveness and training time with different batch sizes.
%(a) Retrieval effectiveness
%(b) Training time
%We investigate the impact of varying batch size, cell size, and the dimension of trajectory representations on the retrieval effectiveness of the YouTube dataset. %We fix other parameters (except for the parameter that is testing) to the value mentioned in \Cref{sec:setup}. 
{\indshihao We investigate the impact of varying (1) batch size, (2) cell size, and (3) the dimension of trajectory representations for sports video retrieval. The results are presented in \cref{tab:paramter_study_batch_size}, \cref{tab:paramter_study_cell_size}, and \cref{tab:paramter_embedding}, respectively.} For (1), the best results are achieved with a batch size of 128. This aligns with the expectation that a larger batch size typically improves performance in contrastive learning by incorporating more negative samples. For (2), the best results are achieved with a cell size of 3 meters, which balances the risk of multiple trajectories being encoded to the same representation (if the cell size is too large) and the risk of representing similar trajectories differently (if the cell size is too small). For (3),  we find that increasing the dimension of trajectory representations improves retrieval performance. When the embedding dimension reaches 256, the retrieval performance HR@1 is 0.564. It suggests that a higher dimensional representation tends to preserve more information, which often results in better performance.

\if 0
We observe that the best results are obtained when using a batch size of 128 (HR@1=0.5), which is in line with the expectation that a larger batch size usually leads to a better performance in contrastive learning as it consists of more negative samples. {\indshihao At the same time, larger batch sizes also lead to faster training. However, due to GPU memory limitations, the maximum batch size is set to 128.} %{\xshrebuttal However, the retrieval performance seems not sensitive to the batch size, since the performance with a batch size of 64 is worse than the performance with a batch size of 32.} It may be because the number of videos in the Youtube dataset is still small. 
%
Besides, we find that using a cell size of 3 meters produces the best results. This is because if the cell size value is set too large, it can result in multiple trajectories being encoded to the same representation, which can hinder accurate retrieval. Conversely, if the cell size is too small, similar trajectories may be represented differently, which can also negatively impact retrieval performance. Therefore, setting the cell size to 3 meters provides an optimal balance between these two issues, leading to improved results.
\fi

\smallskip
{\tcsvt{\noindent\textbf{(6) Transferability Study and of Generalization Discussion.} 
% Transferability test (SoccerNet training and YouTube testing).
%(a) Retrieval effectiveness 
%(b) Training time of finetuning
We evaluate the transferability of TrajSV for sports video retrieval on YouTube using SoccerNet as the training data. The transferability is studied in two ways: 1) training the model on SoccerNet and directly testing on YouTube (zero-shot), and 2) training on SoccerNet, fine-tuning using the YouTube training set, and further testing on YouTube. For comparison, we present results using ResNet features, and both training and testing on the YouTube dataset. As shown in \cref{tab:Transferability}, fine-tuning outperforms zero-shot for both TrajSV and ResNet. Fine-tuning with TrajSV is superior to ResNet, but zero-shot transfer lacks the same improvement, potentially attributed to differences in trajectory distributions between the two datasets. Moreover, fine-tuning performs worse than direct training on YouTube, likely due to differences in video length and content.

We further discuss the generalization capability of TrajSV from two perspectives. (1) Cross-Dataset Transferability: TrajSV can be easily transferred between datasets for the same sport (e.g., soccer). As shown in \cref{tab:Transferability}, it outperforms state-of-the-art methods (e.g., ResNet) and can achieve further improvements with additional training. (2) Cross-Sport Adaptability: The design of TrajSV is adaptable to other sports (e.g., basketball and volleyball), as demonstrated in \cref{tab:retrieval_all_datasets}. This suggests that trajectory features provide a shared foundation across different sports videos, as they inherently capture both spatial and temporal information.
}}

\if 0
\SH{
To investigate the transferability of our methods, we evaluate the effectiveness of sports video retrieval on the YouTube dataset using the SoccerNet dataset as the training data (as shown in \cref{tab:Transferability}). Due to the SoccerNet dataset containing a much larger number of trajectories than the YouTube dataset, we study transferability in two ways: 1) we train the model on SoccerNet and directly test the retrieval performance on the YouTube dataset by searching the test set against the entire YouTube corpus, named zero-shot, and 2) we train the model on SoccerNet, fine-tuned it using the YouTube training set, and further test it on the YouTube dataset. For comparison, we also present two other results: 1) the retrieval results using ResNet features and 2) both training and testing are performed on the YouTube dataset. We set the number of clips contained in the video to 64 ($n=64$), and the other parameters are introduced in \cref{sec:setup}. 
%
The results show that fine-tuning performs better than zero-shot for both clip-level and ResNet features. Besides, fine-tuning with TrajSV is superior to ResNet features, but zero-shot transfer does not show the same improvement. Additionally, fine-tuning results are not better than directly training on the YouTube dataset, possibly due to differences in video length and content between the datasets affecting model performance.
}
\fi

\smallskip
\noindent{\tcsvt{\textbf{(7) Scalability Study.} We conduct a scalability test based on the setup for sports video retrieval, where we vary the database size from 500 to 3,000 videos. We extract representations for all videos in the database and use the HNSW index to manage these representations. As shown in \cref{tab:scalability}, TrajSV demonstrates good scalability, which aligns with the expected vector search scalability complexity. It is important to note that the main computational complexity comes from embedding the given query sports videos, which involves necessary attention operations. This embedding occurs only once per query video and can then be used for search across a large database. Additionally, the embedding supports other analytical tasks, such as action spotting and video captioning.
}}

\smallskip
\noindent{\Comment \textbf{(8) Qualitative Results.} 
To qualitatively validate the effectiveness, we present some typical cases of (1) sports video retrieval and (2) video captioning in \cref{fig:case_study_retrieval} and \cref{fig:case_study}, respectively. %Overall, we observe that the representations based on visual and trajectory signals can successfully retrieve similar video clips, predict video caption timing, and aid in generating captions for sports videos. 
For (1), we query a video clip, and then return the Top-1 retrieved clip with TrajSV. We note that the retrieved clip generally aligns with the query, featuring same sports actions such as corners or direct free-kicks. This observation extends to volleyball and basketball as well. {\tcsvt{Additionally, TrajSV incorporates trajectory (temporal) information alongside the visual features captured by X-CLIP~\cite{DBLP:conf/eccv/NiPCZMFXL22}. To further explore the contribution of trajectory information, we illustrate the Top-1 retrieved clip using X-CLIP for corner and direct free-kick queries in \cref{fig:case_study_retrieval}. We observe that the clips returned by TrajSV better align with the player movement patterns in the queries, suggesting that trajectory information is crucial for capturing similar movement patterns that visual features alone may not adequately represent.
}}
%and the extracted trajectories align well with the areas of sports actions. 
%
For (2), we present a case study of the SDVC, where we present the ground truth and the captions generated by the Baidu-VC, Baidu-VC + play2vec embeddings, and Baidu-VC + TrajSV embeddings. We observe that Baidu-VC + TrajSV achieves more accurate captions in content and timing. However, both the Baidu-VC and Baidu-VC + play2vec produce a different caption, possibly from previous time steps.
%the ground truth describes a player receiving a yellow card, with the referee ignoring the protests.

\smallskip
\noindent{{\tcsvtminor{\textbf{(9) Comparison with General Video Models.} We compare recent state-of-the-art general video models to justify the necessity of the trajectory-specific design in TrajSV for sports analytics. In particular, we adopt StreamMind~\cite{ding2025streammind}, a recent LLM-based model trained on various video understanding tasks. Notably, the SoccerNet dataset~\cite{DBLP:conf/cvpr/DeliegeCGSDNGMD21} is included in its pre-training. To represent a video, we average the frame-level embeddings produced by StreamMind. As shown in Table~\ref{tab:sota}, TrajSV outperforms StreamMind by 10.8\% in HR@1 and 6.7\% in MRR. This gain is largely attributed to the incorporation of trajectory features. When these features are excluded, the performance of TrajSV drops—by 16.8\% in HR@1 and 11.3\% in MRR. We also observe that StreamMind slightly outperforms TrajSV without trajectory features, likely due to its large-scale LLM-based pre-training.
}}

\smallskip
\noindent{{\tcsvtminor{\textbf{(10) Model Generalization Across Diverse Sports Scenarios.} We present a detailed breakdown of performance across various sports scenarios on the SoccerNet dataset~\cite{DBLP:conf/cvpr/DeliegeCGSDNGMD21}, which provides rich annotations to support scenario-specific analysis. As shown in Table~\ref{tab:scenarios}, we report results in terms of MRR with a noise rate of $\delta=0.6$, covering three categories: 1) Concrete events, including free kicks (both indirect and direct), corners, goals, and penalties. 2) Abstract patterns, involving transition-related events such as throw-ins and kick-offs. 3) Variable-length actions, comprising short-term actions like clearance (typically lasting several seconds) and long-term actions like substitution (which can span several minutes).
For reference, we also include the performance of X-CLIP (MLP), the strongest baseline on SoccerNet. We observe that TrajSV consistently outperforms X-CLIP (MLP) by approximately 39.98\% across diverse sports scenarios. This performance gain is consistent with the overall results reported in Table~\ref{tab:retrieval_all_datasets}, and is primarily attributed to TrajSV’s ability to capture fine-grained trajectory patterns that are critical for identifying similar videos in retrieval tasks. This is further illustrated in the case study of the corner event in Figure~\ref{fig:case_study_retrieval}, where the player movement patterns retrieved by TrajSV closely resemble those in the query video.
}}
\section{CONCLUSIONS}

This paper proposes TrajSV, a trajectory-based framework for learning sports video representations. The TrajSV framework consists of three components: data preprocessing, CRNet, and VRNet. It is built on raw broadcast sports videos, making it widely applicable, and derives both video-level and clip-level representations to support various applications. Our experiments demonstrate its superior performance in sports video retrieval, action spotting, and video captioning tasks.
In the future, we plan to explore more sports analytics tasks, such as player performance analysis and team tactics discovery, to further demonstrate the effectiveness of TrajSV.

%\balance
\bibliographystyle{IEEEtran}
\bibliography{ref}

\begin{thebibliography}{10}
\providecommand{\url}[1]{#1}
\csname url@samestyle\endcsname
\providecommand{\newblock}{\relax}
\providecommand{\bibinfo}[2]{#2}
\providecommand{\BIBentrySTDinterwordspacing}{\spaceskip=0pt\relax}
\providecommand{\BIBentryALTinterwordstretchfactor}{4}
\providecommand{\BIBentryALTinterwordspacing}{\spaceskip=\fontdimen2\font plus
\BIBentryALTinterwordstretchfactor\fontdimen3\font minus \fontdimen4\font\relax}
\providecommand{\BIBforeignlanguage}[2]{{%
\expandafter\ifx\csname l@#1\endcsname\relax
\typeout{** WARNING: IEEEtran.bst: No hyphenation pattern has been}%
\typeout{** loaded for the language `#1'. Using the pattern for}%
\typeout{** the default language instead.}%
\else
\language=\csname l@#1\endcsname
\fi
#2}}
\providecommand{\BIBdecl}{\relax}
\BIBdecl

\bibitem{DBLP:conf/kdd/WangLCJ19}
Z.~Wang, C.~Long, G.~Cong, and C.~Ju, ``Effective and efficient sports play retrieval with deep representation learning,'' in \emph{{KDD}}.\hskip 1em plus 0.5em minus 0.4em\relax {ACM}, 2019, pp. 499--509.

\bibitem{DBLP:conf/iui/ShaLYCRM16}
L.~Sha, P.~Lucey, Y.~Yue, P.~Carr, C.~Rohlf, and I.~A. Matthews, ``Chalkboarding: {A} new spatiotemporal query paradigm for sports play retrieval,'' in \emph{{IUI}}.\hskip 1em plus 0.5em minus 0.4em\relax {ACM}, 2016, pp. 336--347.

\bibitem{DBLP:conf/sdm/ZhangWLY22}
Q.~Zhang, Z.~Wang, C.~Long, and S.~Yiu, ``On predicting and generating a good break shot in billiards sports,'' in \emph{{SDM}}.\hskip 1em plus 0.5em minus 0.4em\relax {SIAM}, 2022, pp. 109--117.

\bibitem{DBLP:conf/kdd/DecroosHD18}
T.~Decroos, J.~V. Haaren, and J.~Davis, ``Automatic discovery of tactics in spatio-temporal soccer match data,'' in \emph{{KDD}}.\hskip 1em plus 0.5em minus 0.4em\relax {ACM}, 2018, pp. 223--232.

\bibitem{DBLP:journals/tkdd/DiKSL18}
M.~Di, D.~Klabjan, L.~Sha, and P.~Lucey, ``Large-scale adversarial sports play retrieval with learning to rank,'' \emph{{ACM} Trans. Knowl. Discov. Data}, vol.~12, no.~6, pp. 69:1--69:18, 2018.

\bibitem{DBLP:conf/mmasia/HaruyamaTOH20}
T.~Haruyama, S.~Takahashi, T.~Ogawa, and M.~Haseyama, ``Similar scene retrieval in soccer videos with weak annotations by multimodal use of bidirectional {LSTM},'' in \emph{MMAsia}.\hskip 1em plus 0.5em minus 0.4em\relax {ACM}, 2020, pp. 27:1--27:8.

\bibitem{DBLP:conf/bigdataconf/ProbstRSSR18}
L.~Probst, F.~Rauschenbach, H.~Schuldt, P.~Seidenschwarz, and M.~Rumo, ``Integrated real-time data stream analysis and sketch-based video retrieval in team sports,'' in \emph{{IEEE} BigData}.\hskip 1em plus 0.5em minus 0.4em\relax {IEEE}, 2018, pp. 548--555.

\bibitem{DBLP:conf/cikm/KabaryS13}
I.~A. Kabary and H.~Schuldt, ``Sportsense: using motion queries to find scenes in sports videos,'' in \emph{{CIKM}}.\hskip 1em plus 0.5em minus 0.4em\relax {ACM}, 2013, pp. 2489--2492.

\bibitem{DBLP:conf/cvpr/CioppaDGGDGM20}
A.~Cioppa, A.~Deli{\`{e}}ge, S.~Giancola, B.~Ghanem, M.~V. Droogenbroeck, R.~Gade, and T.~B. Moeslund, ``A context-aware loss function for action spotting in soccer videos,'' in \emph{{CVPR}}.\hskip 1em plus 0.5em minus 0.4em\relax Computer Vision Foundation / {IEEE}, 2020, pp. 13\,123--13\,133.

\bibitem{DBLP:conf/mm/HaruyamaTOH19}
T.~Haruyama, S.~Takahashi, T.~Ogawa, and M.~Haseyama, ``Retrieval of similar scenes based on multimodal distance metric learning in soccer videos,'' in \emph{MMSports@MM}.\hskip 1em plus 0.5em minus 0.4em\relax {ACM}, 2019, pp. 10--15.

\bibitem{kong2019joint}
L.~Kong, D.~Huang, J.~Qin, and Y.~Wang, ``A joint framework for athlete tracking and action recognition in sports videos,'' \emph{IEEE TCSVT}, vol.~30, no.~2, pp. 532--548, 2019.

\bibitem{Mkhallati2023SoccerNetCaption}
H.~Mkhallati, A.~Cioppa, S.~Giancola, B.~Ghanem, and M.~Van~Droogenbroeck, ``{SoccerNet}-{C}aption: Dense video captioning for soccer broadcasts commentaries,'' in \emph{CVPRW, CVsports}, 2023.

\bibitem{qi2019sports}
M.~Qi, Y.~Wang, A.~Li, and J.~Luo, ``Sports video captioning via attentive motion representation and group relationship modeling,'' \emph{IEEE TCSVT}, vol.~30, no.~8, pp. 2617--2633, 2019.

\bibitem{DBLP:journals/tkde/WangLC23}
Z.~Wang, C.~Long, and G.~Cong, ``Similar sports play retrieval with deep reinforcement learning,'' \emph{{IEEE} Trans. Knowl. Data Eng.}, vol.~35, no.~4, pp. 4253--4266, 2023.

\bibitem{datasource}
``Stats perform,'' \url{https://www.stats.com/artificial-intelligence}.

\bibitem{DBLP:journals/corr/abs-2203-12602}
Z.~Tong, Y.~Song, J.~Wang, and L.~Wang, ``Videomae: Masked autoencoders are data-efficient learners for self-supervised video pre-training,'' \emph{CoRR}, vol. abs/2203.12602, 2022.

\bibitem{DBLP:conf/eccv/NiPCZMFXL22}
B.~Ni, H.~Peng, M.~Chen, S.~Zhang, G.~Meng, J.~Fu, S.~Xiang, and H.~Ling, ``Expanding language-image pretrained models for general video recognition,'' in \emph{ECCV}, 2022, pp. 1--18.

\bibitem{zhang2025billiards}
Q.~Zhang, Z.~Wang, C.~Long, and S.-M. Yiu, ``Billiards sports analytics: Datasets and tasks,'' \emph{TKDD}, 2025.

\bibitem{xu2025sportstraj}
Y.~Xu and Y.~Fu, ``Sports-traj: A unified trajectory generation model for multi-agent movement in sports,'' in \emph{ICLR}, 2025.

\bibitem{gu2023mamba}
A.~Gu and T.~Dao, ``Mamba: Linear-time sequence modeling with selective state spaces,'' \emph{arXiv preprint arXiv:2312.00752}, 2023.

\bibitem{DBLP:conf/sigir/KabaryS14}
I.~A. Kabary and H.~Schuldt, ``Enhancing sketch-based sport video retrieval by suggesting relevant motion paths,'' in \emph{{SIGIR}}.\hskip 1em plus 0.5em minus 0.4em\relax {ACM}, 2014, pp. 1227--1230.

\bibitem{DBLP:conf/cvpr/GiancolaG21}
S.~Giancola and B.~Ghanem, ``Temporally-aware feature pooling for action spotting in soccer broadcasts,'' in \emph{{CVPR} Workshops}.\hskip 1em plus 0.5em minus 0.4em\relax Computer Vision Foundation / {IEEE}, 2021, pp. 4490--4499.

\bibitem{DBLP:conf/icpr/TomeiBCBC20}
M.~Tomei, L.~Baraldi, S.~Calderara, S.~Bronzin, and R.~Cucchiara, ``Rms-net: Regression and masking for soccer event spotting,'' in \emph{{ICPR}}.\hskip 1em plus 0.5em minus 0.4em\relax {IEEE}, 2020, pp. 7699--7706.

\bibitem{DBLP:journals/corr/abs-2106-14447}
X.~Zhou, L.~Kang, Z.~Cheng, B.~He, and J.~Xin, ``Feature combination meets attention: Baidu soccer embeddings and transformer based temporal detection,'' \emph{CoRR}, vol. abs/2106.14447, 2021.

\bibitem{DBLP:journals/tmm/WuWBDLCDX23}
F.~Wu, Q.~Wang, J.~Bian, N.~Ding, F.~Lu, J.~Cheng, D.~Dou, and H.~Xiong, ``A survey on video action recognition in sports: Datasets, methods and applications,'' \emph{{IEEE} Trans. Multim.}, vol.~25, pp. 7943--7966, 2023.

\bibitem{li2024contextualized}
J.~Li, L.~Zhang, Q.~Wu, Z.~Qi, H.~Lu, M.~Wang, and D.~Tao, ``Contextualized relation predictive model for self-supervised group activity representation learning,'' \emph{IEEE TMM}, vol.~26, pp. 354--368, 2024.

\bibitem{wang2024knowledge}
Z.~Wang, Z.~Li, X.~Lang, Y.~Zheng, M.~Tian, L.~Wu, L.~Wang, and C.~Chen, ``Knowledge augmented relation inference for group activity recognition,'' \emph{IEEE TCSVT}, pp. 11\,645--11\,658, 2024.

\bibitem{DBLP:conf/icmcs/HuHWL07}
Y.~Hu, B.~Han, G.~Wang, and X.~Lin, ``Enhanced shot change detection using motion features for soccer video analysis,'' in \emph{{ICME}}.\hskip 1em plus 0.5em minus 0.4em\relax {IEEE} Computer Society, 2007, pp. 1555--1558.

\bibitem{DBLP:conf/icassp/WangCX05}
J.~Wang, E.~Chng, and C.~Xu, ``Soccer replay detection using scene transition structure analysis,'' in \emph{{ICASSP} {(2)}}.\hskip 1em plus 0.5em minus 0.4em\relax {IEEE}, 2005, pp. 433--436.

\bibitem{DBLP:conf/wacv/TheinerE23}
J.~Theiner and R.~Ewerth, ``Tvcalib: Camera calibration for sports field registration in soccer,'' in \emph{{WACV}}.\hskip 1em plus 0.5em minus 0.4em\relax {IEEE}, 2023, pp. 1166--1175.

\bibitem{DBLP:conf/spieSR/FarinKWE04}
D.~Farin, S.~Krabbe, P.~H.~N. de~With, and W.~Effelsberg, ``Robust camera calibration for sport videos using court models,'' in \emph{Storage and Retrieval Methods and Applications for Multimedia}, ser. {SPIE} Proceedings, vol. 5307.\hskip 1em plus 0.5em minus 0.4em\relax {SPIE}, 2004, pp. 80--91.

\bibitem{DBLP:journals/pami/LuTLM13}
W.~Lu, J.~Ting, J.~J. Little, and K.~P. Murphy, ``Learning to track and identify players from broadcast sports videos,'' \emph{{IEEE} Trans. Pattern Anal. Mach. Intell.}, vol.~35, no.~7, pp. 1704--1716, 2013.

\bibitem{DBLP:conf/eccv/SullivanC06}
J.~Sullivan and S.~Carlsson, ``Tracking and labelling of interacting multiple targets,'' in \emph{{ECCV} {(3)}}, ser. Lecture Notes in Computer Science, vol. 3953.\hskip 1em plus 0.5em minus 0.4em\relax Springer, 2006, pp. 619--632.

\bibitem{kim2022cost}
H.~Kim, C.~J. Kim, M.~Jeong, J.~Lee, J.~Yoon, and S.-K. Ko, ``Cost-efficient and bias-robust sports player tracking by integrating gps and video,'' in \emph{International Workshop on Machine Learning and Data Mining for Sports Analytics}.\hskip 1em plus 0.5em minus 0.4em\relax Springer, 2022, pp. 74--86.

\bibitem{DBLP:journals/tmm/Tejero-de-Pablos18}
A.~Tejero{-}de{-}Pablos, Y.~Nakashima, T.~Sato, N.~Yokoya, M.~Linna, and E.~Rahtu, ``Summarization of user-generated sports video by using deep action recognition features,'' \emph{{IEEE} Trans. Multim.}, vol.~20, no.~8, pp. 2000--2011, 2018.

\bibitem{DBLP:conf/cvpr/DeliegeCGSDNGMD21}
A.~Deli{\`{e}}ge, A.~Cioppa, S.~Giancola, M.~J. Seikavandi, J.~V. Dueholm, K.~Nasrollahi, B.~Ghanem, T.~B. Moeslund, and M.~V. Droogenbroeck, ``Soccernet-v2: {A} dataset and benchmarks for holistic understanding of broadcast soccer videos,'' in \emph{{CVPR} Workshops}.\hskip 1em plus 0.5em minus 0.4em\relax Computer Vision Foundation / {IEEE}, 2021, pp. 4508--4519.

\bibitem{DBLP:conf/cvpr/GiancolaADG18}
S.~Giancola, M.~Amine, T.~Dghaily, and B.~Ghanem, ``Soccernet: {A} scalable dataset for action spotting in soccer videos,'' in \emph{{CVPR} Workshops}.\hskip 1em plus 0.5em minus 0.4em\relax Computer Vision Foundation / {IEEE} Computer Society, 2018, pp. 1711--1721.

\bibitem{shih2017survey}
H.-C. Shih, ``A survey of content-aware video analysis for sports,'' \emph{IEEE Transactions on circuits and systems for video technology}, vol.~28, no.~5, pp. 1212--1231, 2017.

\bibitem{han2019video}
T.~Han, W.~Xie, and A.~Zisserman, ``Video representation learning by dense predictive coding,'' in \emph{Proceedings of the IEEE/CVF International Conference on Computer Vision Workshops}, 2019, pp. 0--0.

\bibitem{sun2019videobert}
C.~Sun, A.~Myers, C.~Vondrick, K.~Murphy, and C.~Schmid, ``Videobert: A joint model for video and language representation learning,'' in \emph{Proceedings of the IEEE/CVF international conference on computer vision}, 2019, pp. 7464--7473.

\bibitem{piergiovanni2020evolving}
A.~Piergiovanni, A.~Angelova, and M.~S. Ryoo, ``Evolving losses for unsupervised video representation learning,'' in \emph{Proceedings of the IEEE/CVF Conference on Computer Vision and Pattern Recognition}, 2020, pp. 133--142.

\bibitem{DBLP:conf/iccv/YangBG21}
J.~Yang, Y.~Bisk, and J.~Gao, ``Taco: Token-aware cascade contrastive learning for video-text alignment,'' in \emph{{ICCV}}.\hskip 1em plus 0.5em minus 0.4em\relax {IEEE}, 2021, pp. 11\,542--11\,552.

\bibitem{DBLP:journals/ijon/LuoJZCLDL22}
H.~Luo, L.~Ji, M.~Zhong, Y.~Chen, W.~Lei, N.~Duan, and T.~Li, ``Clip4clip: An empirical study of {CLIP} for end to end video clip retrieval and captioning,'' \emph{Neurocomputing}, vol. 508, pp. 293--304, 2022.

\bibitem{DBLP:conf/cvpr/0005XZWGHH22}
S.~Guo, Z.~Xiong, Y.~Zhong, L.~Wang, X.~Guo, B.~Han, and W.~Huang, ``Cross-architecture self-supervised video representation learning,'' in \emph{{CVPR}}.\hskip 1em plus 0.5em minus 0.4em\relax {IEEE}, 2022, pp. 19\,248--19\,257.

\bibitem{DBLP:conf/cvpr/QianMG0WBC21}
R.~Qian, T.~Meng, B.~Gong, M.~Yang, H.~Wang, S.~J. Belongie, and Y.~Cui, ``Spatiotemporal contrastive video representation learning,'' in \emph{{CVPR}}.\hskip 1em plus 0.5em minus 0.4em\relax Computer Vision Foundation / {IEEE}, 2021, pp. 6964--6974.

\bibitem{DBLP:journals/corr/ChenPSA17}
L.~Chen, G.~Papandreou, F.~Schroff, and H.~Adam, ``Rethinking atrous convolution for semantic image segmentation,'' \emph{CoRR}, vol. abs/1706.05587, 2017.

\bibitem{DBLP:journals/ijcv/ZhangWWZL21}
Y.~Zhang, C.~Wang, X.~Wang, W.~Zeng, and W.~Liu, ``Fairmot: On the fairness of detection and re-identification in multiple object tracking,'' \emph{Int. J. Comput. Vis.}, vol. 129, no.~11, pp. 3069--3087, 2021.

\bibitem{DBLP:journals/corr/abs-1904-07850}
X.~Zhou, D.~Wang, and P.~Kr{\"{a}}henb{\"{u}}hl, ``Objects as points,'' \emph{CoRR}, vol. abs/1904.07850, 2019.

\bibitem{DBLP:conf/naacl/DevlinCLT19}
J.~Devlin, M.~Chang, K.~Lee, and K.~Toutanova, ``{BERT:} pre-training of deep bidirectional transformers for language understanding,'' in \emph{{NAACL-HLT} {(1)}}.\hskip 1em plus 0.5em minus 0.4em\relax Association for Computational Linguistics, 2019, pp. 4171--4186.

\bibitem{DBLP:conf/nips/VaswaniSPUJGKP17}
A.~Vaswani, N.~Shazeer, N.~Parmar, J.~Uszkoreit, L.~Jones, A.~N. Gomez, L.~Kaiser, and I.~Polosukhin, ``Attention is all you need,'' in \emph{{NIPS}}, 2017, pp. 5998--6008.

\bibitem{DBLP:journals/corr/BaKH16}
L.~J. Ba, J.~R. Kiros, and G.~E. Hinton, ``Layer normalization,'' \emph{CoRR}, vol. abs/1607.06450, 2016.

\bibitem{DBLP:journals/pami/MalkovY20}
Y.~A. Malkov and D.~A. Yashunin, ``Efficient and robust approximate nearest neighbor search using hierarchical navigable small world graphs,'' \emph{{IEEE} Trans. Pattern Anal. Mach. Intell.}, vol.~42, no.~4, pp. 824--836, 2020.

\bibitem{DBLP:books/daglib/0015804}
D.~H. Ballard, \emph{An introduction to natural computation}, ser. Complex adaptive systems.\hskip 1em plus 0.5em minus 0.4em\relax {MIT} Press, 2000.

\bibitem{he2016deep}
K.~He, X.~Zhang, S.~Ren, and J.~Sun, ``Deep residual learning for image recognition,'' in \emph{Proceedings of the IEEE conference on computer vision and pattern recognition}, 2016, pp. 770--778.

\bibitem{cui2023sportsmot}
Y.~Cui, C.~Zeng, X.~Zhao, Y.~Yang, G.~Wu, and L.~Wang, ``Sportsmot: A large multi-object tracking dataset in multiple sports scenes,'' \emph{arXiv preprint arXiv:2304.05170}, 2023.

\bibitem{cioppa2023soccernet}
A.~Cioppa, S.~Giancola, V.~Somers, F.~Magera, X.~Zhou, H.~Mkhallati, A.~Deliege, J.~Held, C.~Hinojosa, A.~M. Mansourian \emph{et~al.}, ``Soccernet 2023 challenges results,'' \emph{arXiv preprint arXiv:2309.06006}, 2023.

\bibitem{soille2013morphological}
P.~Soille, \emph{Morphological image analysis: principles and applications}.\hskip 1em plus 0.5em minus 0.4em\relax Springer Science \& Business Media, 2013.

\bibitem{loshchilov2017decoupled}
I.~Loshchilov and F.~Hutter, ``Decoupled weight decay regularization,'' \emph{arXiv preprint arXiv:1711.05101}, 2017.

\bibitem{smith2019super}
L.~N. Smith and N.~Topin, ``Super-convergence: Very fast training of neural networks using large learning rates,'' in \emph{Artificial intelligence and machine learning for multi-domain operations applications}, vol. 11006.\hskip 1em plus 0.5em minus 0.4em\relax SPIE, 2019, pp. 369--386.

\bibitem{DBLP:conf/cvpr/CioppaGDKZCGD22}
A.~Cioppa, S.~Giancola, A.~Deli{\`{e}}ge, L.~Kang, X.~Zhou, Z.~Cheng, B.~Ghanem, and M.~V. Droogenbroeck, ``Soccernet-tracking: Multiple object tracking dataset and benchmark in soccer videos,'' in \emph{{CVPR} Workshops}, 2022.

\bibitem{zhou2019objects}
X.~Zhou, D.~Wang, and P.~Kr{\"a}henb{\"u}hl, ``Objects as points,'' \emph{arXiv preprint arXiv:1904.07850}, 2019.

\bibitem{kingma2014adam}
D.~P. Kingma, ``Adam: A method for stochastic optimization,'' \emph{arXiv preprint arXiv:1412.6980}, 2014.

\bibitem{DBLP:conf/mm/MaXSYZJ22}
Y.~Ma, G.~Xu, X.~Sun, M.~Yan, J.~Zhang, and R.~Ji, ``{X-CLIP:} end-to-end multi-grained contrastive learning for video-text retrieval,'' in \emph{{ACM} Multimedia}.\hskip 1em plus 0.5em minus 0.4em\relax {ACM}, 2022, pp. 638--647.

\bibitem{DBLP:conf/kdd/LiuZC18}
Y.~Liu, K.~Zhao, and G.~Cong, ``Efficient similar region search with deep metric learning,'' in \emph{{KDD}}.\hskip 1em plus 0.5em minus 0.4em\relax {ACM}, 2018, pp. 1850--1859.

\bibitem{ding2025streammind}
X.~Ding, H.~Wu, Y.~Yang, S.~Jiang, D.~Bai, Z.~Chen, and T.~Cao, ``Streammind: Unlocking full frame rate streaming video dialogue through event‑gated cognition,'' \emph{ICCV}, 2025.

\end{thebibliography}

\begin{IEEEbiography}
[{\includegraphics[width=1in,height=1.35in, clip,keepaspectratio]{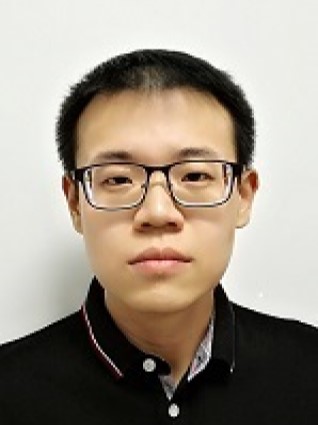}}]
\noindent\textbf{Zheng Wang} is currently a Principal Researcher and Huawei TopMinds at Huawei Technologies, Co., Ltd.. His current research interest focuses on multimodal search. He received his PhD degree at Nanyang Technological University. His research has been recognized by some prestigious awards, including Rising Star Award in Spatial Data Intelligence from ACM SIGSPATIAL China, one of the Best PhD Thesis Awards, WAIC Yunfan Award, Nominated Schmidt Science Fellows, Google PhD Fellowship, and AISG PhD Fellowship.
\end{IEEEbiography}
\vspace{-1.25cm}

\begin{IEEEbiography}
[{\includegraphics[width=1in,height=1.35in, clip,keepaspectratio]{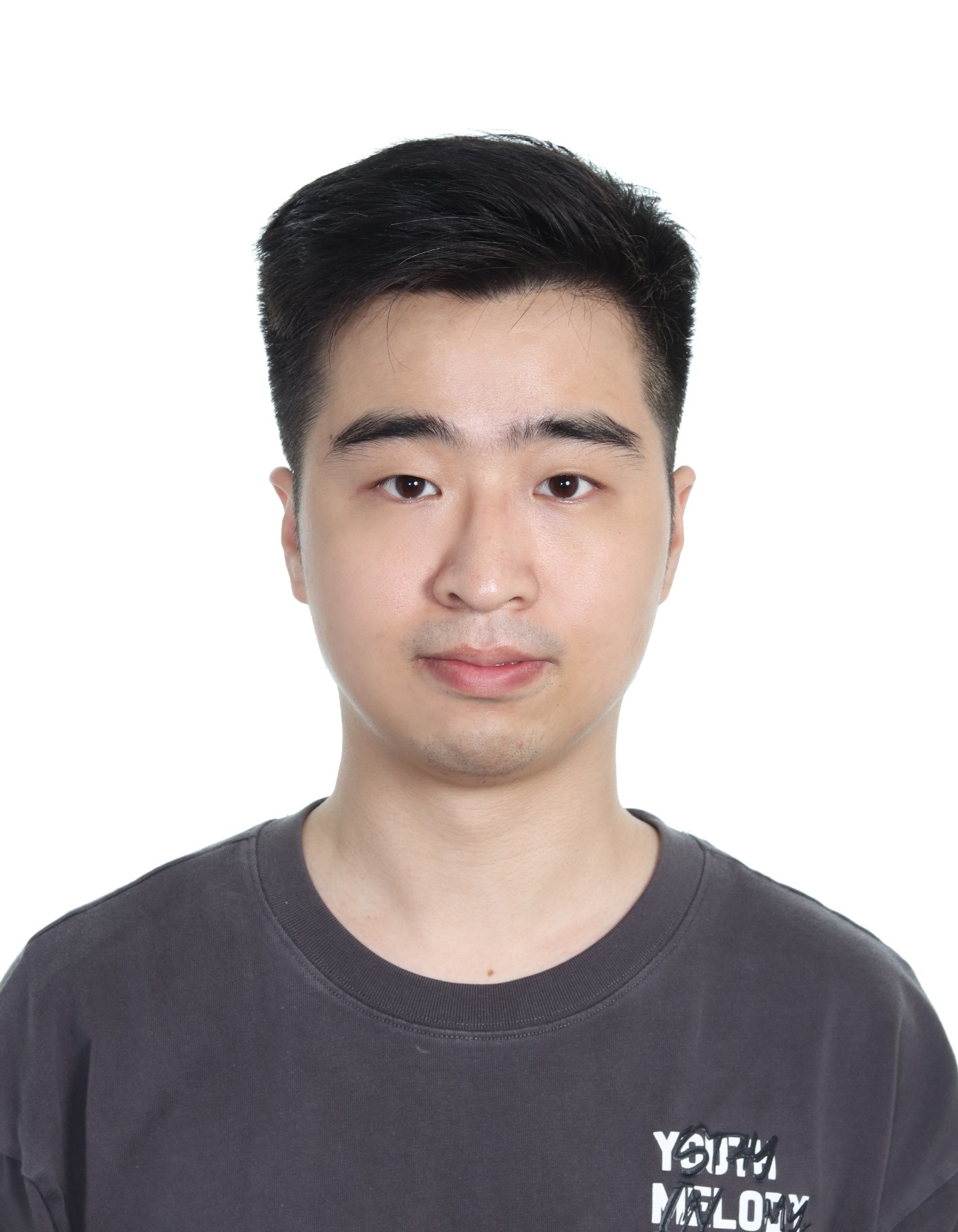}}]
\noindent\textbf{Shihao Xu} is a Research Scientist at Huawei Technologies, Co., Ltd., multimodal search and recommendation lab. His current research interests and works fill in multimodal applications including sports video representations, user intention
generation, multimodal geometry problem solving, and multimodal
prompting. He got his PhD degree in Nanyang Technological University at 2022, during which, he was working on the audio-visual understanding of human behaviours.
\end{IEEEbiography}
\vspace{-1.25cm}

\begin{IEEEbiography}
[{\includegraphics[width=1in,height=1.35in, clip,keepaspectratio]{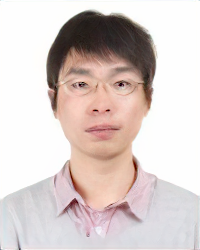}}]
\noindent\textbf{Wei Shi} is currently head of multimodal search team at Huawei Technologies, Co., Ltd.. He received his PhD degree at Department of Computer Science and Technology, Tsinghua University in 2015. His research interests are broadly in multimodal search, vision-language alignment, and big data systems.
\end{IEEEbiography}
\vspace{-1.25cm}

\end{document}